\documentclass{article} 
\usepackage{iclr2024_conference,times}

\usepackage{amsmath,amsfonts,bm}

 \def\eqref#1{(\ref{#1})}

\def\1{\bm{1}}

\def\vtheta{{\bm{\theta}}}

\def\vh{{\bm{h}}}

\def\vp{{\bm{p}}}

\def\vv{{\bm{v}}}
\def\vw{{\bm{w}}}
\def\vx{{\bm{x}}}
\def\vy{{\bm{y}}}

\def\mH{{\bm{H}}}

\def\mW{{\bm{W}}}

\DeclareMathAlphabet{\mathsfit}{\encodingdefault}{\sfdefault}{m}{sl}
\SetMathAlphabet{\mathsfit}{bold}{\encodingdefault}{\sfdefault}{bx}{n}

\newcommand{\R}{\mathbb{R}}

\DeclareMathOperator*{\argmax}{arg\,max}
\DeclareMathOperator*{\argmin}{arg\,min}

\usepackage[colorlinks]{hyperref}
\usepackage{url}

\usepackage{graphicx}
\usepackage{caption}
\usepackage{subcaption}
\usepackage{wrapfig}
\usepackage{enumitem} 

\usepackage{comment}

\usepackage{float}
\usepackage{blindtext}
\usepackage{amsmath}

\usepackage{minitoc}

\graphicspath{ {figs/} }

\def\0{\textbf{0}}
\def\1{\textbf{1}}
\def\a{\boldsymbol{a}}

\def\h{\boldsymbol{h}}
\def\k{\boldsymbol{k}}
\def\v{\boldsymbol{v}}

\def\u{\boldsymbol{u}}
\def\v{\boldsymbol{v}}
\def\w{\boldsymbol{w}}
\def\x{\boldsymbol{x}}
\def\y{\boldsymbol{y}}
\def\z{\boldsymbol{z}}
\def\\Delta{\boldsymbol{\Delta}}

\def\H{\boldsymbol{H}}

\def\I{\boldsymbol{I}}

\def\P{\boldsymbol{P}}

\def\W{\boldsymbol{W}}
\def\Y{\boldsymbol{Y}}

\DeclareMathOperator{\dist}{dist}

\DeclareMathOperator{\trace}{tr}

\DeclareMathOperator{\conv}{conv}

\def\st{\textrm{s.t.}}

\newcommand{\RR}{I\!\!R} 

\renewcommand{\mathbf}{\boldsymbol}

\newcommand{\mb}{\mathbf}
\newcommand{\mc}{\mathcal}

\newcommand{\bb}{\mathbb}

\newcommand{\paren}[1]{\left( #1 \right)}

\newcommand{\proj}{\mathrm{proj}}

\newcommand{\ol}{\overline}

\newcommand{\NC}{$\mc {NC}$}
\newcommand{\GNC}{$\mc {GNC}$}
\usepackage[toc, page]{appendix}
\newcommand{\setS}{\mathbb{S}}
\newcommand{\setR}{\mathbb{R}}

\newcommand{\norm}[2]{\left\| #1 \right\|_{#2}}

\def\ie{i.e.}
\def\eg{e.g.}

\DeclareMathOperator{\OB}{{\mathcal OB}}

\newcommand{\onevsrest}{\textup{one-vs-rest}}
\newcommand{\onevsone}{\textup{one-vs-one}}
\usepackage{amsthm}
\usepackage{cleveref}
\theoremstyle{plain}

\newtheorem{theorem}{Theorem}[section]

\newtheorem{lemma}[theorem]{Lemma}

\newtheorem{definition}[theorem]{Definition}

\newcommand{\vct}[1]{\boldsymbol{#1}}

\newcommand{\vTheta}{\vct{\Theta}}

\newcommand{\zz}[1]{\textcolor{blue}{ [{\em Zhihui:} #1]}}

\newcommand{\cy}[1]{\textcolor{red}{ [{\em CY:} #1]}}

\newcommand{\jz}[1]{\textcolor{green}{ [{\em JZ:} #1]}}
\newcommand{\jc}[1]{\textcolor{magenta}{ [{\em JC:} #1]}}

\title{Generalized Neural Collapse for a Large Number of Classes}

\author{Jiachen Jiang$^*$\\
Ohio State University\\
\texttt{jiang.2880@osu.edu} \\
\And
Jinxin Zhou\thanks{ The first two authors contribute equally to this work.} \\
Ohio State University\\
\texttt{zhou.3820@osu.edu} \\
\And
Peng Wang \\
University of Michigan\\
\texttt{pengwa@umich.edu} \\
\And
Qing Qu \\
University of Michigan\\
\texttt{qingqu@umich.edu} \\
\And
Dustin Mixon \\
Ohio State University\\
\texttt{mixon.23@osu.edu} \\
\And
Chong You$^\dag$ \\
Google Research\\
\texttt{cyou@google.com} \\
\And
Zhihui Zhu\thanks{Corresponding authors: CY and ZZ.} \\
Ohio State University\\
\texttt{zhu.3440@osu.edu} 
}
\iclrfinalcopy 
\begin{document}


\maketitle

\begin{abstract}
Neural collapse provides an elegant mathematical characterization of learned last layer representations (a.k.a. features) and classifier weights in deep classification models. 
Such results not only provide insights but also motivate new techniques for improving practical deep models. 
However, most of the existing empirical and theoretical studies in neural collapse focus on the case that the number of classes is small relative to the dimension of the feature space. 
This paper extends neural collapse to cases where the number of classes are much larger than the dimension of feature space, which broadly occur for language models, retrieval systems, and face recognition applications. 
We show that the features and classifier exhibit a generalized neural collapse phenomenon, where the minimum one-vs-rest margins is maximized.
We provide empirical study to verify the occurrence of generalized neural collapse in practical deep neural networks. 
Moreover, we provide theoretical study to show that the generalized neural collapse provably occurs under unconstrained feature model with spherical constraint, under certain technical conditions on feature dimension and number of classes.

\end{abstract}

\section{Introduction}\label{sec:intro}
Over the past decade, deep learning algorithms have achieved remarkable progress across numerous machine learning tasks and have significantly enhanced the state-of-the-art in many practical applications ranging from computer vision to natural language processing and retrieval systems. Despite their tremendous success, a comprehensive understanding of the features learned from deep neural networks (DNNs) is still lacking. The recent work \cite{papyan2020prevalence, papyan2020traces} has empirically uncovered an intriguing phenomenon regarding the last-layer features and classifier of DNNs, called  \emph{Neural Collapse} (\NC)  that can be briefly summarized as the following characteristics:
\begin{itemize}[leftmargin=0.12in,topsep=-0.1em,itemsep=0.11em]
    \item \textbf{\emph{Variability Collapse} ($\mathcal{NC}_1$):} Within-class variability of features collapses to zero.
    \item \textbf{\emph{Convergence to Simplex ETF} ($\mathcal{NC}_2$):} 
    Class-mean features converge to a simplex Equiangular Tight Frame (ETF), achieving equal lengths, equal pair-wise angles, and  maximal distance in the feature space.  
    \item  \textbf{\emph{Self-Duality} ($\mathcal{NC}_3$):} Linear classifiers converge to class-mean features, up to a global rescaling. 
\end{itemize}
Neural collapse provides a mathematically elegant characterization of learned representations or features in deep learning based classification models, independent of network architectures, dataset properties, and optimization algorithms. Building on the so-called \emph{unconstrained feature model} \citep{mixon2020neural} or the \emph{layer-peeled model} \citep{fang2021exploring}, subsequent research \citep{zhu2021geometric, lu2020neural, ji2021unconstrained, yarasneural, wojtowytsch2020emergence, jiunconstrained, zhouall, hanneural, tirer2022extended, zhou2022optimization, poggio2020explicit, thrampoulidis2022imbalance, tirer2023perturbation, nguyen2022memorization} has provided theoretical evidence for the existence of the \NC\ phenomenon when using a family of loss functions including cross-entropy (CE) loss, mean-square-error (MSE) loss and variants of CE loss. Theoretical results regarding \NC\ not only contribute to a new understanding of the working of DNNs but also provide inspiration for developing new techniques to enhance their practical performance in various settings, such as imbalanced learning \citep{xie2023neural, liu2023inducing}, transfer learning \citep{galanti2022generalization, li2022principled, xie2022hidden, galanti2022role}, continual learning \citep{yu2022continual,yang2023neural}, loss and architecture designs \citep{chan2022redunet, yu2020learning,zhu2021geometric}, etc.  

However, most of the existing empirical and theoretical studies in \NC\ focus on the case that the number of classes is small relative to the dimension of the feature space. Nevertheless, there are many cases in practice where the number of classes can be extremely large, such as
\begin{itemize} [leftmargin=0.13in,topsep=0.01em,itemsep=0.01em]
    \item Person identification \citep{deng2019arcface}, where each identity is regarded as one class. 
    \item Language models \citep{devlin2018bert}, where the number of classes equals the vocabulary size\footnote{Language models are usually trained to classify a token (or a collection of them) that is either masked in the input (as in BERT \citep{devlin2018bert}), or the next one following the context (as in language modeling), or a span of masked tokens in the input (as in T5 \citep{raffel2020exploring}), etc. In such cases, the number of classes is equal to the number of all possible tokens, \ie, the vocabulary size.}.
    \item Retrieval systems \citep{mitra2018introduction}, where each document in the dataset represents one class.
    \item Contrastive learning \citep{chen2020simple}, where each training data can be regarded as one class.
\end{itemize}

In such cases, it is usually infeasible to have a feature dimension commensurate with the number of classes due to computational and memory constraints. Therefore, it is crucial to develop a comprehensive understanding of the characteristics of learned features in such cases, particularly with the increasing use of web-scale datasets that have a vast number of classes.

\begin{figure} [!t]
     \centering
     \begin{subfigure}[b]{0.32\textwidth}
         \centering
         \includegraphics[width=0.9\textwidth]{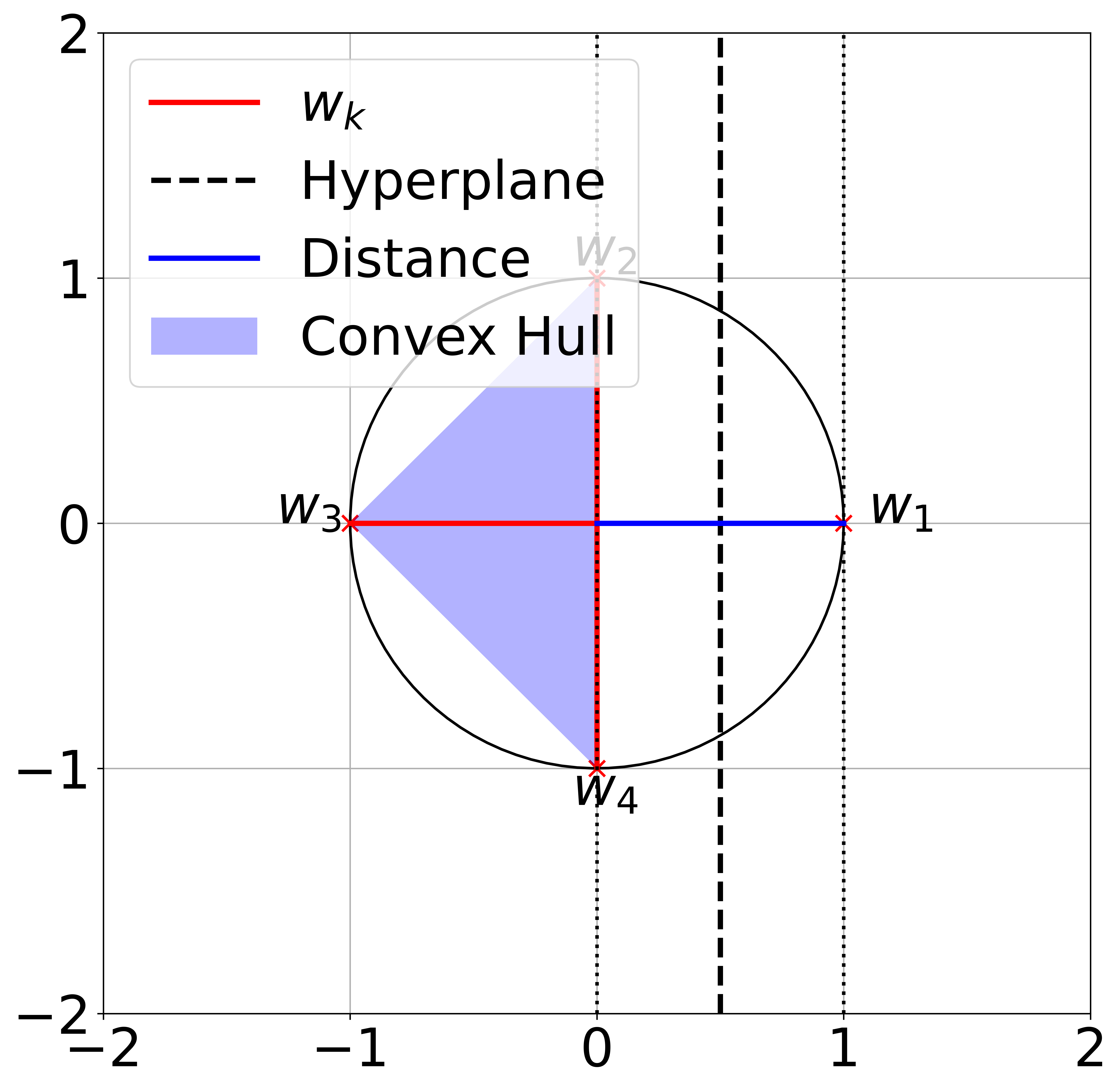}
         \caption{One-vs-rest distance: Case 1}
         \label{fig:uniform_convexhull_margin}
     \end{subfigure}
     \hfill
     \begin{subfigure}[b]{0.32\textwidth}
         \centering
         \includegraphics[width=0.9\textwidth]{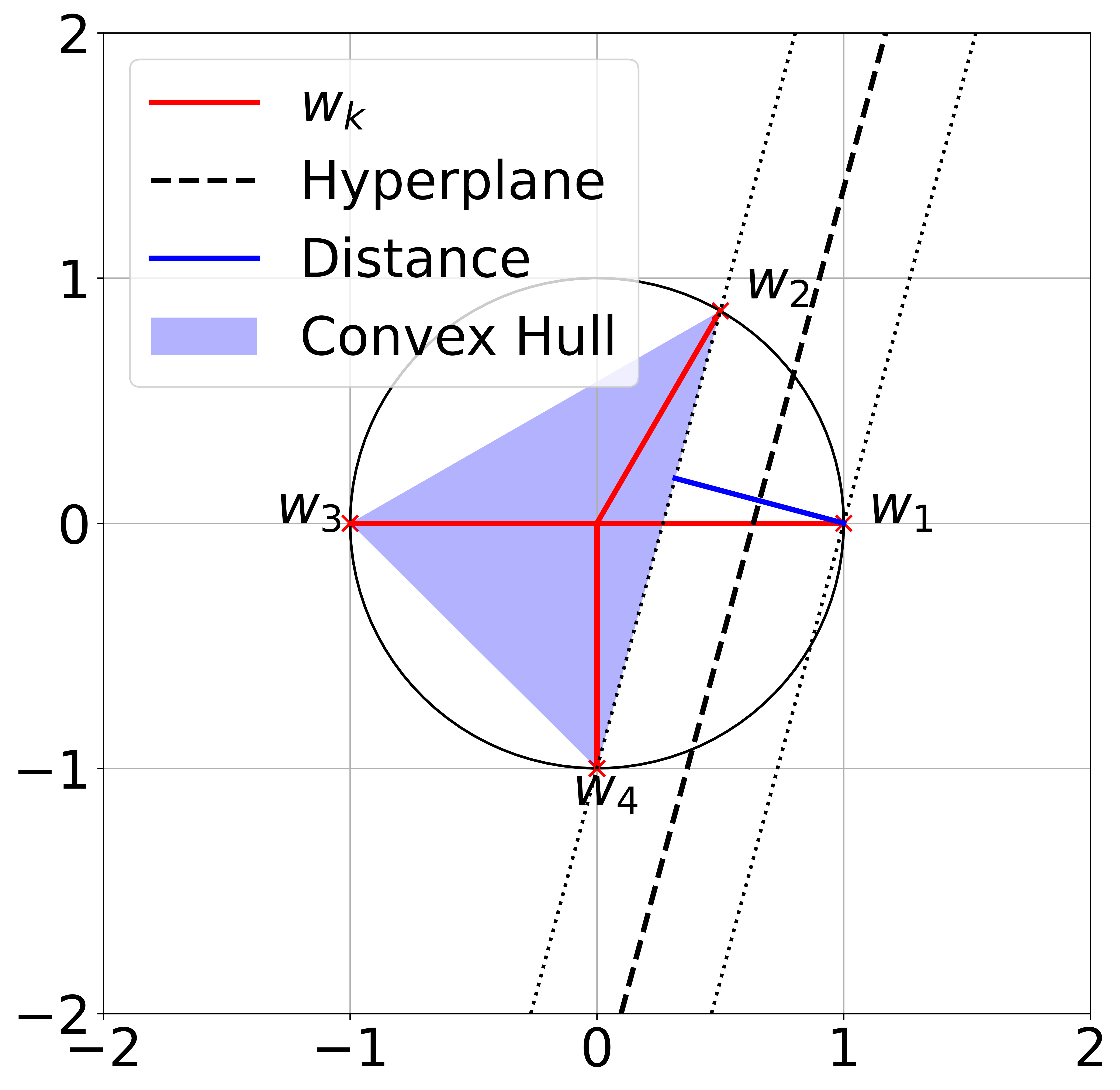}
          \caption{One-vs-rest distance: Case 2}
         \label{fig:nonuniform_convexhull_margin}
     \end{subfigure}
     \hfill
     \begin{subfigure}[b]{0.32\textwidth}
    
         \centering
         \includegraphics[width=0.9\textwidth]{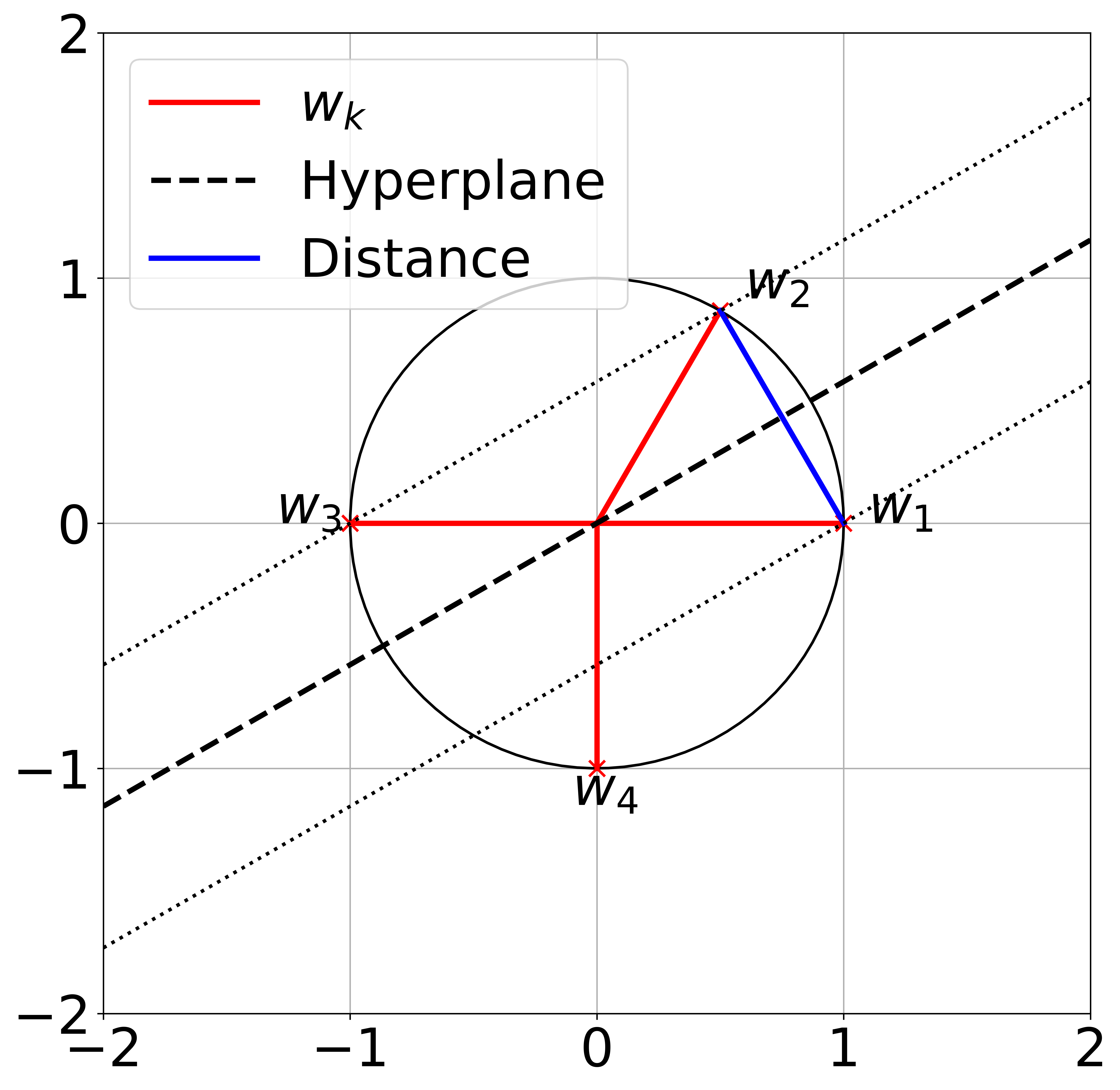}
         \caption{One-vs-one distance}
         \label{fig:tammes-margin}
     \end{subfigure}
    
        \caption{In Generalized Neural Collapse (\GNC), the optimal classifier weight $\{\vw_k\}$ is a \emph{Softmax Code} defined from maximizing the \emph{one-vs-rest distance} (see \Cref{def:softmax-code}). \emph{(a, b)} Illustration of the one-vs-rest distance using the example of $\vw_1$-vs-$\{\w_2, \w_3, \w_4\}$ distance, under two configurations of $\{\vw_k\}_{k=1}^4$ in a two-dimensional space. The distance in Case 1 is larger than that in Case 2. \emph{(c)} Illustration of the \emph{one-vs-one distance} used to define the Tammes problem (see Eq.~\eqref{eq:tammes-problem}). We prove \GNC\ under technical conditions on Softmax Code and Tammes problem (see \Cref{sec:main_result}).}
        \label{fig:comparison-tammes}
\vspace{-.2in}
\end{figure}

\paragraph{Contributions.} 
This paper studies the geometric properties of the learned last-layer features and the classifiers for cases where the number of classes can be arbitrarily large compared to the feature dimension. 
Motivated by the use of spherical constraints in learning with a large number of classes, such as person identification and contrastive learning, we consider networks trained with \emph{spherical constraints} on the features and classifiers. Our contributions can be summarized as follows.
\begin{itemize}[leftmargin=0.13in,topsep=0.2em,itemsep=0.11em]

    \item \textbf{The Arrangement Problem: Generalizing \NC\ to a Large Number of Classes.}    
    In \Cref{sec:setup} we introduce the generalized \NC\ (\GNC) for characterizing the last-layer features and classifier. 
    In particular, \GNC$_1$\ and \GNC$_3$ state the same as \NC$_1$\ and \NC$_3$, respectively. 
    \GNC$_2$ states that the classifier weight is a \emph{Softmax Code}, which generalizes the notion of a simplex ETF and is defined as the collection of points on the unit hyper-sphere that maximizes the minimum one-vs-all distance (see \Cref{fig:comparison-tammes} (a,b) for an illustration). 
    Empirically, we verify that the \GNC\ approximately holds in practical DNNs trained with a small temperature in CE loss.
    Furthermore, we conduct theoretical study in \Cref{sec:main_result} to show that under the unconstrained features model (UFM)~\citep{mixon2020neural,fang2021exploring,zhu2021geometric} and with a vanishing temperature, the global solutions satisfy \GNC\ under technical conditions on Softmax Code and solutions to the Tammes problem \citep{tammes1930origin}, the latter defined as a collection of points on the unit hyper-sphere that maximizes the minimum one-vs-one distance (see \Cref{fig:comparison-tammes}(c) for an illustration).

    \item \textbf{The Assignment Problem: Implicit Regularization of Class Semantic Similarity.}
    Unlike the simplex ETF (as in \NC$_2$) in which the distance between any pair of vectors is the same, not all pairs in a Softmax Code (as in \GNC$_2$) are of equal distant when the number of classes is greater than the feature space dimension.
    This leads to the ``assignment'' problem, \ie, the correspondence between the classes and the weights in a Softmax Code.
    In \Cref{subsec:permutation}, we show empirically an implicit regularization effect by the semantic similarity of the classes, \ie, conceptually similar classes (\eg, Cat and Dog) are often assigned to closer classifier weights in Softmax Code, compared to those that are conceptually dissimilar (\eg, Cat and Truck). Moreover, such an implicit regularization is beneficial, \ie, enforcing other assignments produces inferior model quality.

    \item \textbf{Cost Reduction for Practical Network Training/Fine-tuning.} The universality of alignment between classifier weights and class means (i.e., \GNC$_3$) implies that training the classifier is unnecessary and the weight can be simply replaced by the class-mean features. Our experiments in \Cref{subsec:ocm} demonstrate that such a strategy achieves comparable performance to classical training methods, and even better out-of-distribution performance than classical fine-tuning methods with significantly reduced parameters.
\end{itemize}

\paragraph{Related work.} The recent work \cite{liu2023generalizing} also introduces a notion of 
generalized \NC\ for the case of large number of classes, 
which predicts equal-spaced features. However, their work focuses on networks trained with weight decay, for which empirical results in \Cref{app:exp} and \cite{yaras2023neural} show to not produce equal-length and equal-spaced features for a relatively large number of classes.
Moreover, the work \cite{liu2023generalizing} relies on a specific choice of kernel function to describe the uniformity. Instead, we concretely define \GNC$_2$ through the softmax code. When preparing this submission, we notice a concurrent work \cite{gao2023study} that provides analysis for generalized \NC, but again for networks trained with weight decay. In addition, that work analyzes gradient flow for the corresponding UFM with a particular choice of weight decay, while our work studies the global optimality of the training problem. The work \cite{zhou2022optimization} empirically shows that MSE loss is inferior to the CE loss when $K>d+1$, but no formal analysis is provided for CE loss. Finally, the global optimality of the UFM with spherical constraints has been studied in \cite{lu2022neural,yaras2023neural} but only for the cases $K\le d+1$ or $K\rightarrow \infty$. 

\section{Generalized Neural Collapse for A Large Number of Classes}\label{sec:setup}
In this section, we begin by providing a brief overview of DNNs and introducing notations used in this study in \Cref{subsec:concepts}. We will also introduce the concept of the UFM which is used in theoretical study of the subsequent section.
Next, we introduce the notion of \emph{Softmax Code} for describing the distribution of a collection of points on the unit sphere, which prepares us to present a formal definition of \emph{Generalized Neural Collapse} and empirical verification of its validity in \Cref{subsec:gnc}.

\subsection{Basics Concepts of DNNs}\label{subsec:concepts}

A DNN classifier aims to learn a feature mapping $\phi_{\vtheta}(\cdot):\mathbb{R}^D\rightarrow\mathbb{R}^d$ with learnable parameters $\vtheta$ 
that maps from input $\x \in \mathbb{R}^D$ to a deep representation called the feature $\phi_{\vtheta}(\x) \in \mathbb{R}^d$, and a linear classifier $\mW = \left[ \vw_1 \quad \vw_2 \quad \cdots \quad \vw_K\right] \in \mathbb{R}^{d \times K}$ such that the output (also known as the logits) $\Psi_{\vTheta}(\x) =  \mW^\top \phi_{\vtheta} \paren{\x} \in\mathbb{R}^K$ can make a correct prediction. Here, $\vTheta=\{\vtheta,\mW\}$ represents \emph{all} the learnable parameters of the DNN.\footnote{We ignore the bias term in the linear classifier since $(i)$ the bias term is used to compensate the global mean of the features and vanishes when the global mean is zero \citep{papyan2020prevalence,zhu2021geometric}, $(ii)$ it is the default setting across a wide range of applications such as person identification \citep{wang2018cosface, deng2019arcface}, contrastive learning \citep{chen2020simple,chen2021exploring}, etc.} 

Given a balanced training set $\left\{\paren{\x_{k,i}, \y_k}\right\}_{i\in[n],k\in[K]} \subseteq \mathbb{R}^D \times \mathbb{R}^K$, where $\vx_{k,i}$ is the $i$-th sample in the $k$-th class and $\vy_k$ is the corresponding one-hot label with all zero entries except for unity in the $k$-th entry, the network parameters $\vTheta$ 
are typically optimized by minimizing the following CE loss  
\begin{align}
    \min_{\vTheta} \frac{1}{nK} \sum_{k=1}^{K} \sum_{i=1}^{n} \mathcal{L}_{\text{CE}}\paren{\Psi_{\vTheta} \paren{\vx_{k,i}} , \vy_k, \tau}, \ \mathcal{L}_{\text{CE}}\paren{\z, \y_k, \tau} = -\log \Big(\frac{\exp(z_k/\tau)}{\sum_{j=1}^{K} \exp(z_j/\tau)}\Big).
    \label{def:ce}
\end{align}
{In above, we assume that a spherical constraint is imposed on the feature and classifier weights and that the logit $z_k$ is divided by the temperature parameter $\tau$. This is a common practice when dealing with a large number of classes \citep{wang2018cosface, chang2019pre,chen2020simple}.}
Specifically, we enforce $\{\vw_k, \phi_{\vTheta}(\vx_{k,i})\} \subseteq \mathbb{S}^{d-1}:= \{\a \in \mathbb{R}^{d} : \|\a\|_2 = 1 \}$ for all $i\in [n]$ and $k\in [K]$.
{An alternative regularization is weight decay on the model parameters $\vTheta$, the effect of which we study in \Cref{app:exp}.}

To simplify the notation, we denote the \textit{oblique manifold} embedded in Euclidean space by $\OB(d,K) := \left\{ \mW \in \setR^{d \times K} \ | \ \vw_{k} \in \setS^{d-1}, \ \forall k \in [K] \right\}$. In addition, we denote the last-layer features by $\vh_{k,i}:=\phi_{\vtheta}(\vx_{k,i})$. 
We rewrite all the features in a matrix form as
\[
\H :=  \left[ \H_1 \quad \H_2 \quad \cdots \quad \H_K\right] \in \mathbb{R}^{d \times nK}, \text {with}\ \mb H_k:= \begin{bmatrix}\mb h_{k,1} & \cdots & \mb h_{k,n} \end{bmatrix} \in \bb R^{ d \times n}.
\]
Also we denote by $\overline{\vh}_k := \frac{1}{n} \sum_{i = 1}^{n} \vh_{k,i}$ the class-mean feature for each class.

\paragraph{Unconstrained Features Model (UFM).} The UFM \citep{mixon2020neural} or layer-peeled model \citep{fang2021exploring}, wherein the last-layer features are treated as free optimization variables, are widely used for theoretically understanding the \NC\ phenomena.
In this paper, we will consider the following UFM with a spherical constraint on classifier weights $\mW$ and unconstrained features $\mH$: 
\begin{equation}\label{eqn:ufm-sphere}
        \min_{\mW, \mH} \frac{1}{nK}\sum_{k=1}^{K} \sum_{i=1}^{n} \mathcal{L}_{\text{CE}}\paren{\mW^\top\vh_{k,i} , \vy_k,\tau} \quad \st \quad \mW\in\OB(d,K),\ \mH\in\OB(d,nK).
\end{equation}

\subsection{Generalized Neural Collapse}\label{subsec:gnc}

{
We start by introducing the notion of \emph{softmax code} which will be used for describing \GNC.
\begin{definition} [Softmax Code] Given positive integers $d$ and $K$, a softmax code is an arrangement of $K$ points on a unit sphere of $\mathbb{R}^d$ that maximizes the minimal distance between one point and the convex hull of the others:
    \begin{align}\label{eq:softmax-code}
    \max_{\mW\in\OB(d,K)} \rho_\onevsrest(\mW), ~~\text{where}~~\rho_\onevsrest(\mW) \doteq \min_{k} \dist\!\Big(\vw_k, \{\vw_j\}_{j\in[K]\setminus k}\Big).
    \end{align}
    In above, the distance between a point $\v$ and a set $\mathcal{W}$ is defined as $\dist(\v, \mathcal{W}) = \inf_{\w \in \conv(\mathcal{W})}\{\|\v - \w\|\}$, where $\conv(\cdot)$ denotes the convex hull of a set.
\label{def:softmax-code}\end{definition}
}

We now extend  \NC\ to the \emph{Generalized Neural Collapse} (\GNC) that captures the properties of the features and classifiers at the terminal phase of training. 
With a vanishing temperature (i.e., $\tau \rightarrow 0$),
the last-layer features and classifier exhibit the following \GNC\ phenomenon: 
\begin{itemize}[leftmargin=0.12in,topsep=0.01em,itemsep=0.01em]
    \item \textbf{\emph{Variability Collapse} ($\mathcal{GNC}_1$).} All features of the same class collapse to the corresponding class mean. 
     Formally, as used in \cite{papyan2020prevalence}, the quantity $\mathcal{GNC}_1 \doteq \frac{1}{K} \trace \big( {\boldsymbol{\Sigma}_{W} \boldsymbol{\Sigma}_B^{\dagger} }\big)\rightarrow 0$, where $\boldsymbol{\Sigma}_B := \frac{1}{K} \sum_{k=1}^{K} \overline{\vh}_k\overline{\vh}_k^\top$ and $\boldsymbol{\Sigma}_W := \frac{1}{nK} \sum_{ k =1}^{k} \sum_{i=1}^{n} \paren{\vh_{k,i} - \overline{\vh}_k} \paren{\vh_{k,i} - \overline{\vh}_k}^\top$ denote the between-class and within-class covariance matrices, respectively. 
    
    \item \textbf{\emph{Softmax Codes} ($\mathcal{GNC}_2$).} 
    Classifier weights converge to the softmax code in \cref{def:softmax-code}. This property may be measured by $\mathcal{GNC}_2 \doteq \rho_\onevsrest(\mW) \to \max_{\mW\in\OB(d,K)}\rho_\onevsrest(\mW)$. 
    
    \item  \textbf{\emph{Self-Duality} ($\mathcal{GNC}_3$).} Linear classifiers converge to the class-mean features. Formally, this alignment can be measured by $\mathcal{GNC}_3 \doteq \frac{1}{K} \sum_{k =1}^{K} \paren{1 - \w_{k}^{\top} \ol\vh_{k}} \rightarrow 0$.
\end{itemize}

\begin{figure}[!ht]
    \centering
    {\includegraphics[width=0.23\textwidth]{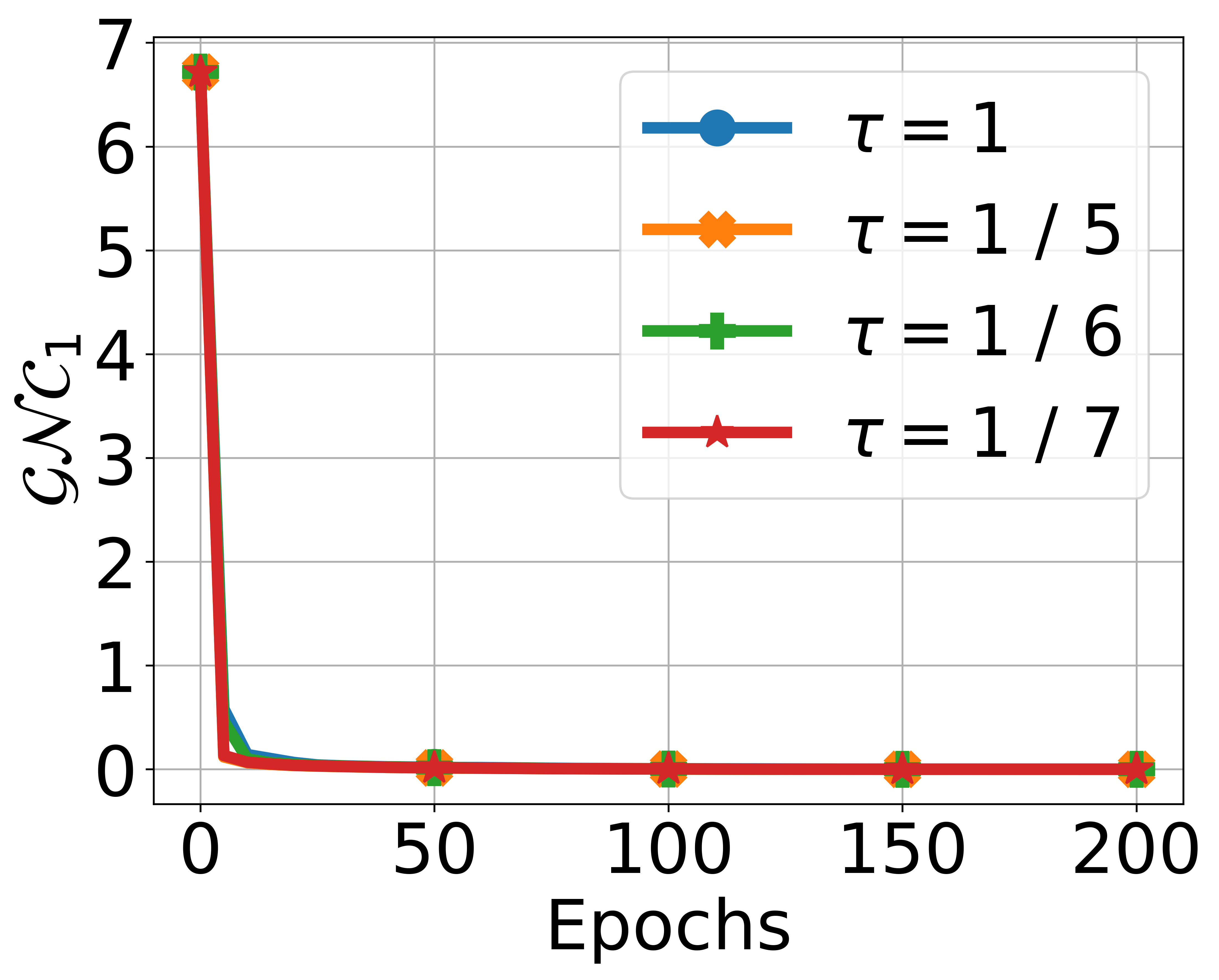}} 
    \hfill
  {\includegraphics[width=0.24\textwidth]{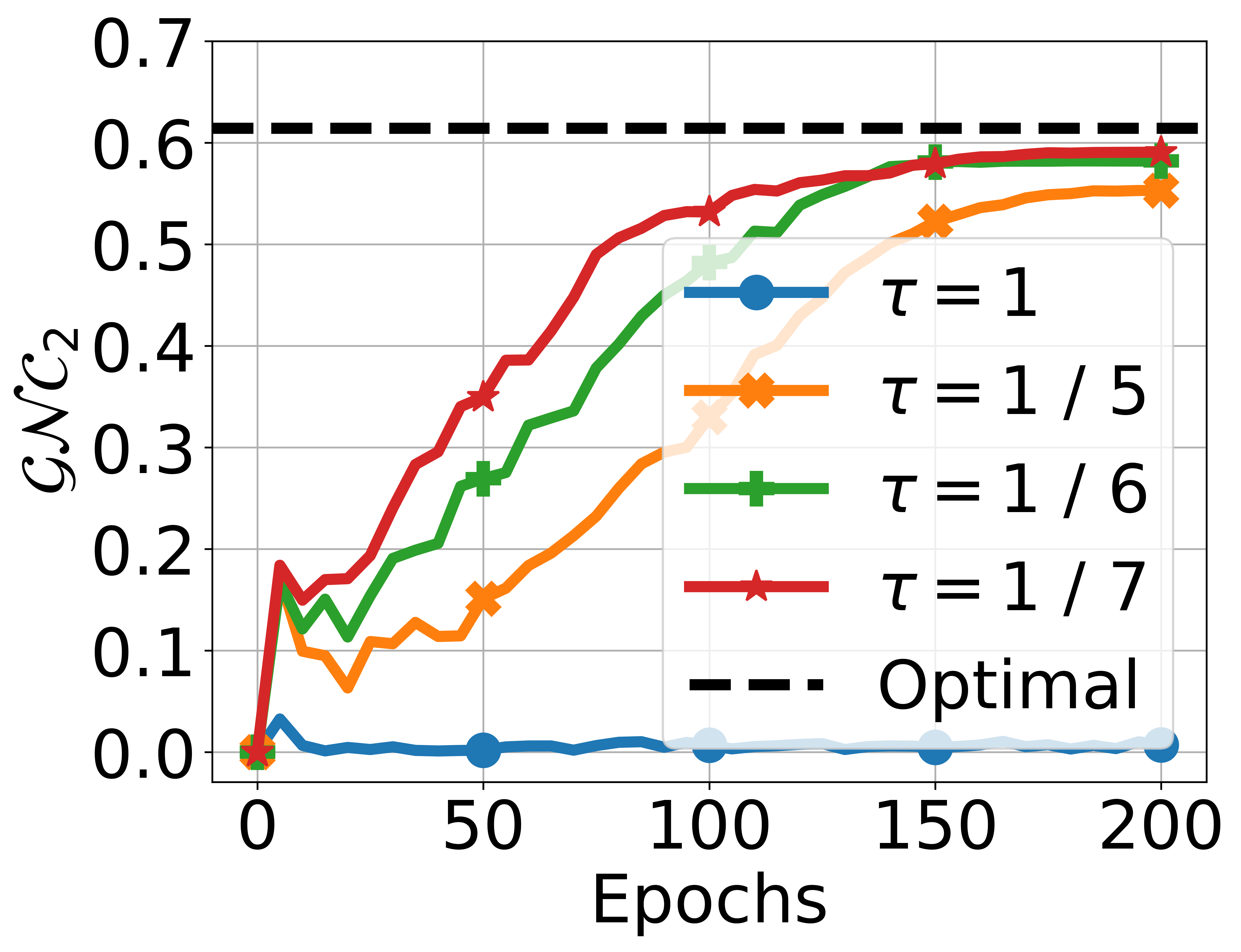}} 
  \hfill
{\includegraphics[width=0.24\textwidth]{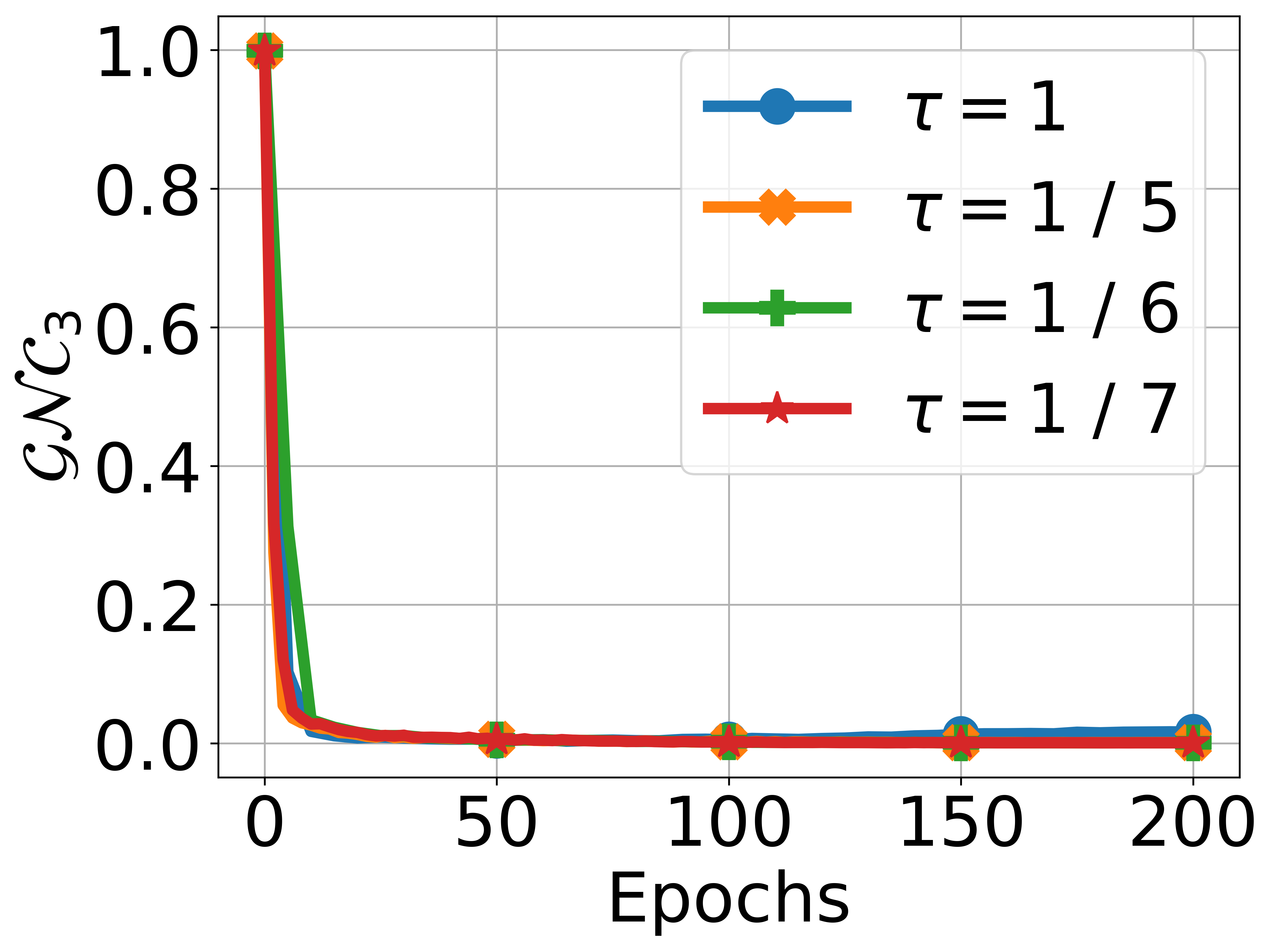}}   
\hfill
   {\includegraphics[width=0.23\textwidth]{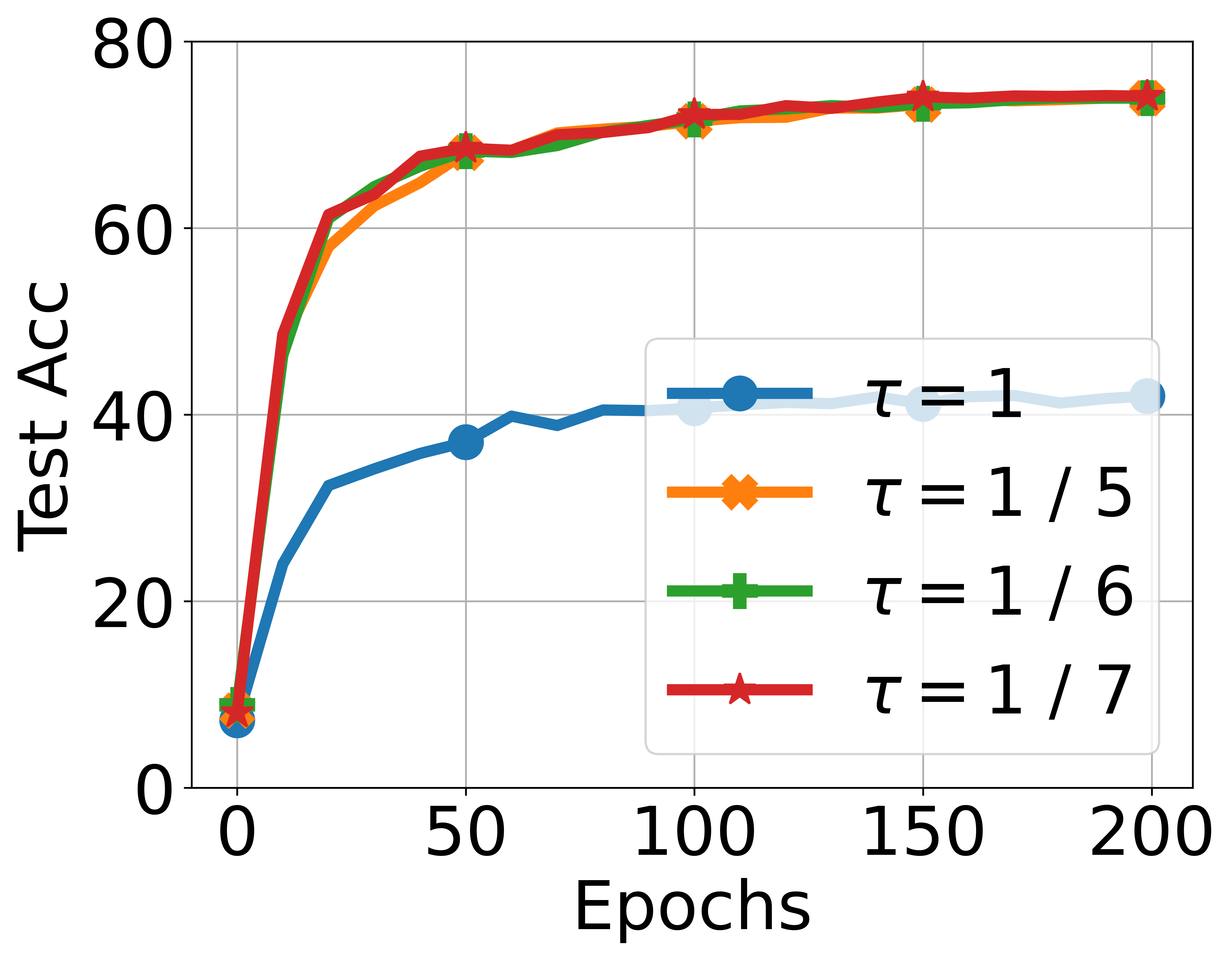}}
    \caption{\textbf{Illustration of \GNC\ and test accuracy across different temperatures $\tau$} in training a ResNet18 on CIFAR100 with $d=10$ and $K=100$. ``Optimal'' in the second left figure refers to $\max_{\mW\in\OB(d,K)}\rho_\onevsrest(\mW)$.
        }
    \label{fig:NC-dynamic}
\end{figure}

{
The main difference between \GNC\ and \NC\ lies in \GNC$_2$ / \NC$_2$, which describe the configuration of the classifier weight $\W$. 
In \NC$_2$, the classifier weights corresponding to different classes are described as a simplex ETF, which is a configuration of vectors that have equal pair-wise distance and that distance is maximized.
Such a configuration does not exist in general when the number of classes is large, i.e., $K > d + 1$. 
\GNC$_2$ introduces a new configuration described by the notion of softmax code. 
By \Cref{def:softmax-code}, a softmax code is a configuration where each vector is maximally separated from all the other points, measured by its distance to their convex hull.
Such a definition is motivated from theoretical analysis (see \Cref{sec:main_result}). In particular, it reduces to simplex ETF when $K \le d + 1$ (see \Cref{thm:special-case-NC3}). 
}

{
\paragraph{Interpretation of Softmax Code.}  
Softmax Code abides a max-distance interpretation. 
Specifically, consider the features $\{\vh_{k,i}\}_{k \in [K], i \in [n]}$ from $n$ classes. In multi-class classification, one commonly used distance (or margin) measurement is the one-vs-rest (also called one-vs-all or one-vs-other) distance \citep{murphy2022probabilistic}, i.e., the distance of class $k$ vis-a-vis other classes. Noting that the  distance between two classes is equivalent to the distance between the convex hulls of the data from each class \citep{murphy2022probabilistic}, the distance of class $k$ vis-a-vis other classes is given by $\dist(\{\vh_{k, i}\}_{i \in [n]}, \{\vh_{k', i}\}_{k'\in[K]\setminus k, i\in [n]})$. From \GNC$_1$ and \GNC$_3$ we can rewrite the distance as
\begin{equation}
    \dist\!\big(\{\vh_{k, i}\}_{i \in [n]}, \{\vh_{k', i}\}_{k'\in[K]\setminus k, i\in [n]} \big) 
    = \dist\!\big(\overline{\vh}_k, \{\overline{\vh}_{k'}\}_{k'\in[K]\setminus k}\big) 
    = \dist\!\big(\vw_k, \{\vw_{k'}\}_{k'\in[K]\setminus k}\big). 
\end{equation}
By noticing that the rightmost term is minimized in a Softmax Code, it follows from \GNC$_2$ that the learned features have the property that their one-vs-rest distance minimized over all classes $k \in [K]$ is maximized. 
In other words, measured by one-vs-rest distance, the learned features are such that they are maximally separated. 
Finally, we mention that the separation of classes may be characterized by other measures of distance as well, such as the one-vs-one distance (also known as the sample margin in \cite{cao2019learning, zhou2022learning}) which leads to the well-known Tammes problem. We will discuss this in \Cref{subsec:GNC}.
}

\paragraph{Experimental Verification of \GNC.}  
We verify the occurence of \GNC\ by training a ResNet18 \citep{he2016deep} for image classification on the CIFAR100 dataset \citep{krizhevsky2009learning}, and report the results in \Cref{fig:NC-dynamic}.
To simulate the case of $K > d + 1$, we use a modified ResNet18 where the feature dimension is $10$. From \Cref{fig:NC-dynamic}, we can observe that both \GNC$_1$ and \GNC$_3$ converge to $0$, and \GNC$_2$ converges towards the spherical code with relatively small temperature $\tau$. Additionally, selecting a small $\tau$ is not only necessary for achieving \GNC, but also for attaining high testing performance. Due to limited space, we present experimental details and other experiments with different architectures and datasets in \Cref{app:exp}. In the next section, we provide a theoretical justification for \GNC\ under UFM in \eqref{eqn:ufm-sphere}.

\section{Theoretical Analysis of GNC}\label{sec:main_result}
In this section, we provide a theoretical analysis of \GNC\ under the UFM in \eqref{eqn:ufm-sphere}. 
We first show  in \Cref{subsec:hardmax} that under appropriate temperature parameters, the solution to \eqref{eqn:ufm-sphere} can be approximated by the solution to a ``HardMax'' problem, which is of a simpler form amenable for subsequent analysis.
We then provide a theoretical analysis of \GNC\ in \Cref{subsec:GNC}, by first proving the optimal classifier forms a Softmax Code (\GNC$_2$), and then establishing \GNC$_1$ and \GNC$_3$ under technical conditions on Softmax Code and solutions to the Tammes problem. In addition, we provide insights for the design of feature dimension $d$ given a number of classes $K$ by analyzing the upper and lower bound for the one-vs-rest distance of a Softmax Code.
All proofs can be found in \Cref{app:thm}. 

\subsection{Preparation: the Asymptotic CE Loss}\label{subsec:hardmax}


Due to the nature of the softmax function which blends the output vector, analyzing the CE loss can be difficult even for the unconstrained features model. The previous work \cite{yaras2023neural} analyzing the case $K \leq d+1$ relies on the simple structure of the global solutions, where the classifiers form a simplex ETF. However, this approach cannot be directly applied to the case $K > d+1$ due to the absence of an informative characterization of the global solution. 
Motivated by the fact that the temperature $\tau$ is often selected as a small value ($\tau<1$, e.g., $\tau = 1/30$ in \cite{wang2018cosface}) in practical applications \citep{wang2018cosface, chen2021exploring}, we consider the case of $\tau\rightarrow 0$ where the CE loss \eqref{eqn:ufm-sphere} converges to the following ``HardMax'' problem:
\begin{equation}
    \label{def:hardmax}
    \begin{aligned}
        \min_{
        \substack{\mW\in\OB(d,K)\\\mH\in\OB(d,nK)}} 
        \mathcal{L}_{\text{HardMax}}(\mW, \mH), ~\text{where}~ \mathcal{L}_{\text{HardMax}}(\mW, \mH) \doteq \max_{k \in [K]} \max_{i \in [n]} \max_{k' \ne k} \langle \w_{k'} - \w_{k} , \h_{k,i} \rangle,
    \end{aligned}
\end{equation}
where $\langle\cdot , \cdot \rangle$ denotes the inner-product operator. More precisely, we have the following result.

\begin{lemma}[Convergence to the HardMax problem] For any positive integers $K$ and $n$, we have
\begin{equation}\label{eq:converge-to-hardmax}
    \limsup_{\tau \to 0} \paren{ 
    \argmin_{\substack{\mW\in\OB(d,K)\\\mH\in\OB(d,nK)}}  
    \frac{1}{nK}\sum_{k=1}^{K} \sum_{i=1}^{n} \mathcal{L}_{\text{CE}}\paren{\mW^\top\vh_{k,i} , \vy_k,\tau} }
    \! \subseteq \!
    \argmin_{\substack{\mW\in\OB(d,K)\\\mH\in\OB(d,nK)}} 
    \mathcal{L}_{\text{HardMax}}(\mW, \mH). \!\!
    \vspace{-1em}
\end{equation}
\label{thm:converge-to-hardmax}
\end{lemma}
Our goal is not to replace CE with the HardMax function in practice. 
Instead, we will analyze the HardMax problem in \eqref{def:hardmax} to gain insight into the global solutions and the \GNC\ phenomenon.

\subsection{Main Result: Theoretical Analysis of \GNC}
\label{subsec:GNC}

\paragraph{\GNC$_2$\ and Softmax Code.} 
Our main result for \GNC$_2$ is the following.

\begin{theorem}[\GNC$_2$]\label{thm:NC2} 
Let $(\mW^\star,\mH^\star)$ be an optimal solution to  \eqref{def:hardmax}. Then, it holds that $\mW^\star$ is a Softmax Code, \ie,
\begin{equation}
    \mW^\star = \argmax\limits_{\mW\in\OB(d,K)} \rho_\onevsrest(\mW).
\end{equation}
\end{theorem}
\GNC$_2$ is described by the Softmax Code, which is defined from an optimization problem (see \Cref{def:softmax-code}).
This optimization problem may not have a closed form solution in general. 
Nonetheless, the one-vs-rest distance that is used to define Softmax Code has a clear geometric meaning, making an intuitive interpretation of Softmax Code tractable.
Specifically, maximizing the one-vs-rest distance results in the classifier weight vectors $\{\vw_k^\star\}$ to be maximally distant. 
As shown in \Cref{fig:uniform_convexhull_margin,fig:nonuniform_convexhull_margin} for a simple setting of four classes in a 2D plane,  
the  weight vectors $\{\vw_k\}$  that are uniformly distributed (and hence maximally distant) have a larger margin than the non-uniform case. 

For certain choices of $(d, K)$ the Softmax Code bears a simple form.

\begin{theorem}
\label{thm:special-case-NC3} 
For any positive integers $K$ and $d$, let $\mW^\star \in \OB(d,K)$ be a Softmax Code. 
Then, 
\begin{itemize}[leftmargin=0.22in,topsep=0.0em,itemsep=-0.4em]
    \item $d=2$: $\{\vw_k^\star\}$ is uniformly distributed on the unit circle, i.e., $\{\vw_k^\star\} = \{\big(\cos(\frac{2\pi k}{K} + \alpha), \sin(\frac{2\pi k}{K} + \alpha)\big)\}$ for some $\alpha$;
    \item $K\le d+1$: $\{\vw_k^\star\}$ forms a simplex ETF, i.e., $\W^\star = \sqrt{\frac{K}{K-1}}\P(\I_K - \frac{1}{K}\1_K \1_K^\top)$ for some orthonomal $\P \in \RR^{d\times K}$;
    \item $d+1< K\le 2d $: $\rho_\onevsrest(\mW^\star) =1$ which can be achieved when $\{\vw_k^\star\}$ are a subset of vertices of a cross-polytope; 
\end{itemize}
\end{theorem}
For the cases of $K \le d+1$, the optimal $\mW^\star$ from \Cref{thm:special-case-NC3} is the same as that of \cite{lu2022neural}. 
However, \Cref{thm:special-case-NC3} is an analysis of the HardMax loss while \cite{lu2022neural} analyzed the CE loss. 

\paragraph{\GNC$_1$\ and Within-class Variability Collapse.} 
To establish the within-class variability collapse property, we require a technical condition associated with the Softmax Code.
Recall that Softmax Codes are those that maximize the \emph{minimum} one-vs-rest distance over all classes. 
We introduce \emph{rattlers}, which are classes that do not attain such a \emph{minimum}.
\begin{definition}[Rattler of Softmax Code]
    Given positive integers $d$ and $K$, a rattler associated with a Softmax Code $\mW^\text{SC} \in \OB(d,K)$ is an index $k_\text{rattler} \in [K]$ for which
    \begin{equation}
        \min_{k \in [K]} \dist(\vw_k^\text{SC},\{\vw_j^\text{SC}\}_{j\in[K]\setminus k}) \ne \dist(\vw^\text{SC}_{k_\text{rattler}},\{\vw^\text{SC}_j\}_{j\in[K]\setminus k_\text{rattler}}).
    \end{equation}
\end{definition}
In other words, rattlers are points in a Softmax Code with no neighbors at the minimum one-to-rest distance.
This notion is borrowed from the literature of the \emph{Tammes Problem} \citep{cohn2022small,wang2009finding}, which we will soon discuss in more detail\footnote{The occurrence of rattlers is rare: Among the $182$ pairs of $(d, K)$ for which the solution to Tammes problem is known, only $31$ have rattlers \citep{cohn2022small}. This has excluded the cases of $d=2$ or $K \le 2d$ where there is no rattler. 
The occurrence of ratter in Softmax Code may be rare as well.}.

We are now ready to present the main results for \GNC$_1$.

\begin{theorem}[\GNC$_1$]\label{thm:NC1} 
Let $(\mW^\star,\mH^\star)$ be an optimal solution to  \eqref{def:hardmax}. For all $k$ that is not a rattler of $\mW^\star$, it holds that
\begin{equation}
    \overline{\vh}_{k}^\star \doteq \vh_{k,1}^\star=\cdots=\vh_{k,n}^\star = \mathcal{P}_{\setS^{d-1}} \left(\vw_k^\star - \mathcal{P}_{ \{\vw_j^\star\}_{j\in[K]\setminus k}}(\vw_k^\star) \right),
\end{equation}
where $\mathcal{P}_\mathcal{W}(\vv) \doteq \argmin_{\vw \in \conv(\mathcal{W})}\{\|\vv - \vw\|_2 \}$ denotes the projection of $\vv$ on $\conv(\mathcal{W})$.
\end{theorem}

The following result shows that the requirement in the \Cref{thm:NC1} that $k$ is not a rattler is satisfied in many cases. 
\begin{theorem}
    If $d = 2$, or $K \le d+1$,  Softmax Code has no rattler for all classes. 
\end{theorem}

\paragraph{\GNC$_3$\ and Self-Duality.} 
To motivate our technical conditions for establishing self-duality, assume that any optimal solution $(\mW^\star, \mH^\star)$ to \eqref{def:hardmax} satisfies self-duality as well as \GNC$_1$. 
This implies that 
\begin{equation}
    \argmin_{\mW\in\OB(d,K),\mH\in\OB(d,nK)}  \mathcal{L}_{\text{HardMax}}(\mW, \mH) 
    = \argmin_{\mW \in\OB(d,nK)} \max_{k \in [K]} \max_{i \in [n]} \max_{k' \ne k} \langle \w_{k'} - \w_{k} , \w_{k} \rangle.
\end{equation}
After simplification we may rewrite the optimization problem on the right hand side equivalently as:
\begin{equation}\label{eq:tammes-problem}
     \max_{\mW\in\OB(d,K)} \rho_\onevsone(\mW), ~~\text{where}~~\rho_\onevsone(\mW) \doteq \min_{k \in [K]} \min_{k' \ne k} \dist(\vw_k,\vw_{k'}).
\end{equation}
Eq.~\eqref{eq:tammes-problem} is the well-known \emph{Tammes problem}.
Geometrically, the problem asks for a distribution of $K$ points on the unit sphere of $\RR^d$ so that the minimum distance between any pair of points is maximized. 
The Tammes problem is unsolved in general, except for certain pairs of $(K, d)$.

Both the Tammes problem and the Softmax Code are problems of arranging points to be maximally separated on the unit sphere, with their difference being the specific measures of separation.
Comparing \eqref{eq:tammes-problem} and \eqref{eq:softmax-code}, the Tammes problem maximizes for all $k \in [K]$ the \emph{one-vs-one distance}, i.e., $\min_{k' \ne k} \dist(\vw_k,\vw_{k'})$, whereas the Softmax Code maximizes the minimum \emph{one-vs-rest distance}, i.e., $\dist(\vw_k, \{\vw_j\}_{j\in[K]\setminus k})$. 
Both one-vs-one distance and one-vs-rest distances characterize the separation of the weight vector $\vw_k$ from $\{\vw_j\}_{j\in[K]\setminus k}$. 
As illustrated in Figure \ref{fig:comparison-tammes}, taking $k=1$, the former is the distance between $\vw_1$ and its closest point in the set $\{\vw_2, \vw_3, \vw_4\}$, in this case $\vw_2$ (see \Cref{fig:tammes-margin}), whereas the later captures the minimal distance from $\vw_1$ to the convex hull of the rest vectors $\{\vw_1, \vw_2, \vw_3\}$ (see \Cref{fig:nonuniform_convexhull_margin}). 

Since the Tammes problem can be derived from the self-duality constraint on the HardMax problem, it may not be surprising that the Tammes problem can be used to describe a condition for establishing self-duality.
Specifically, we have the following result. 

\begin{theorem}[\GNC$_3$]\label{thm:NC3}
For any $K, d$ such that both Tammes problem and Softmax Code have no rattler, the following two statements are equivalent:
\begin{itemize}[leftmargin=0.22in,topsep=0.0em,itemsep=-0.3em]
    \item Any optimal solution $(\mW^\star,\mH^\star)$ to  \eqref{def:hardmax} satisfies $\vh_{k, i}^\star = \vw_k^*, \forall i \in [n], \forall k \in [K]$;
    \item The Tammes problem and the Softmax codes are equivalent, \ie,
    \begin{equation}
        \argmax_{\mW\in\OB(d,K)} \rho_\onevsrest(\mW) = \argmax_{\mW\in\OB(d,K)} \rho_\onevsone(\mW).
    \end{equation}
\end{itemize}

\end{theorem}

In words, \Cref{thm:NC3} states that \GNC$_3$ holds if and only if the Tammes problem in \eqref{eq:tammes-problem} and the Softmax codes are equivalent. 
As both the Tammes problem and Softmax Code maximize separation between one vector and the others, though their notions of separation are different, we conjecture that they are equivalent and share the same optimal solutions. We prove this conjecture for some special cases and leave the study for the general case as future work. 

\begin{theorem}
    If $d = 2$, or $K \le d+1$, the Tammes problem and the Softmax codes are equivalent. 
\label{thm:Tammes-equivalent-Softmax}\end{theorem}

\subsection{Insights for Choosing Feature Dimension $d$ Given Class Number $K$}

Given a class number $K$, how does the choice of feature dimension $d$ affect the model performance? Intuitively, smaller $d$ reduces the separability between classes in a Softmax Code.
We define this rigorously by providing bounds for the one-vs-rest distance of a Softmax Code based on $d$ and $K$. 
\begin{theorem}\label{thm:softmax-code-objective-bound} 
Assuming $K \geq \sqrt{2 \pi \sqrt{e} d}$ and letting $\Gamma(\cdot)$ denote the Gamma function, we have
\begin{align}
    \frac{1}{2} \biggr[\frac{\sqrt{\pi}}{K}\frac{\Gamma\paren{\frac{d+1}{2}}}{\Gamma\paren{\frac{d}{2} +1}} \biggr]^{\frac{2}{d-1}} \le \max_{\mW \in \OB(d,K)}\rho_\onevsrest(\mW) \le 2 \biggr[ \frac{2\sqrt{\pi}}{K} \frac{\Gamma \paren{\frac{d+1}{2}}}{\Gamma\paren{\frac{d}{2}}} \biggr]^{\frac{1}{d-1}}.
\end{align}\label{eq:margin-bound}
\vspace{-.12in}
\end{theorem}

\begin{wrapfigure}{r}{0.4\textwidth}
\vspace{-1em}
\includegraphics[width=0.4 \textwidth]{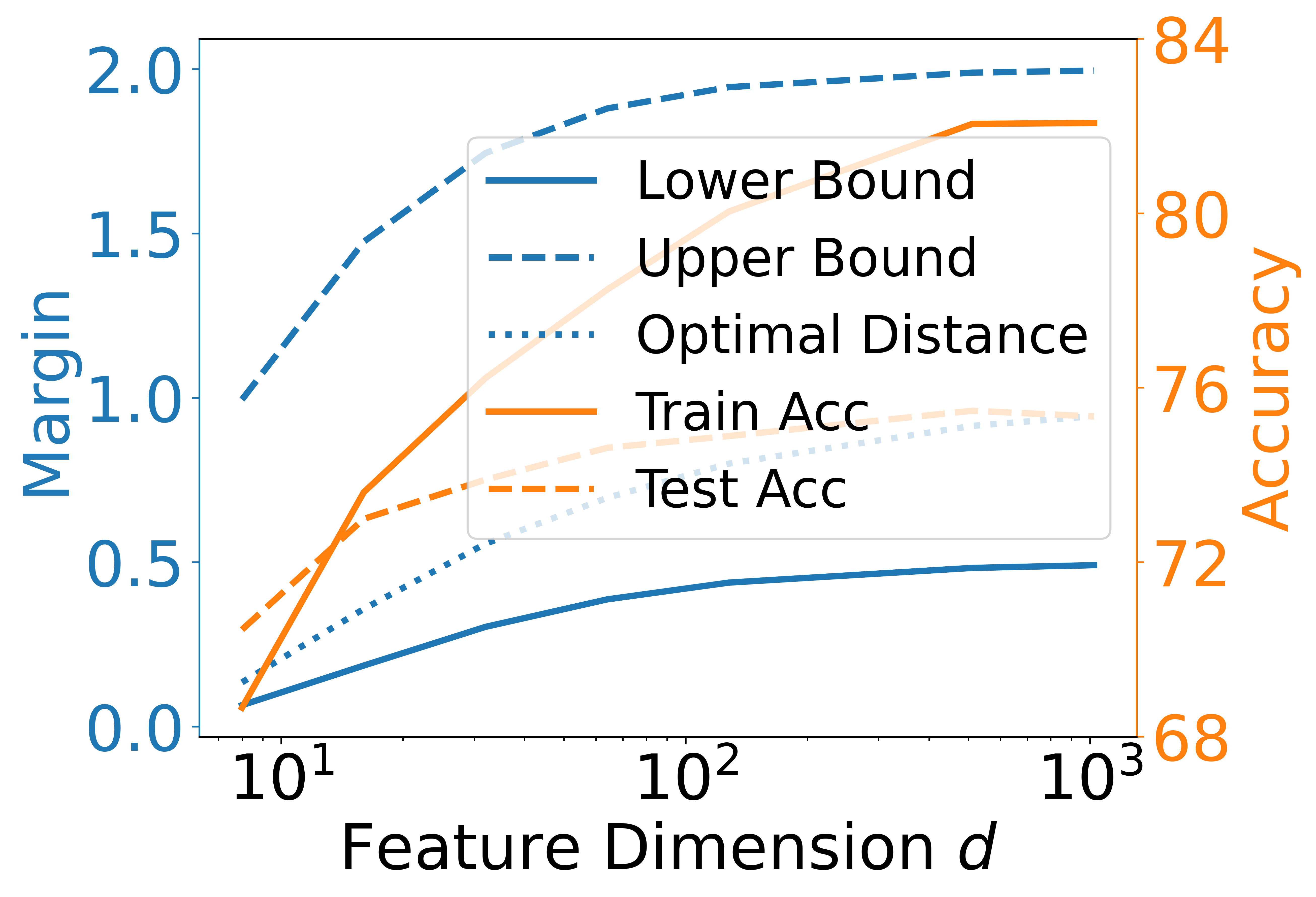}
\centering
\caption{Effect of feature dimension $d$ on (Left $y$-axis): $\rho_\onevsrest(\mW^\star)$ and its upper/lower bounds (in \Cref{thm:softmax-code-objective-bound}), and (Right $y$-axis): training and test accuracies for ResNet-50 on ImageNet.}
\label{fig:ImageNet-vary-d}
\vspace{-2em}
\end{wrapfigure}


The bounds characterize the separability for $K$ classes in $d$-dimensional space. Given the number of classes $K$ and desired margin $\rho$, the minimal feature dimension is roughly an order of $\log(K^2 / \rho)$, showing classes separate easily in higher dimensions. This also provides a justification for applications like face classification and self-supervised learning, where the number of classes (e.g., millions of classes) could be significantly larger than the dimensionality of the features (e.g., $d = 512$).

By conducting experiments on ResNet-50 with varying feature dimensions for ImageNet classification, we further corroborate the relationship between feature dimension and network performance in \Cref{fig:ImageNet-vary-d}. First, we observe that the curve of the optimal distance is closely aligned with the curve of testing performance, indicating a strong correlation between distance and testing accuracy. Moreover, both the distance and performance curves exhibit a slow (exponential) decrease as the feature dimension $d$ decreases, which is consistent with the bounds in  \Cref{eq:margin-bound}.

\section{The Assignment Problem:  An Empirical Study 
}\label{subsec:permutation}
\begin{figure} [!ht]
     \centering
     \begin{subfigure}[b]{0.21\textwidth}
         \centering
         \includegraphics[width=\textwidth]{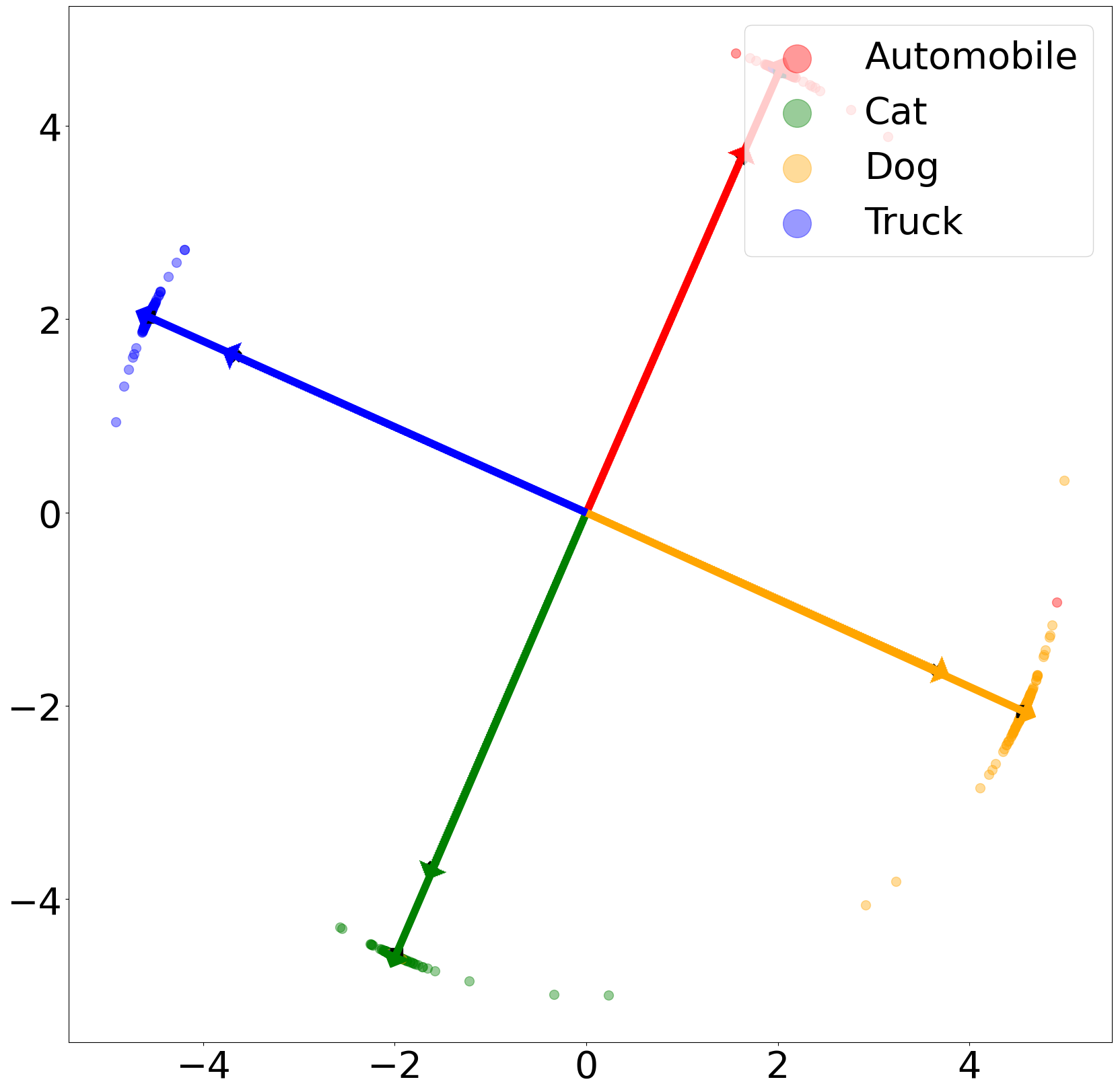}
         \caption{Trainable ($91.45\%$)}
         \label{fig:learned-assignment}
     \end{subfigure}
     \hfill
     \begin{subfigure}[b]{0.21\textwidth}
         \centering
         \includegraphics[width=\textwidth]{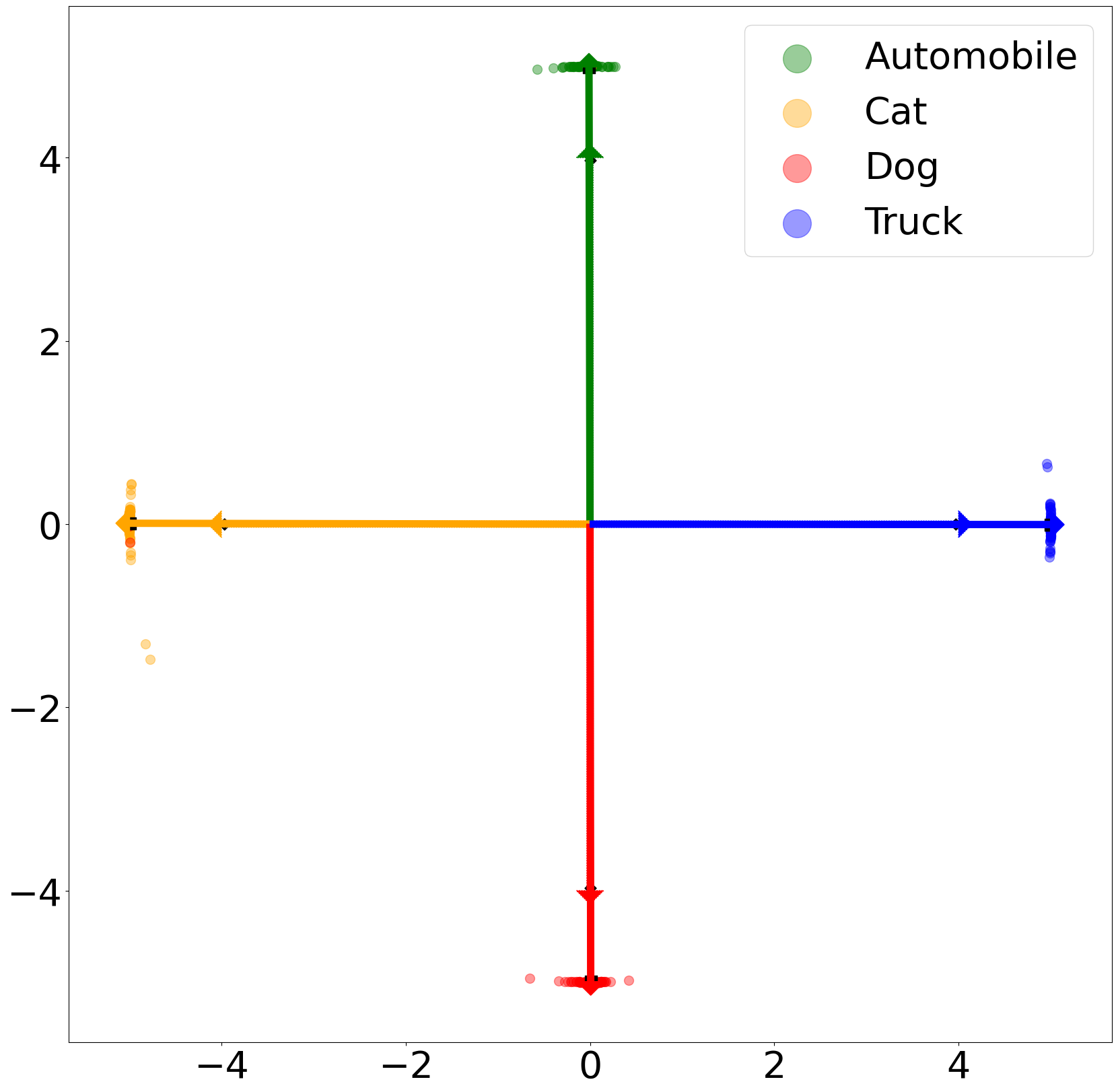}
         \caption{Fixed A1 ($91.90\%$)}
         \label{fig:assignment_1}
     \end{subfigure}
     \hfill
     \begin{subfigure}[b]{0.21\textwidth}
         \centering
         \includegraphics[width=\textwidth]{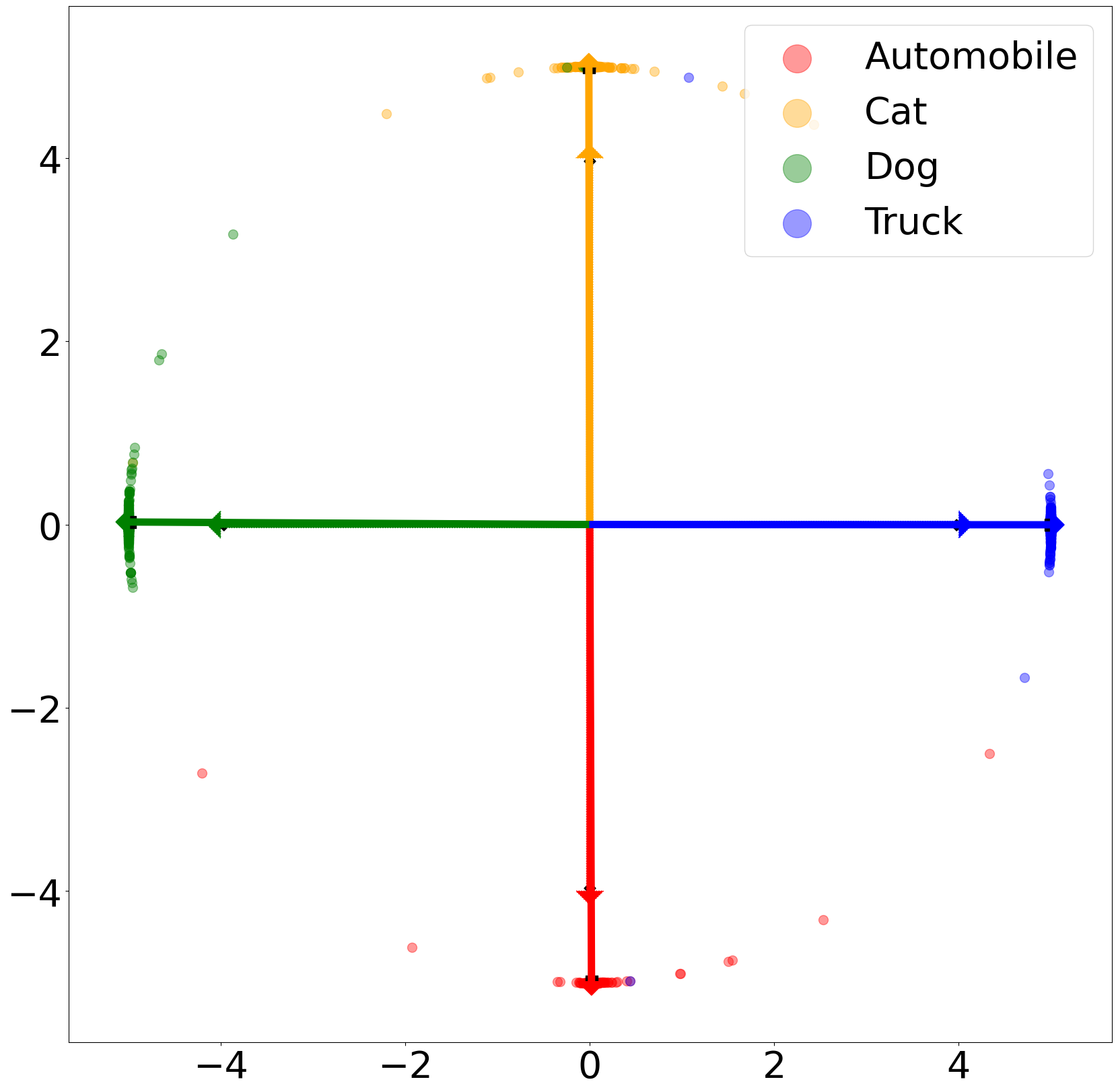}
         \caption{Fixed A2 ($92.13\%$)}
         \label{fig:assignment_3}
    \end{subfigure}
    \hfill
     \begin{subfigure}[b]{0.21\textwidth}
         \centering
         \includegraphics[width=\textwidth]{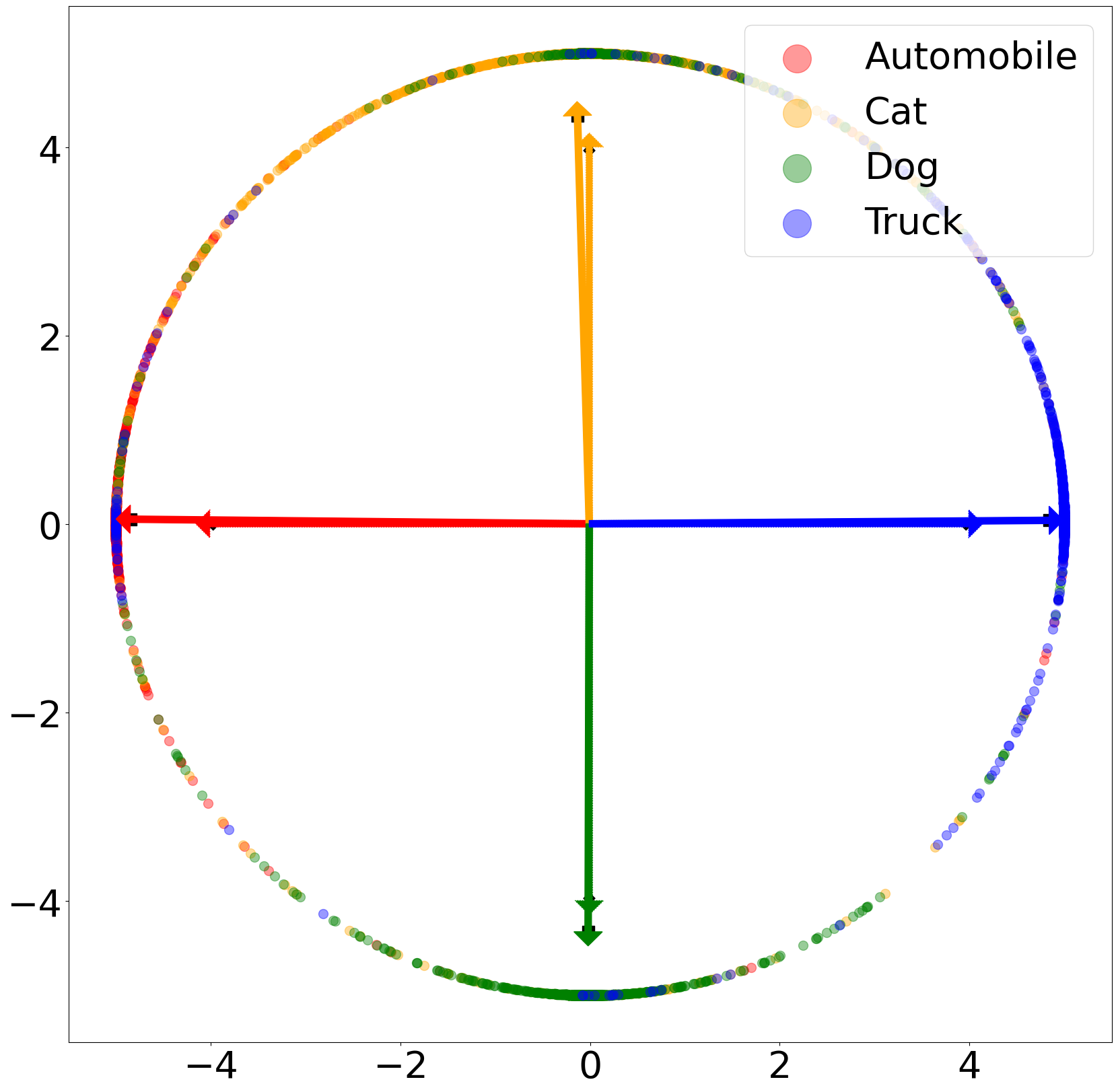}
         \caption{Fixed A3 ($89.95\%$)}
         \label{fig:assignment_2}
     \end{subfigure}\
        \caption{\textbf{Assignment of classes to classifier weights} for a ResNet18 with 2-dimensional feature space trained on the 4 classes \{Automobile, Cat, Dog, Truck\} from CIFAR10. \emph{(a)} Learned classifier. \emph{(b-d)} Classifiers fixed to be three different assignments. Test accuracy is reported in the bracket.} 
        \label{fig:permutation}
        \vspace{-.3in}
\end{figure}

Unlike the case $d\ge K-1$ where the optimal classifier (simplex ETF) has equal angles between any pair of the classifier weights, when $d<K-1$, not all pairs of classifier weights are equally distant 
with the optimal $\mW$ (Softmax Code) predicted in \Cref{thm:NC2}. 
Consequently, this leads to a ``class assignment" problem. 
To illustrate this, we train a ResNet18 network with $d =2$ on four classes \{Automobile, Cat, Dog, Truck\}  from CIFAR10 dataset that are selected due to their clear semantic similarity and discrepancy. 
In this case, according to \Cref{thm:special-case-NC3}, the optimal classifiers are given by $[1, 0], [-1, 0], [0, 1], [0, -1]$, up to a rotation. Consequently, there are three distinct class assignments,
as illustrated in \Cref{fig:assignment_1,fig:assignment_3,fig:assignment_2}. 

When doing standard training, the classifier consistently converges to the case where Cat and Dog are closer together across 5 different trials; \Cref{fig:learned-assignment} shows the learned features (dots) and classifier weights (arrows) in one of such trials.
This demonstrates the implicit algorithmic regularization in training DNNs, which naturally attracts (semantically) similar classes and separates dissimilar ones. 

We also conduct experiments with the classifier fixed to be one of the three arrangements, and present the results in \Cref{fig:assignment_1,fig:assignment_3,fig:assignment_2}.
Among them, we observe that the case where Cat and Dog are far apart achieves a testing accuracy of $89.95 \%$, which is lower than the other two cases with testing accuracies of $91.90 \%$ and $92.13 \%$. 
This demonstrates the important role of class assignment to the generalization of DNNs, and that the implicit bias of the learned classifier is benign, \ie, leads to a more generalizable solutions.

\section{Implications for Practical Network Training/Fine-tuning
}\label{subsec:ocm}

Since the classifier always converges to a simplex ETF when $K \le d+1$, prior work proposes to fix the classifier as a simplex ETF for reducing training cost \citep{zhu2021geometric} and handling imbalance dataset \citep{yang2022we}. 
When $K > d+1$, the optimal classifier is also known to be a Softmax Code according to \GNC$_2$. 
However, the same method as in prior work may become sub-optimal due to the class assignment problem (see \Cref{subsec:permutation}). 
To address this, we introduce the method of class-mean features (CMF) classifiers, where the classifier weights are set to be the exponential moving average of the mini-batch class-mean features during the training process. 
This approach is motivated from \GNC$_3$ which states that the optimal classifier converges to the class-mean features.
We explain the detail of CMF in \Cref{app:exp}.
As in prior work, CMF can reduce trainable parameters as well. For instance, it can reduce $30.91$\% of total parameters in a ResNet18 for BUPT-CBFace-50 dataset \citep{zhang2020class}.
Here, we compare CMF with the standard training where the classifier is learned together with the feature mapping, in both training from scratch and fine-tuning.

\begin{figure} [!ht]
     \centering
     \begin{subfigure}[b]{0.3\textwidth}
         \centering
         \includegraphics[width=\textwidth]{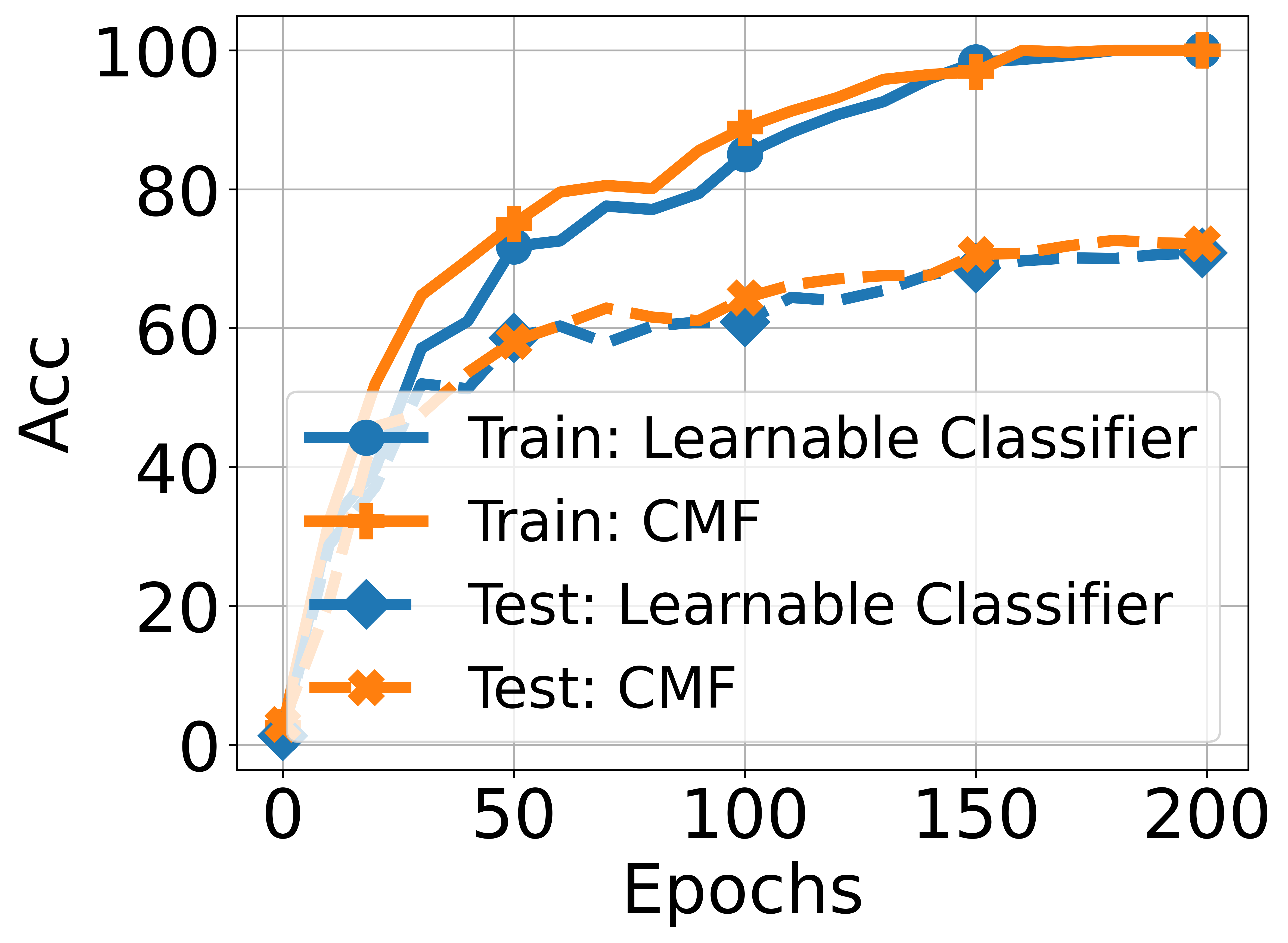}
         \caption{ResNet}
     \end{subfigure}\
     \begin{subfigure}[b]{0.3\textwidth}
         \centering
         \includegraphics[width=\textwidth]{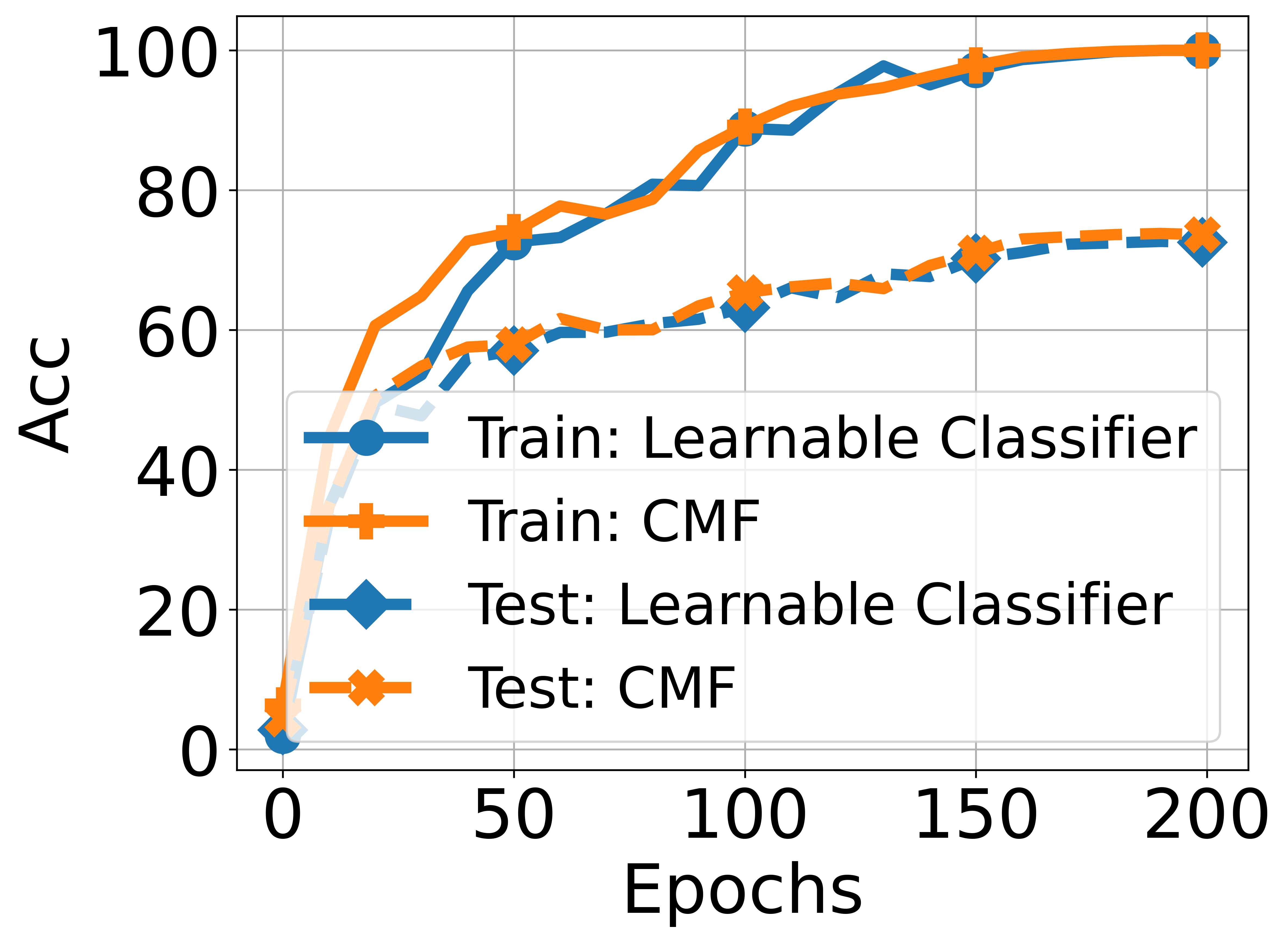}
         \caption{DenseNet}
     \end{subfigure}\
          \begin{subfigure}[b]{0.3\textwidth}
         \centering
         \includegraphics[width=\textwidth]{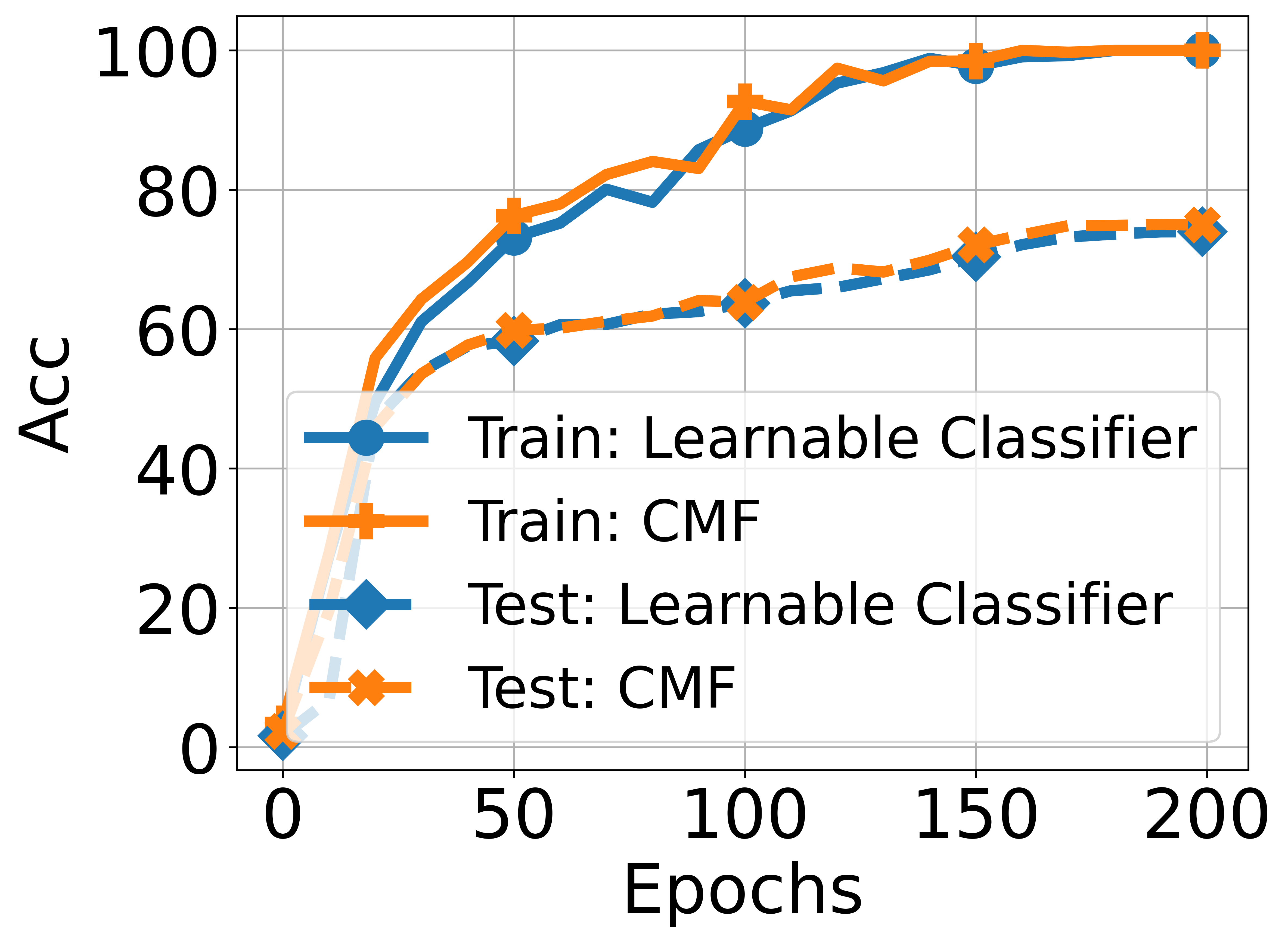}
         \caption{ResNeXt}
    \end{subfigure}
     
        \caption{\textbf{Comparison of the learning curves (training and testing accuracies) with learned classifiers vs. CMF classifiers} trained with ResNet18, DenseNet121, and ResNeXt50 on CIFAR100 dataset and $d = 10$.
        }
        \label{fig:OCM}
        \vspace{-.1in}
\end{figure}

\paragraph{Training from Scratch.} We train a ResNet18 on CIFAR100 by using a learnable classifier or the CMF classifier. 
The learning curves in \Cref{fig:OCM} indicate that the approach with CMF classifier achieves comparable performance to the classical training protocols. 

\paragraph{Fine-tuning.} 
To verify the effectiveness of the CMF classifiers on fine-tuning, we follow the setting in \cite{kumar2022finetuning} to measure the performance of the fine-tuned model on both in-distribution (ID) task (i.e., CIFAR10 \cite{krizhevsky2009learning}) and OOD task (STL10 \cite{pmlr-v15-coates11a}). We compare the standard approach that fine-tunes both the classifier (randomly initialized) and the pre-trained feature mapping with our approach (using the CMF classifier). Our experiments show that the approach with CMF classifier achieves slightly better ID accuracy ($98.00\%$ VS $97.00\%$) and a better OOD performance ($90.67\%$ VS $87.42\%$).
The improvement of OOD performance stems from the ability to align the classifier with the class-means through the entire process, which better preserves the OOD property of the pre-trained model. Our approach also simplifies the two-stage approach of linearly probing and subsequent full fine-tuning in \cite{kumar2022finetuning}.

\section{Conclusion}\label{sec:conclusion}
In this work, we have introduced generalized neural collapse (\GNC)  for characterizing learned last-layer features and classifiers in DNNs under an arbitrary number of classes and feature dimensions. We empirically validate the \GNC\ phenomenon on practical DNNs that are trained with a small temperature in the CE loss and subject to spherical constraints on the features and classifiers. Building upon the unconstrained features model 
we have proven that \GNC\ holds under certain technical conditions. 
\GNC\ could offer valuable insights for the design, training, and generalization of DNNs. For example, the minimal one-vs-rest distance provides implications for designing feature dimensions when dealing with a large number of classes. Additionally, we have leveraged \GNC\ to  enhance training efficiency and fine-tuning performance by fixing the classifier as class-mean features. Further exploration of \GNC\ in other scenarios, such as imbalanced learning, is left for future work. It is also of interest to further study the problem of optimally assigning classifiers from Softmax Code for each class, which could shed light on developing techniques for better classification performance.

\section*{Acknowledgment} 
JJ, JZ, and ZZ acknowledge support from NSF grants CCF-2240708 and IIS-2312840. PW and QQ acknowledge support from NSF CAREER CCF-2143904, NSF CCF-2212066, NSF CCF-2212326, NSF IIS 2312842, ONR N00014-22-1-2529, an AWS AI Award, a gift grant from KLA, and MICDE Catalyst Grant.  We thank
CloudBank (supported by NSF under Award \#1925001) for providing the computational resources.


\bibliography{reference}
\bibliographystyle{iclr2024_conference}

\newpage 
\setcounter{page}{1}
\counterwithin{figure}{section}
\counterwithin{equation}{section}

\appendix
\part{Appendix} 
\parttoc 

\section{Organizations and Basic.} 
The organization of the appendix is as follows. Firstly, we introduce the fundamental theorems that are utilized throughout the appendices. In \Cref{app:exp}, we offer comprehensive information regarding the datasets and computational resources for each figure, along with presenting additional experimental results on practical datasets. Lastly, in \Cref{app:thm}, we provide the theoretical proofs for all the theorems mentioned in \Cref{sec:main_result}. 

\begin{lemma}[LogSumExp]\label{lem:logsumexp} Let denote $\vx=\begin{bmatrix}
    x_1, x_2, \cdots, x_n
\end{bmatrix} \in \mathcal{R}^n$ and the LogSumExp function $\text{LSE}(\vx, \tau)=\log\paren{\sum_{i=1}^n \exp(x_i/\tau)}$. With $\tau\rightarrow 0$, we have
\begin{align*}
    \tau\text{LSE}(\vx, \tau)\rightarrow\max_i x_i
\end{align*}
In other words, the LogSumExp function is a smooth approximation to the maximum function.
\end{lemma}

\begin{proof}[Proof of \Cref{lem:logsumexp}]  According to the definition of the LogSumExp function, we have
    \begin{align*}
        \max_i x_i \leq \tau\text{LSE}(\vx, \tau)\leq  \max_i x_i+\tau\log n
    \end{align*}
    where the first inequality is strict unless $n=1$ and the second inequality is strict unless all arguments are equal $x_1=x_2=\cdots=x_n$. Therefore, with $\tau\rightarrow 0$, $\tau\text{LSE}(\vx, \tau)\rightarrow\max_i x_i$. 
\end{proof}

\begin{theorem}[Carathéodory's theorem \cite{caratheodory1911variabilitatsbereich}] Given $\vw_1,\ldots,\vw_K\in \R^{d}$, if a point $\vv$ lies in the convex hull $\conv(\{\vw_1,\ldots,\vw_K\})$ of $\{\vw_1,\ldots,\vw_K\}$, then $\vv$ resides in the convex hull of at most $d+1$ of the points in $\{\vw_1,\ldots,\vw_K\}$.
\label{thm:Caratheodory}\end{theorem}

\section{Experiment}\label{app:exp}
In this section, we begin by providing additional details regarding the datasets and the computational resources utilized in the paper. Specifically, CIFAR10, CIFAR100, and BUPT-CBFace datasets are publicly available for academic purposes under the MIT license. Additionally, all experiments were conducted on 4xV100 GPU with 32G memory. Furthermore, we present supplementary experiments and implementation specifics for each figure.

\subsection{Implementation details}

We present implementation details for results in the paper.

\paragraph{Results in \Cref{fig:NC-dynamic}.} To illustrate the occurrence of the \GNC\ phenomenon in practical multi-class classification problems, we trained a ResNet18 network \citep{he2016deep} on the CIFAR100 dataset \citep{krizhevsky2009learning} using CE loss with varying temperature parameters. In this experiment, we set the dimensions of the last-layer features to $10$. 
This is achieved by setting the number of channels in the second convolutional layer of the last residual block before the classifier layer to be $10$. 
Prior to training, we applied standard preprocessing techniques, which involved normalizing the images (channel-wise) using their mean and standard deviation. We also employed standard data augmentation methods. For optimization, we utilized SGD with a momentum of $0.9$ and an initial learning rate of $0.1$, which decayed according to the CosineAnnealing over a span of $200$ epochs.

The optimal margin (represented by the dotted line) in the second left subfigure is obtained from numerical optimization of the Softmax Code problem. In this optimization process, we used an initial learning rate of $0.1$ and decreased it by a factor of $10$ every $1000$ iterations, for a total of $5000$ iterations.

\paragraph{Results in \Cref{fig:permutation}.} To demonstrate the implicit algorithmic regularization in training DNNs, we train a ResNet18 network on four classes {Automobile, Cat, Dog, Truck} from CIFAR10 dataset. We set the dimensions of the last-layer features to $2$ so that the learned features can be visualized, and we used a temperature parameter of $0.05$. 
We optimized the networks for a total of $800$ epochs using SGD with a momentum of $0.9$ and an initial learning rate of $0.1$, which is decreased by a factor of $10$ every $200$ epochs.

\paragraph{Results in \Cref{fig:OCM}.}To assess the effectiveness of the proposed CMF (Class Mean Feature) method, we utilized the ResNet18 architecture \citep{he2015deep} as the feature encoder on the CIFAR100 dataset \citep{krizhevsky2009learning}, where we set the feature dimension to $d = 20$ and the temperature parameter to $\tau = 0.1$. Since the model can only access a mini-batch of the dataset for each iteration, it becomes computationally prohibitive to calculate the class-mean features of the entire dataset. As a solution, we updated the classifier by employing the exponential moving average of the feature class mean, represented as $\mW^{(t+1)} \leftarrow \beta \mW^{(t)} + (1-\beta) \overline{\mH}^{(t)}$. Here, $\overline{\mH}^{(t)} \in \mathcal{R}^{K \times d}$ denotes the class mean feature of the mini-batch in iteration $t$, $\mW \in \mathcal{R}^{K \times d}$ represents the classifier weights, and $\beta \in [0,1)$ denotes the momentum coefficient (in our experiment, $\beta = 0.9$). We applied the same data preprocessing and augmentation techniques mentioned previously and 
 employed an initial learning rate of $0.1$ with the CosineAnnealing scheduler.

\paragraph{Results in the fine-tuning experiment of \Cref{subsec:ocm}.} We fine-tune a pretrained ResNet50 on MoCo v2 \citep{chen2020improved} on CIFAR10 \citep{krizhevsky2009learning}. The CIFAR10 test dataset was chosen as the in-distribution (ID) task, while the STL10 dataset \citep{pmlr-v15-coates11a} served as the out-of-distribution (OOD) task. To preprocess the dataset, we resized the images to $224 \times 224$ using BICUBIC interpolation and normalized them by their mean and standard deviation. Since there is no “monkey” class in CIFAR10 dataset, we remove the “monkey” class in the STL10 dataset. Additionally, we reassign the labels of CIFAR10 datasets to the STL10 dataset in order to match the classifier output. For optimization, we employed the Adam optimizer with a learning rate of $1e-5$ and utilized the CosineAnnealing scheduler. The models are fine-tuned for 5 epochs with the batch size of 100. We reported the best ID testing accuracy achieved during the fine-tuning process and also provided the corresponding OOD accuracy.

\paragraph{Results in \Cref{fig:intro_d<K}.} To visualize the different structures learned under weight decay or spherical constraint, we conducted experiments in both practical and unconstrained feature model settings. In the practical setting, we trained a ResNet18 network \citep{he2016deep} on the first $30$ classes of the CIFAR100 dataset \citep{krizhevsky2009learning} using either weight decay or spherical constraint with the cross-entropy (CE) loss. For visualization purposes, we set the dimensions of the last-layer features to $2$, and for the spherical constraint, we used a temperature parameter of $70$. Prior to training, we applied standard preprocessing techniques, which included normalizing the images (channel-wise) using their mean and standard deviation. Additionally, we employed standard data augmentation methods. We optimized the networks for a total of 1600 epochs using SGD with a momentum of $0.9$ and an initial learning rate of 0.1, which decreased by a factor of $10$ every $700$ epochs. For the unconstrained feature model setting, we considered only one sample per class and treated the feature (class-mean features) and weights as free optimization variables. In this case, we initialized the learning rate at $0.1$ and decreased it by a factor of $10$ every $1000$ iterations, for totaling $5000$ iterations. The temperature parameter was set to $0.02$ for the spherical constraint.

\paragraph{Results in \Cref{fig:permutation6classes}.} To demonstrate the implicit algorithmic regularization in training DNNs, we train a ResNet18 network on four classes {Automobile, Cat, Deer, Dog, Horse, Truck} from CIFAR10 dataset. We set the dimensions of the last-layer features to $2$ so that the learned features can be visualized, and we used a temperature parameter of $0.2$. 
We optimized the networks for a total of $600$ epochs using SGD with a momentum of $0.9$ and an initial learning rate of $0.1$, which is decreased by a factor of $10$ every $200$ epochs.

\subsection{Effect of regularization: Spherical constraint vs. weight decay}

\begin{figure} [!ht]
     \centering
     \begin{subfigure}[b]{0.24\textwidth}
         \centering
         \includegraphics[width=\textwidth]{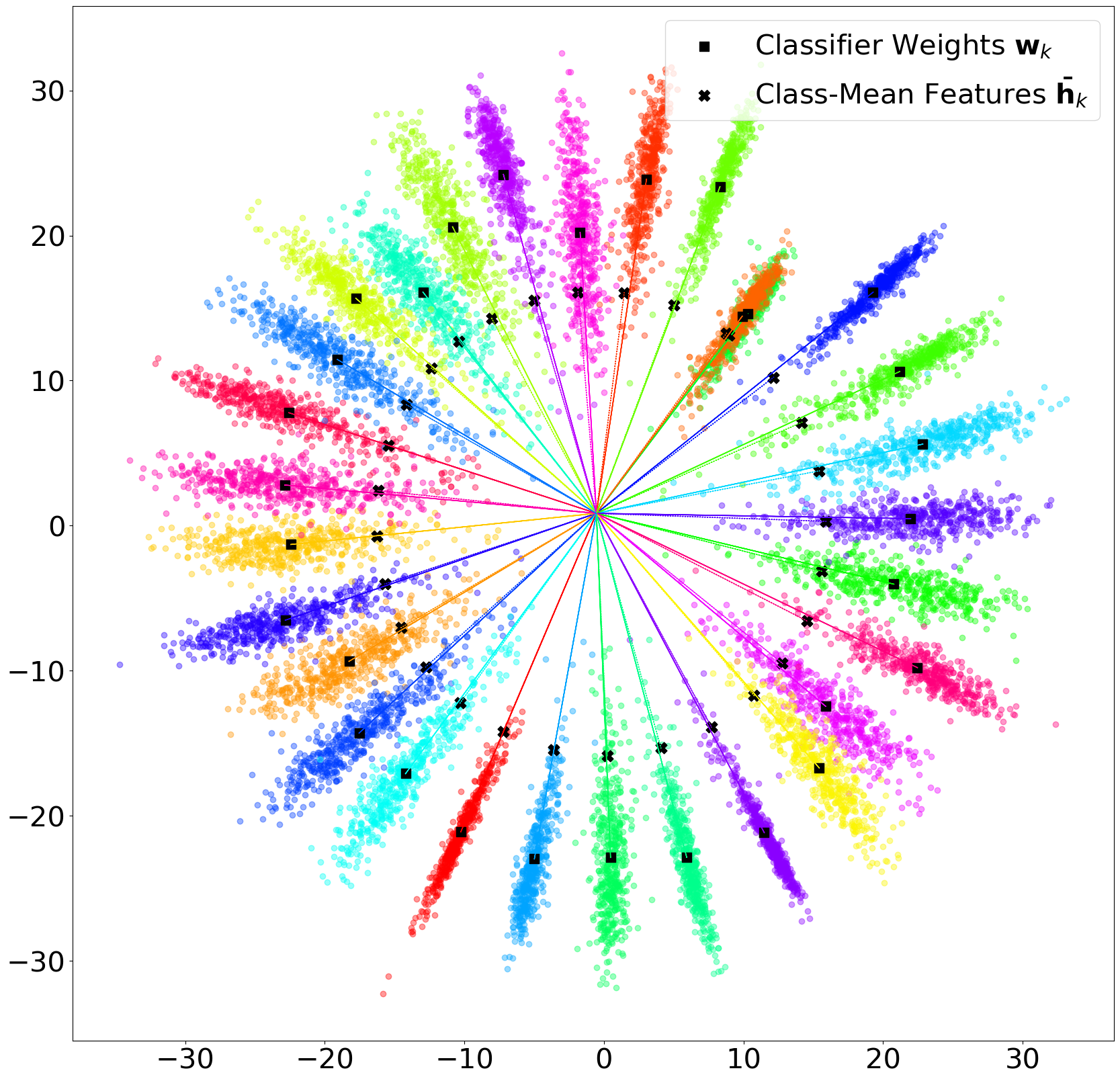}
         \caption{CIFAR100 (WD)}
         \label{fig:cifar100_wd_dist}
     \end{subfigure}\
         \begin{subfigure}[b]{0.24\textwidth}
         \centering
         \includegraphics[width=\textwidth]{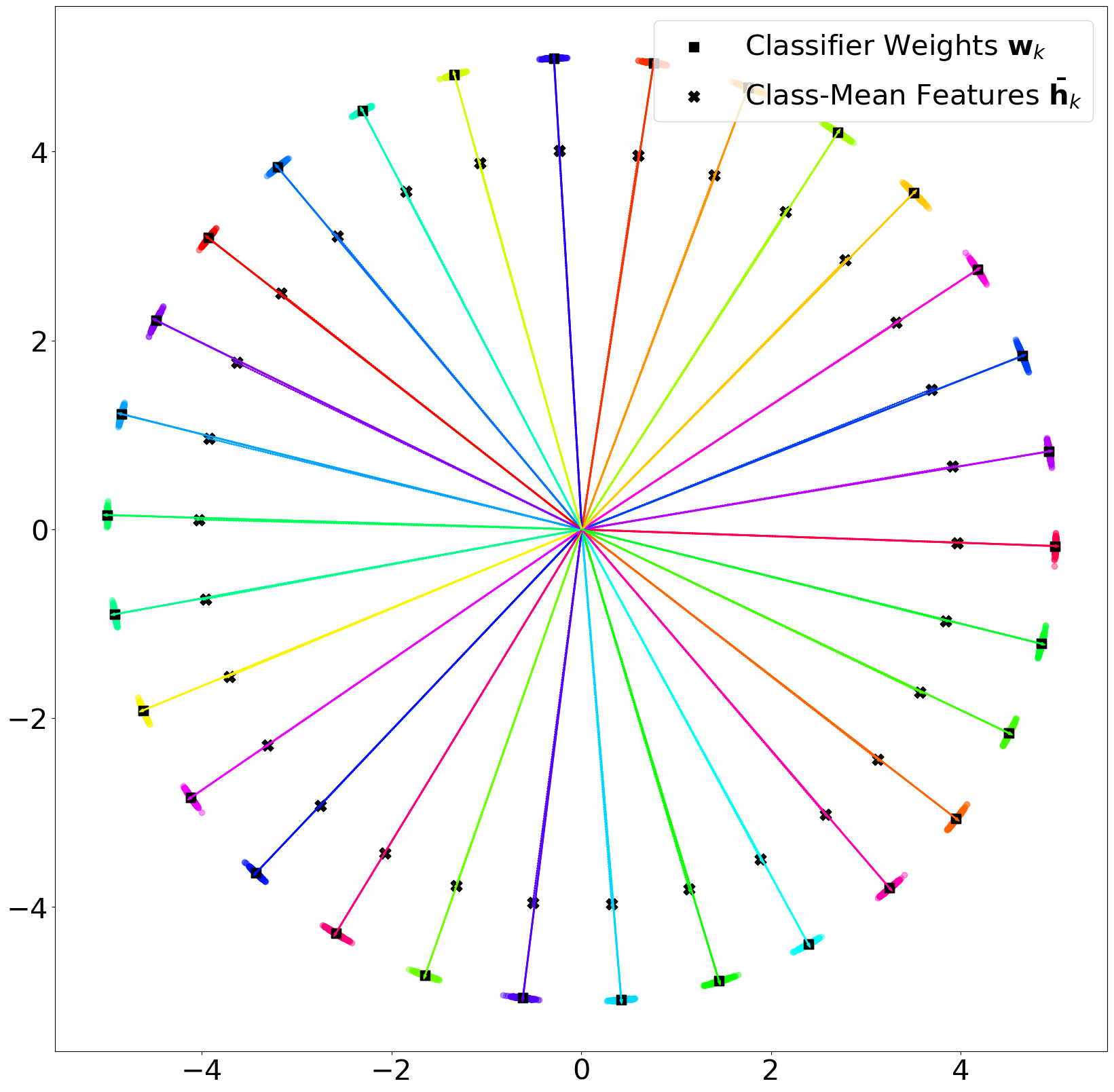}
         \caption{CIFAR100 (SC)}
         \label{fig:ufm_sphere_config}
     \end{subfigure}\
     \begin{subfigure}[b]{0.25\textwidth}
         \centering
         \includegraphics[width=\textwidth]{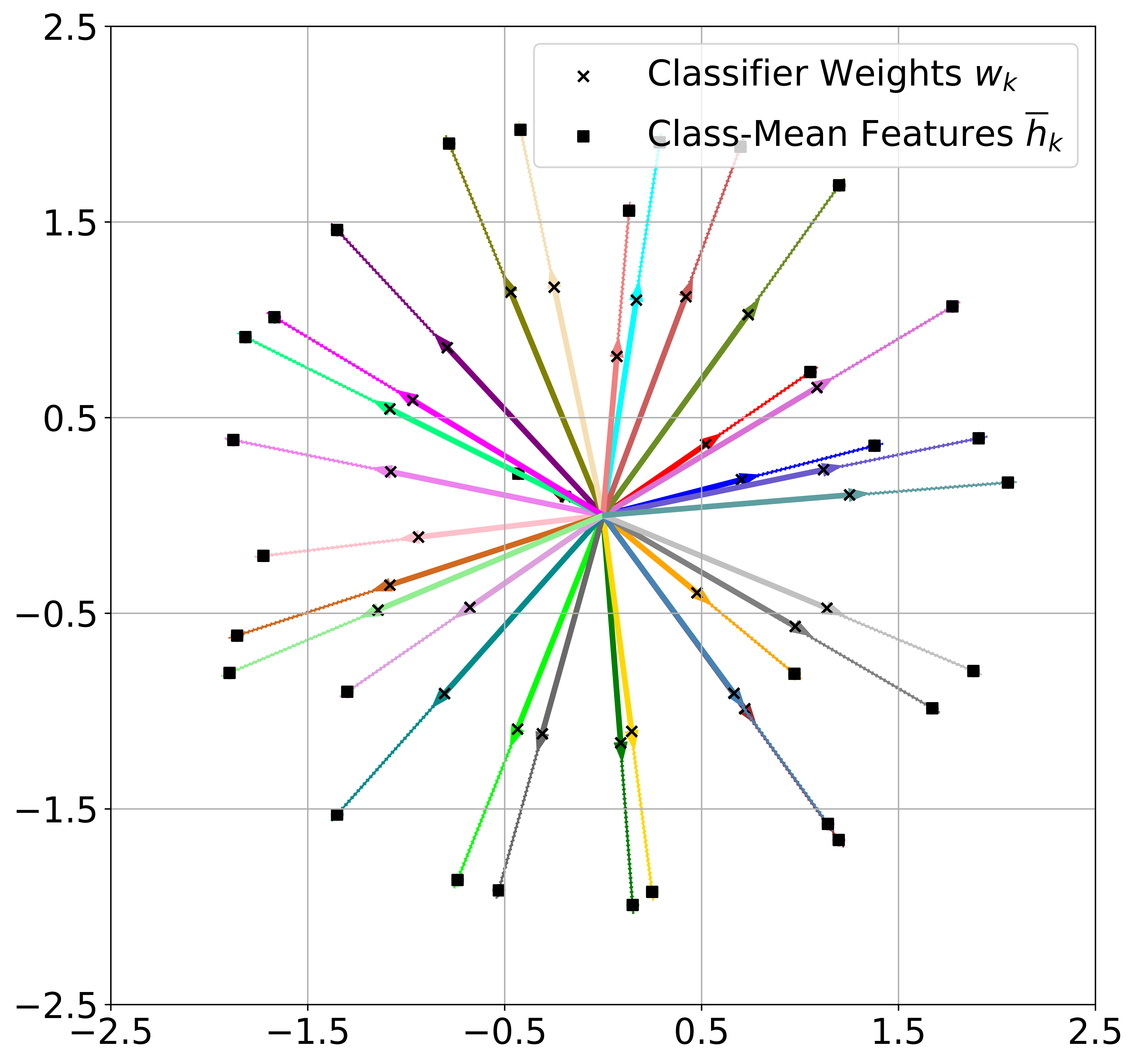}
         \caption{UFM (WD)}
         \label{fig:ufm_wd_dist}
     \end{subfigure}\
     \begin{subfigure}[b]{0.25\textwidth}
         \centering
         \includegraphics[width=\textwidth]{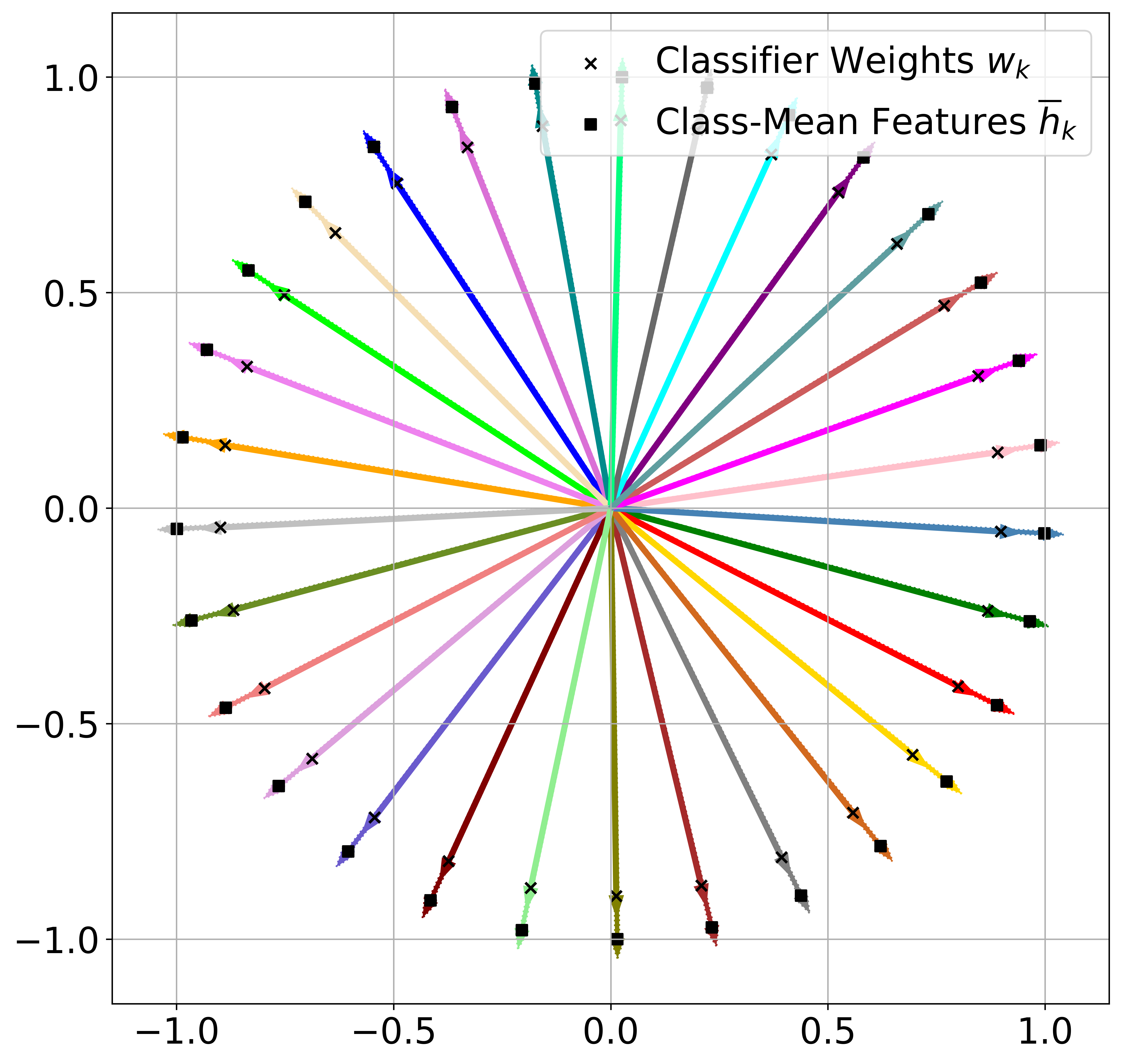}
         \caption{UFM (SC)}
         \label{fig:ufm_sphere_dist}
    \end{subfigure}
        \caption{\textbf{Comparison of classifier weights and last-layer features with weight decay (WD) vs spherical constraint (SC) on features and classifiers.} (a) vs (b): Results with a ResNet18 trained on CFIAR100. (c) vs (d): Results with the unconstrained feature model in \eqref{eqn:ufm-wd} and \eqref{eqn:ufm-sphere}, respectively. In all settings, we set $K=30$ and $d=2$ for visualization; the first $K=30$ classes from CIFAR100 dataset are used to train ResNet18.}
        \label{fig:intro_d<K}
        \vspace{-0.1in}
\end{figure}

\begin{figure}[!ht]
\includegraphics[width=0.5 \textwidth]{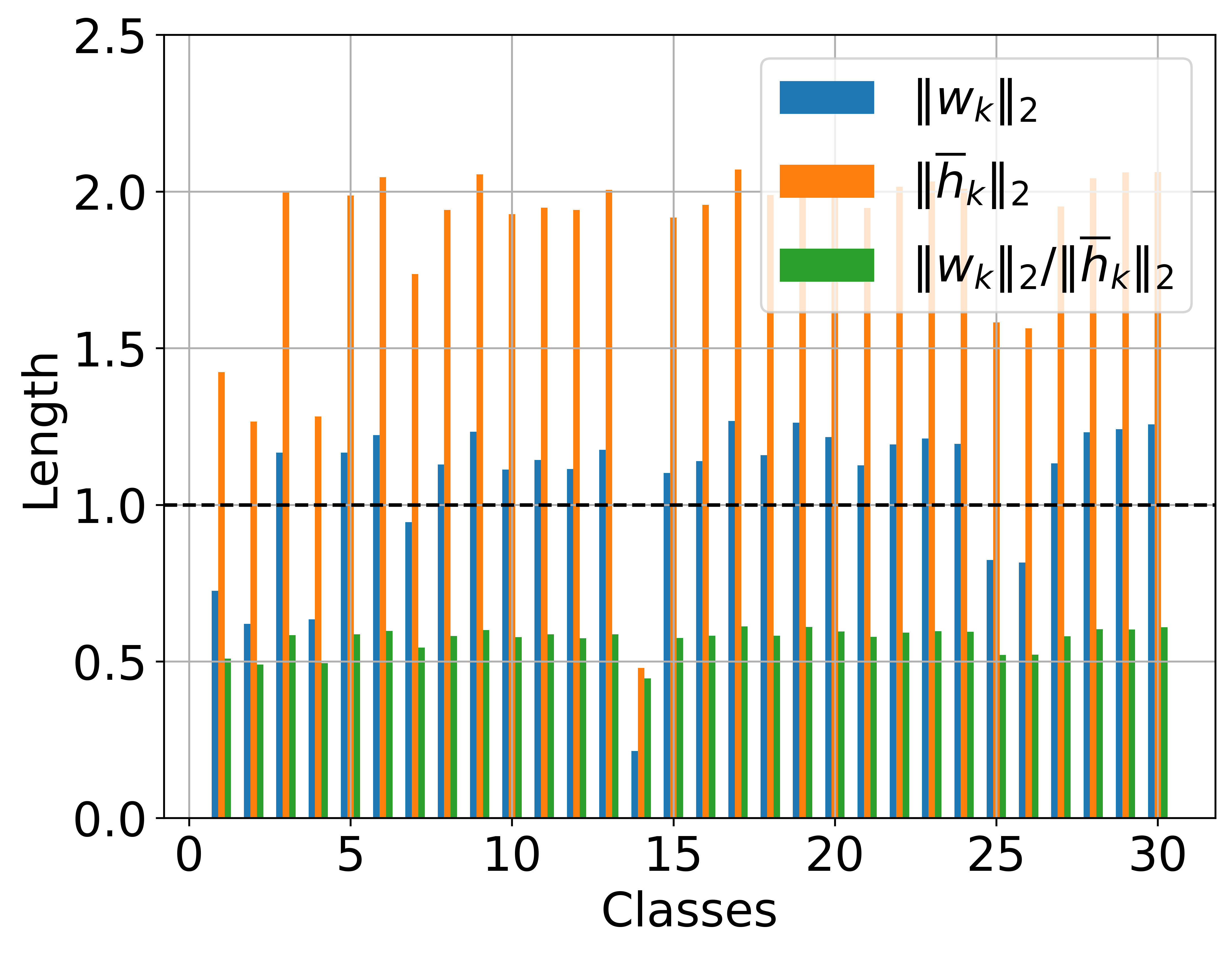}
\centering
\caption{Illustration of the length of the classifier weights and class-mean features in unconstrained feature model (UFM) with weight decay (WD) form in Figure \ref{fig:ufm_wd_dist}. The ratios between the lengths of classifier weights and class-mean features vary across different classes. }
\label{fig:ratios-wd}
\end{figure}

We study the features and classifiers of networks trained with weight decay for the case $K>d+1$, and compare with those obtained with spherical constraint. For the purpose of visualization, we train a ResNet18 with $d = 2$ on the first $K = 30$ classes of CIFAR100 and display the learned features and classifiers in \Cref{fig:intro_d<K}(a). 
Below we summarize observations from \Cref{fig:intro_d<K}(a). 
\begin{itemize}
[leftmargin=0.12in,topsep=0.2em,itemsep=0.01em]
\item \textbf{Non-equal length and non-uniform distribution.} The vectors of class-mean features $\{\ol \vh_k\}$ \emph{do not} have equal length, and also appear to be \emph{not} equally spaced when normalized to the unit sphere. 
\item \textbf{Self-duality only in direction.} The classifiers point towards the same direction as their corresponding class-mean features, but they have different lengths, and there exists no global scaling to exactly align them. As shown in \Cref{fig:ratios-wd}, the ratios between the lengths of classifier weights and class-mean features vary across different classes. Therefore, there exists no global scaling to align them exactly.
\end{itemize}
{In contrast, using spherical constraint produces equal-length and uniformly distributed features, as well as aligned classifier and classifier weights not only in direction but also in length, see \Cref{fig:intro_d<K} (b). Such a result is aligned with \GNC. }
This observation may explain the common practice of using feature normalization in applications with an extremely large number of classes \citep{chen2021exploring,wang2018cosface}, and justify our study of networks trained with sphere constraints in \eqref{eqn:ufm-sphere} rather than weight decay as in \cite{liu2023generalizing}.

We note that the discrepancy between the approaches using weight decay and spherical constraints is not due to the insufficient expressiveness of the networks. In fact, we also observe different performances in the unconstrained feature model with spherical constraints as in \eqref{eqn:ufm-sphere} and with the following regularized form
\citep{zhu2021geometric,mixon2020neural, tirer2022extended, zhou2022optimization}:
\begin{equation}\label{eqn:ufm-wd}
    \min_{\mW, \mH} \frac{1}{nK}\sum_{k=1}^{K} \sum_{i=1}^{n} \mathcal{L}_{\text{CE}}\paren{\mW^\top\vh_{k,i} , \vy_k,\tau} + \frac{\lambda}{2}(\norm{\mW}{F}^2 + \norm{\mH}{F}^2), 
\end{equation}
where $\lambda$ represents the weight decay parameters. We observe similar phenomena in \Cref{fig:intro_d<K}(c, d) as in \Cref{fig:intro_d<K}(a, b) that the weight decay formulation results in features with non-equal lengths, non-uniform distribution, and different lengths than the classifiers. In the next section, we provide a theoretical justification for \GNC\ under the UFM for Problem \eqref{eqn:ufm-sphere}.

\subsection{Additional results on prevalence of \GNC}

We provide additional evidence on the occurrence of the \GNC\ phenomenon in practical multi-class classification problems.
Towards that, we train ResNet18, DenseNet121, and ResNeXt50 network on the CIFAR100, Tiny-ImageNet and BUPT-CBFace-50 datasets using CE loss.
To illustrate the case where the feature dimension is smaller than the number of classes, we insert another linear layer before the last-layer classifier and set the dimensions of the features as $d = 10$ for CIFAR100 and Tiny-ImageNet, and $d = 512$ for BUPT-CBFace-50. 

\begin{figure} [!t]
     \centering
     \subfloat[$\mathcal{GNC}_1$ (CIFAR100)]{\includegraphics[width=0.24\textwidth]{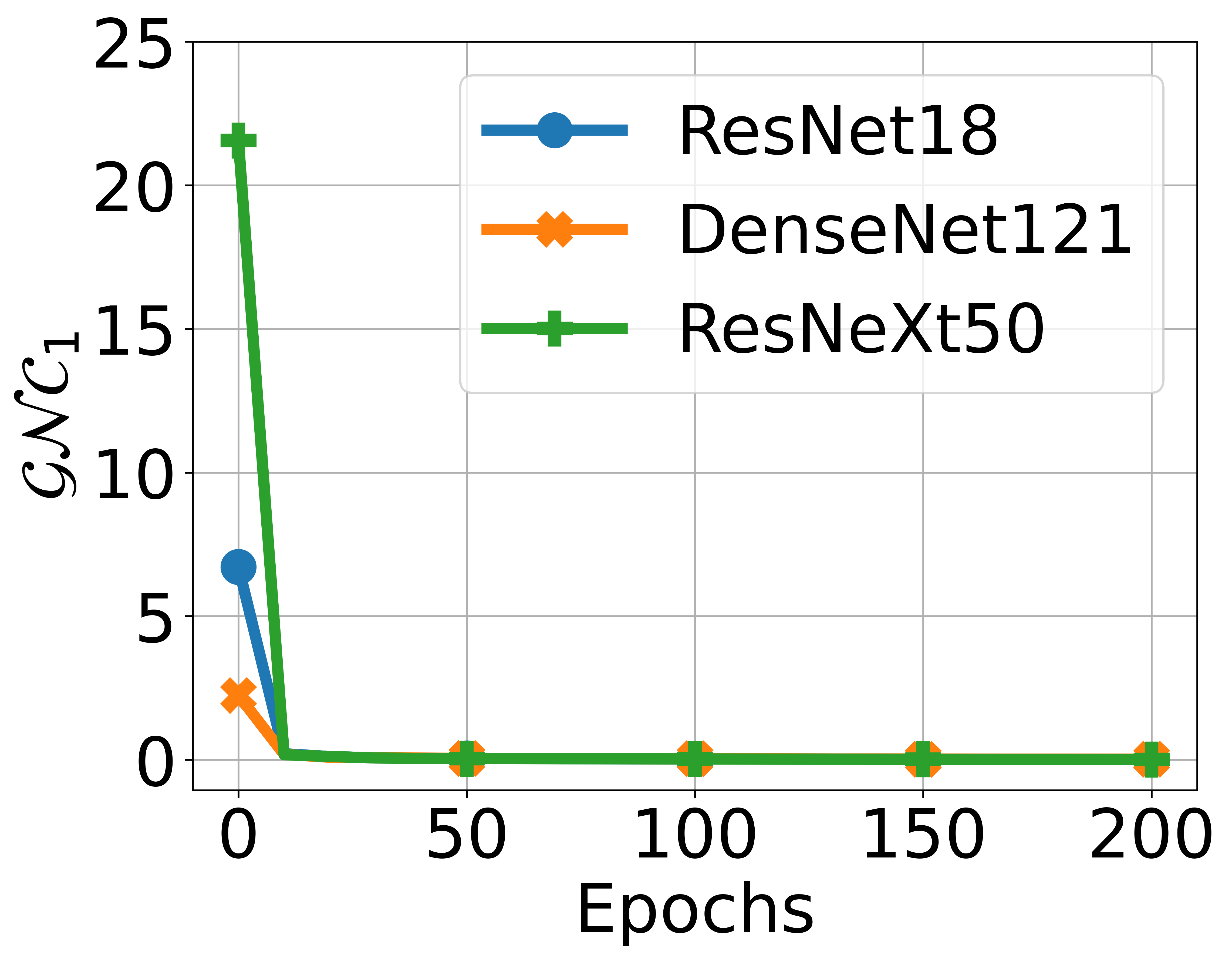}}\
     \subfloat[$\mathcal{GNC}_2$ (CIFAR100)]{\includegraphics[width=0.24\textwidth]{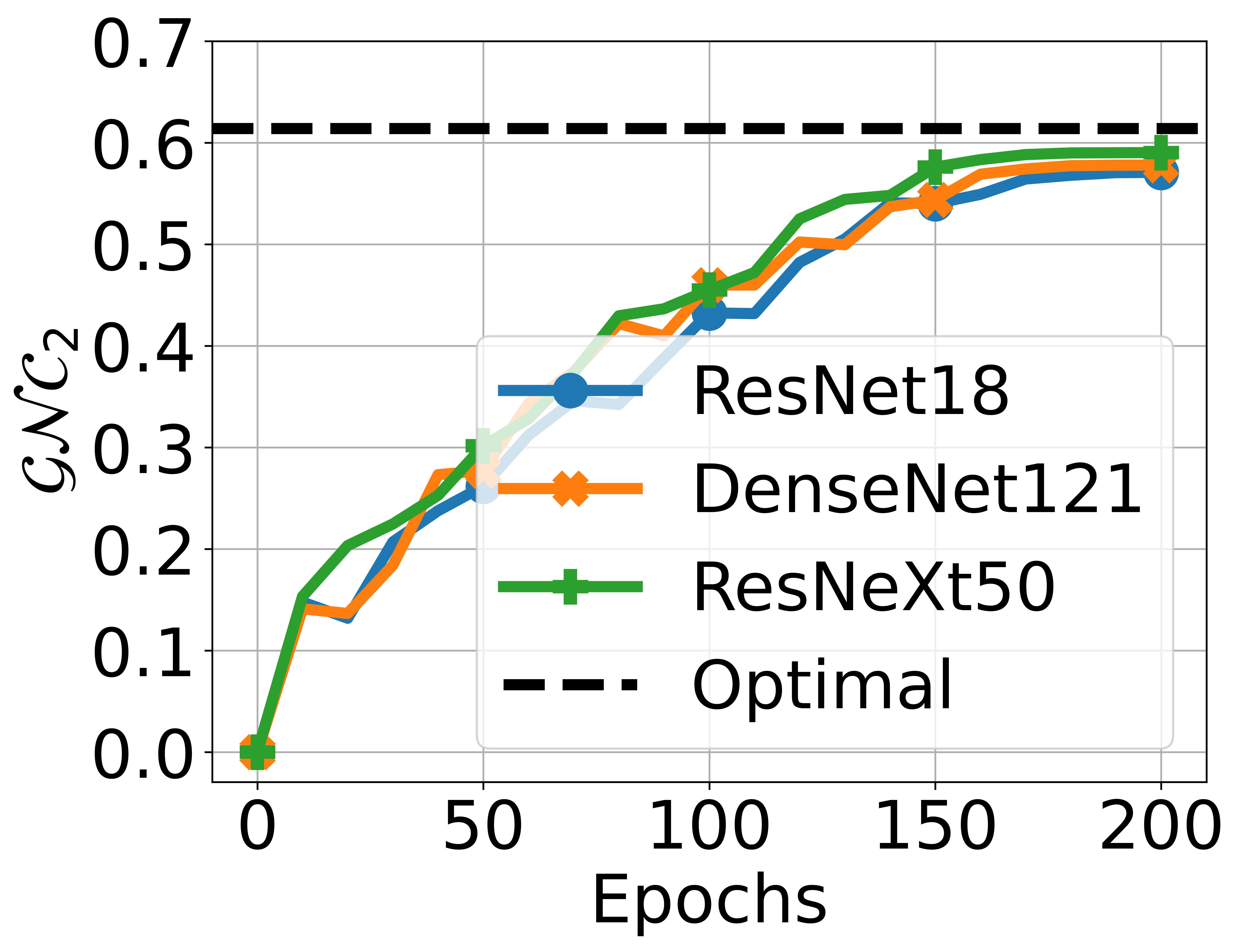}}\
    \subfloat[$\mathcal{GNC}_3$ (CIFAR100)]{\includegraphics[width=0.24\textwidth]{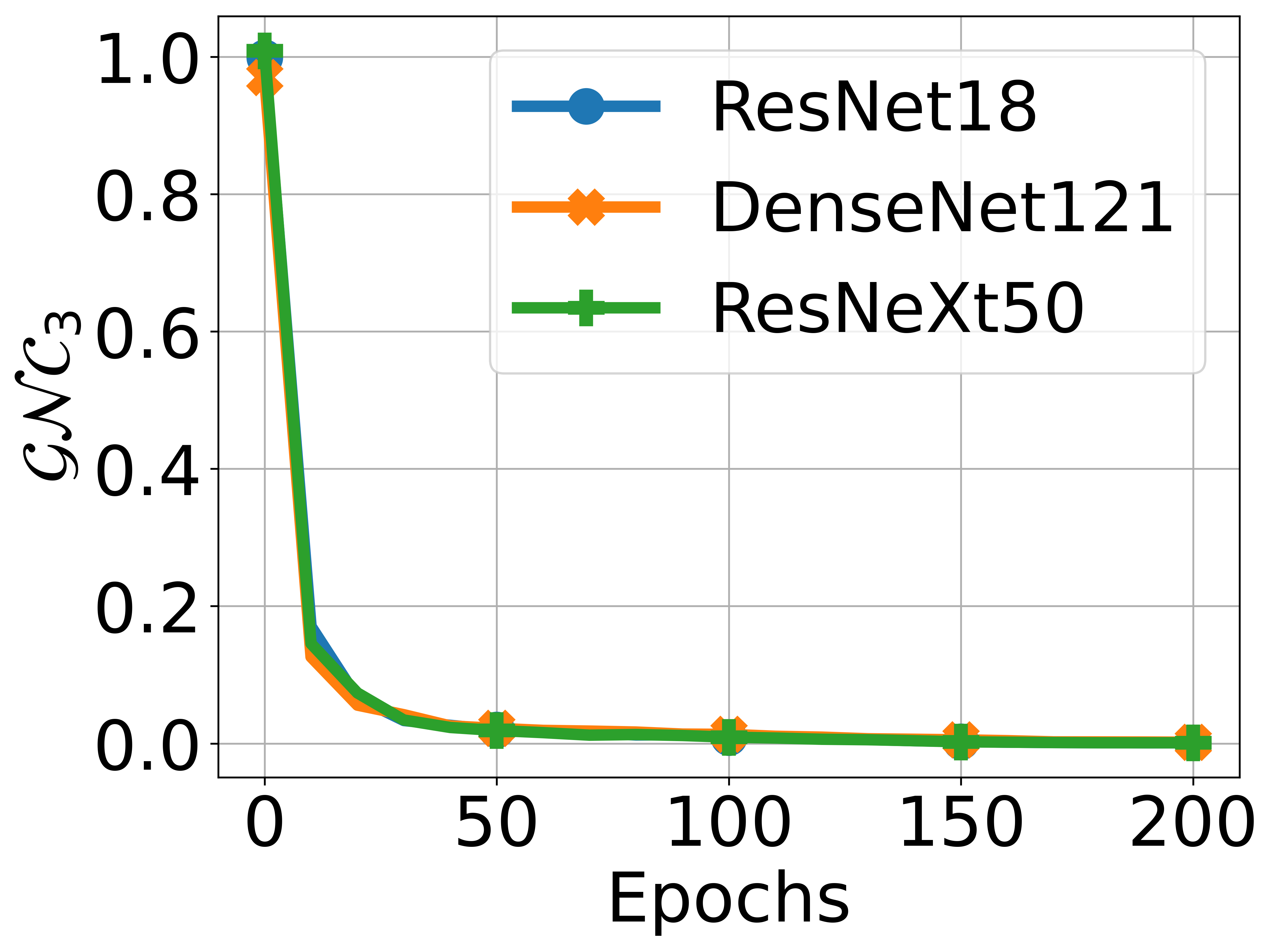}}\
    \subfloat[Train Acc (CIFAR100)]{\includegraphics[width=0.24\textwidth]{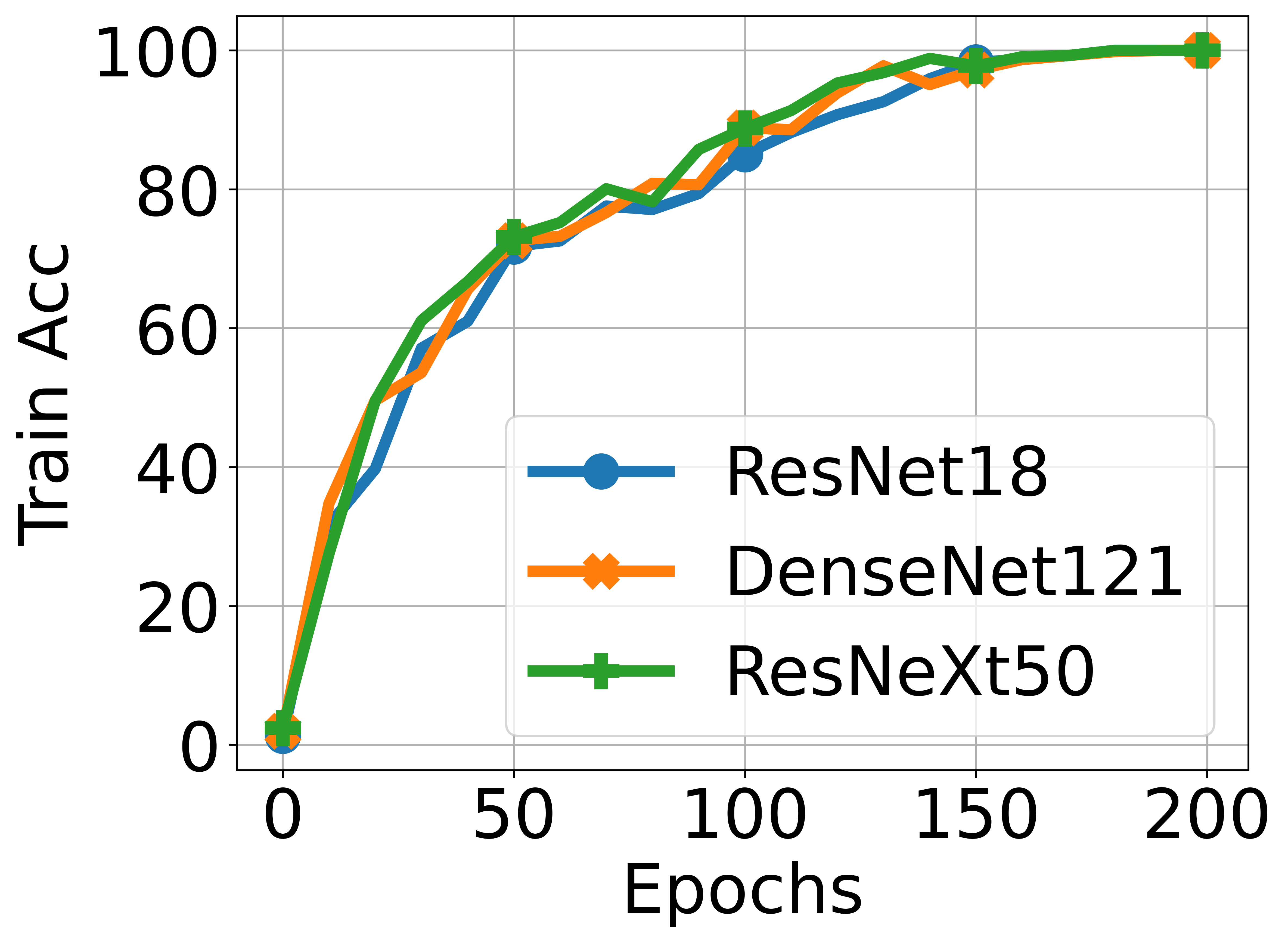}}\\

     \subfloat[$\mathcal{GNC}_1$ (Tiny)]{\includegraphics[width=0.24\textwidth]{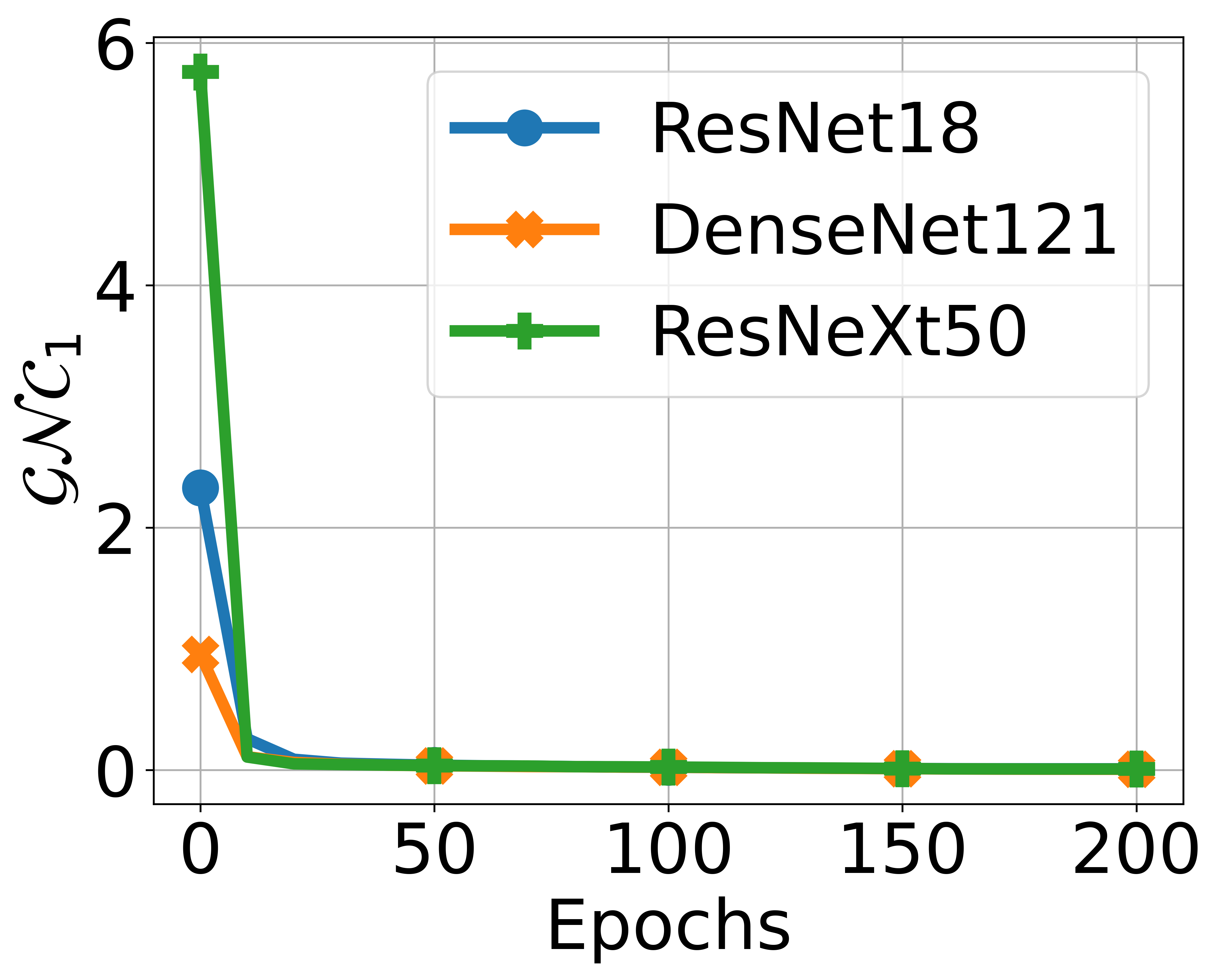}}\
     \subfloat[$\mathcal{GNC}_2$ (Tiny)]{\includegraphics[width=0.24\textwidth]{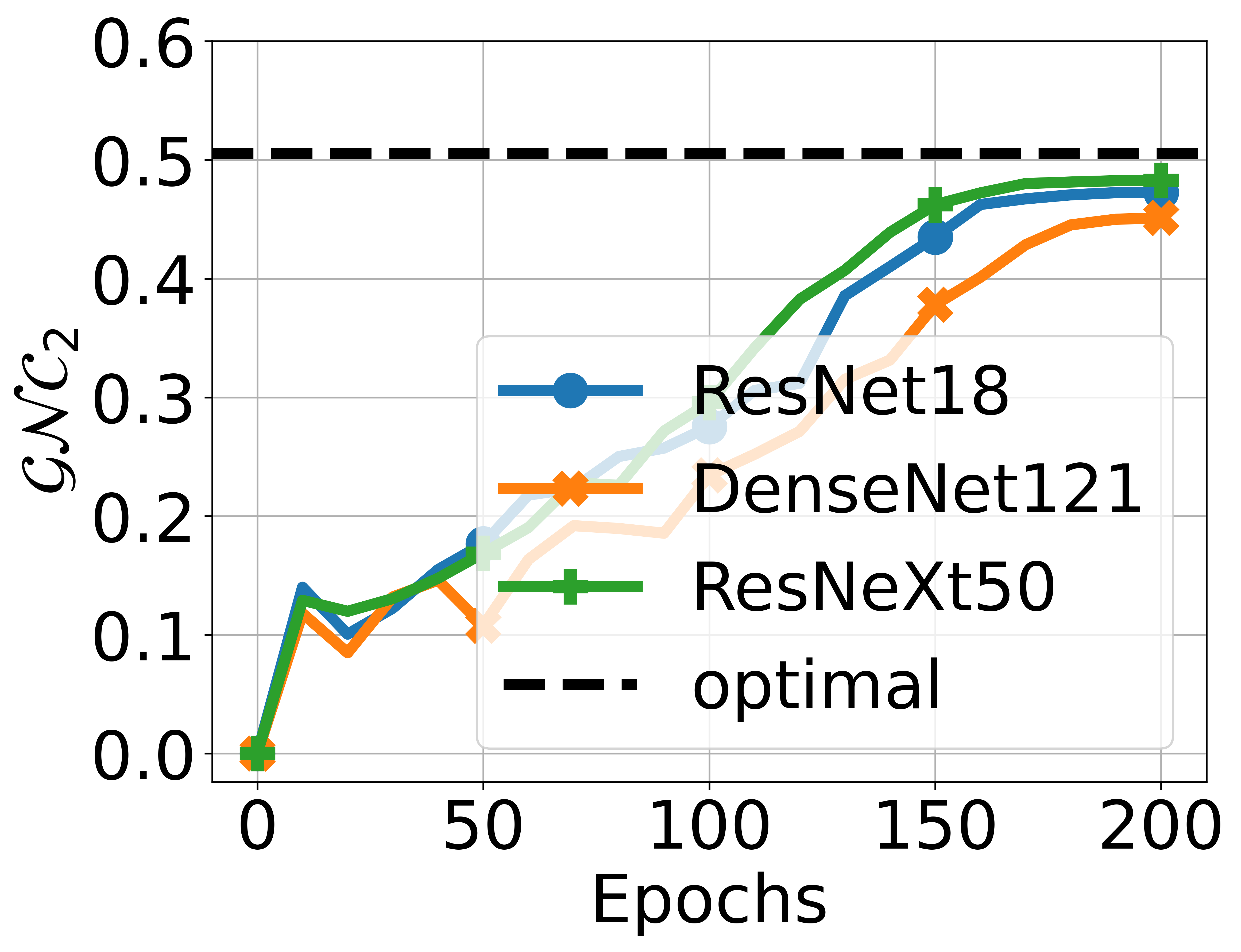}}\
    \subfloat[$\mathcal{GNC}_3$ (Tiny)]{\includegraphics[width=0.24\textwidth]{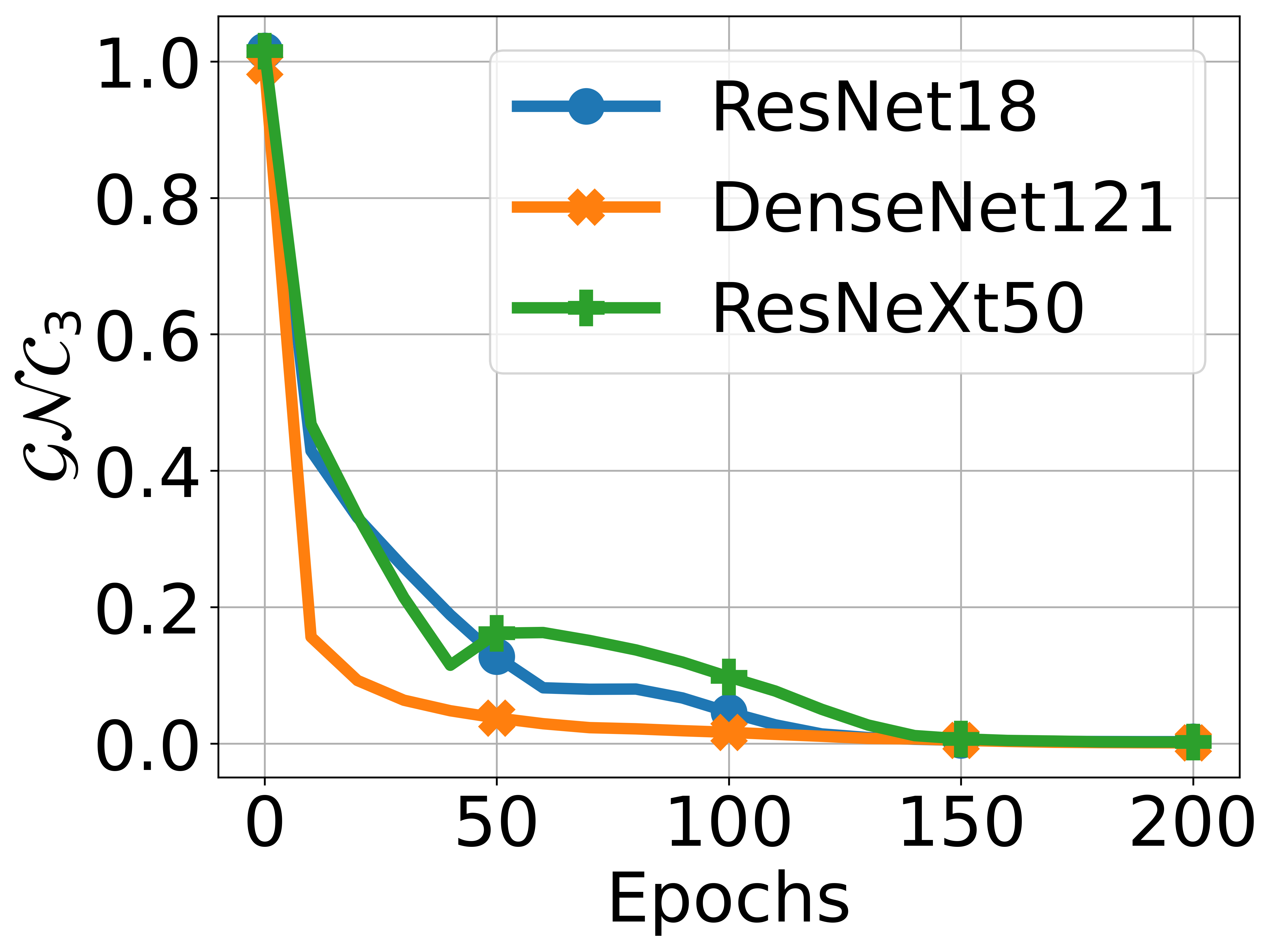}}\
    \subfloat[Train Acc (Tiny)]{\includegraphics[width=0.24\textwidth]{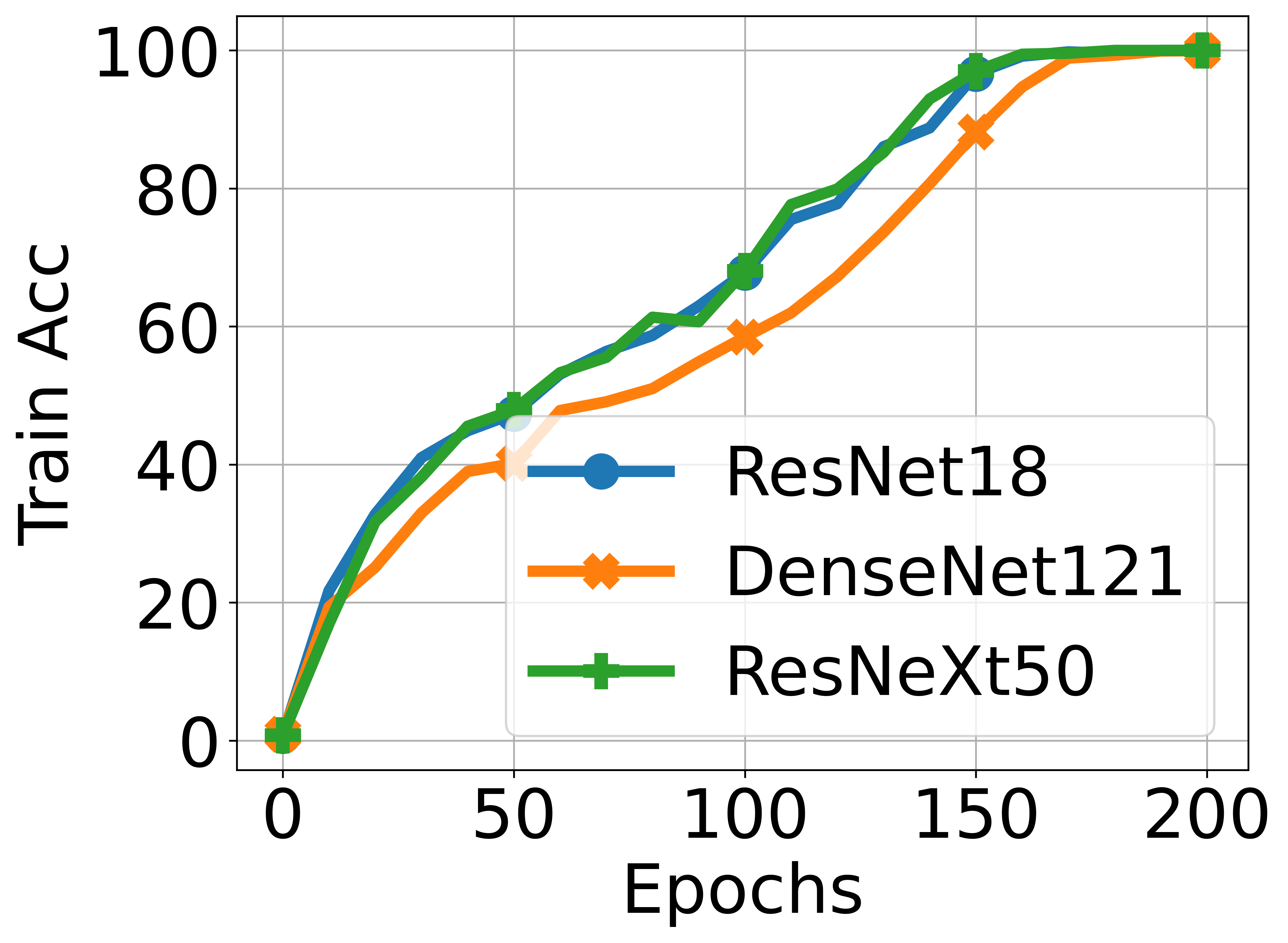}}

    \subfloat[$\mathcal{GNC}_1$ (Face)]{\includegraphics[width=0.24\textwidth]{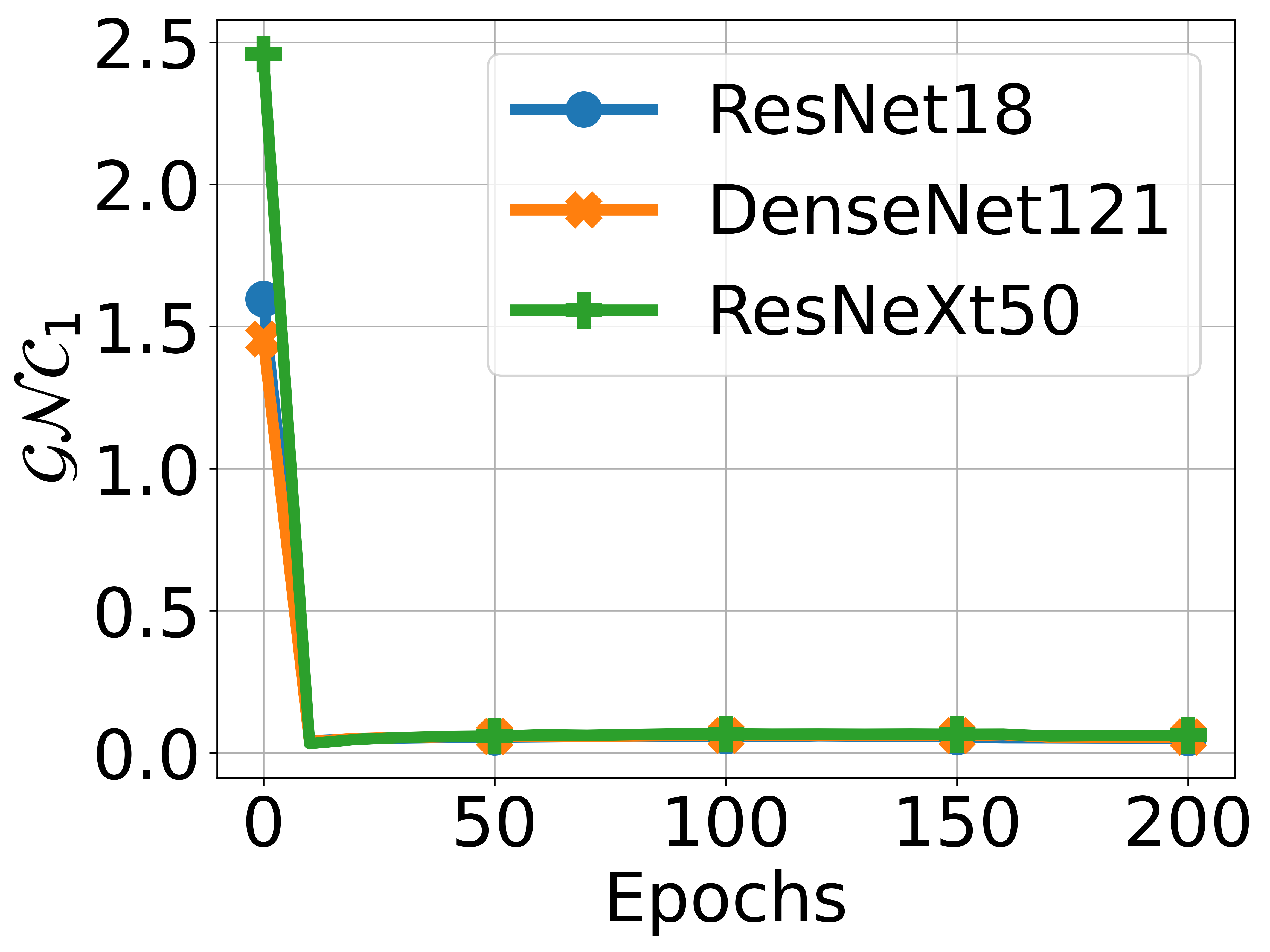}}\
     \subfloat[$\mathcal{GNC}_2$ (Face)]{\includegraphics[width=0.24\textwidth]{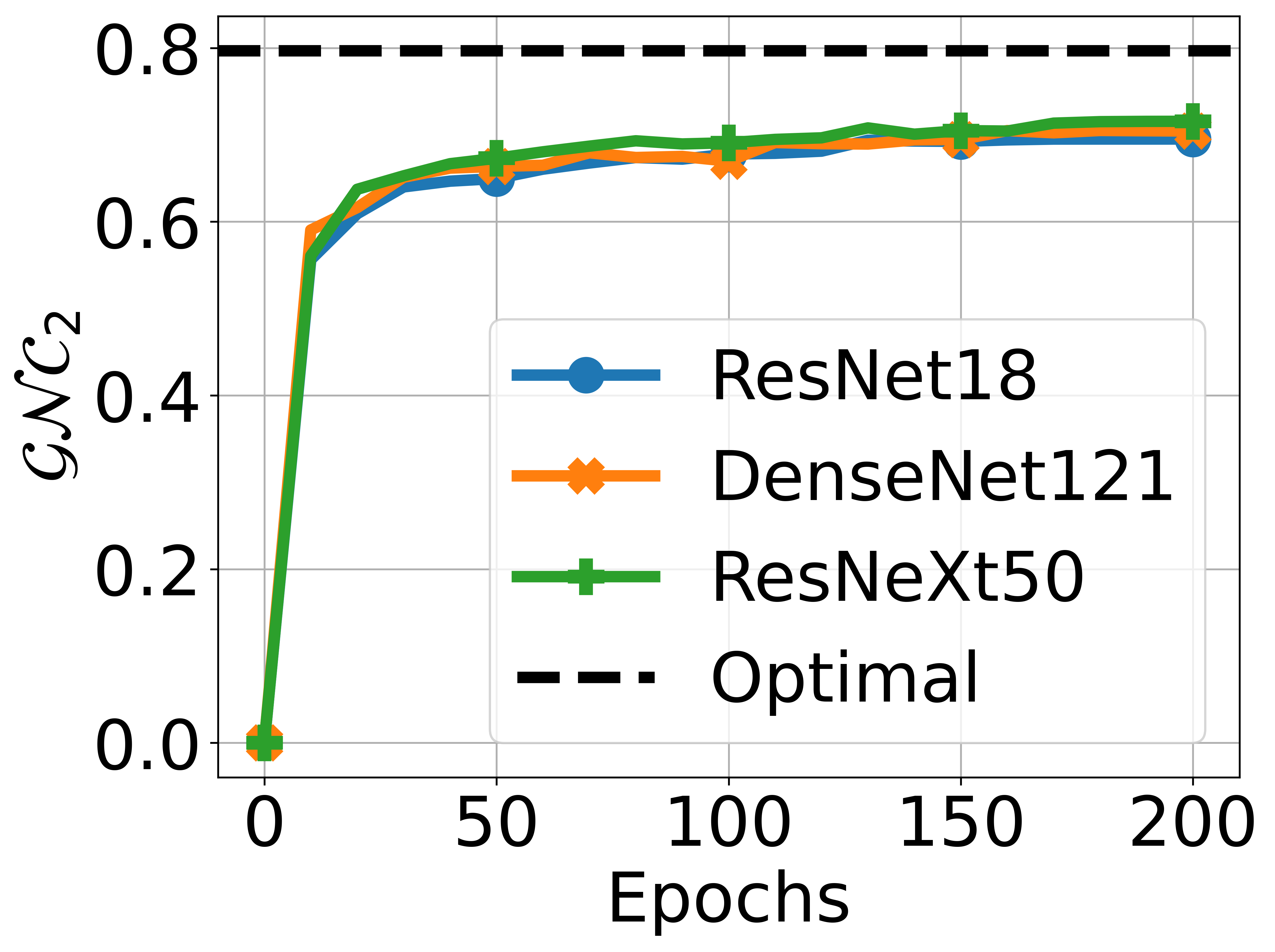}}\
    \subfloat[$\mathcal{GNC}_3$ (Face)]{\includegraphics[width=0.24\textwidth]{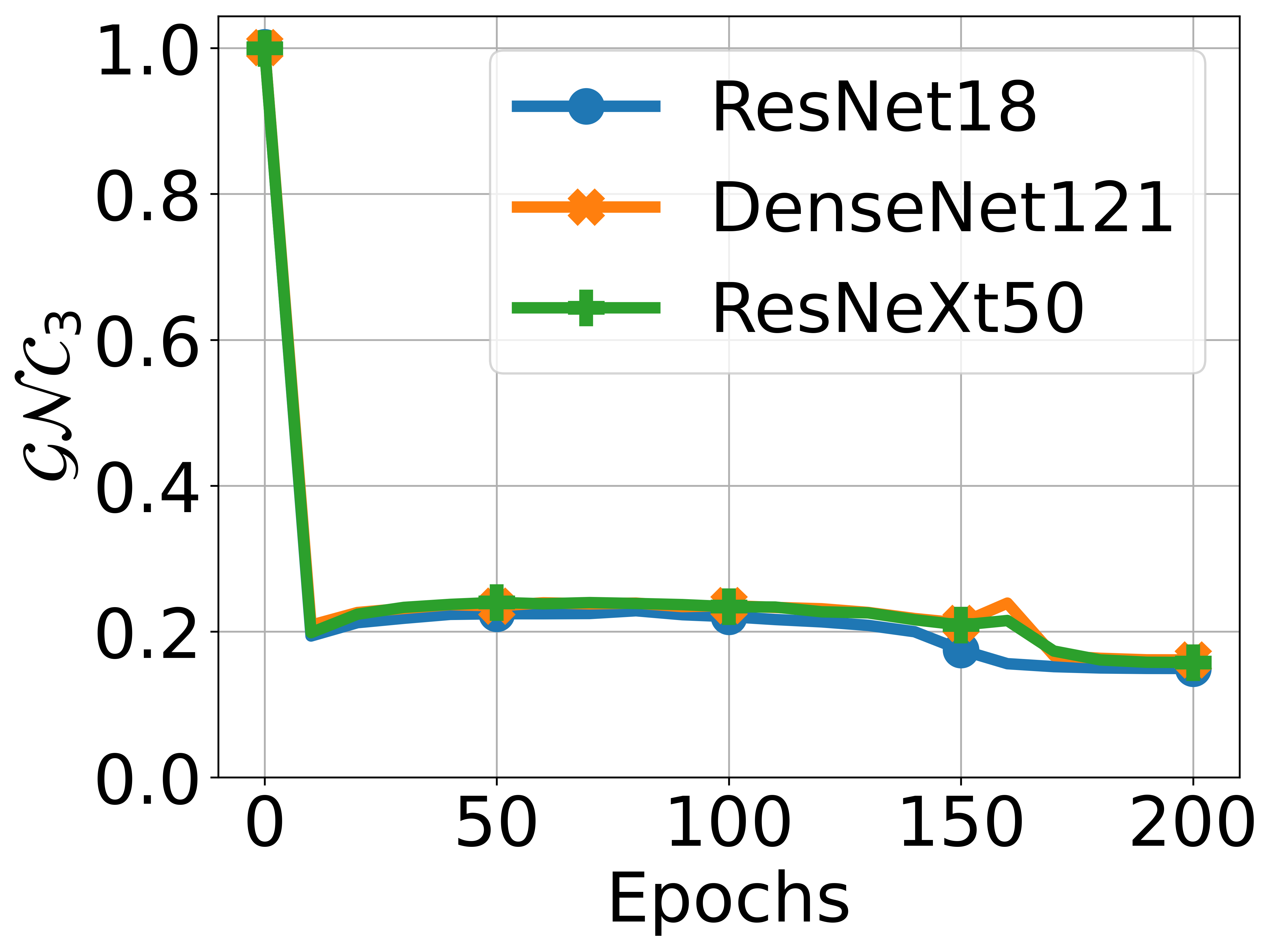}}\
    \subfloat[Train Acc (Face)]{\includegraphics[width=0.24\textwidth]{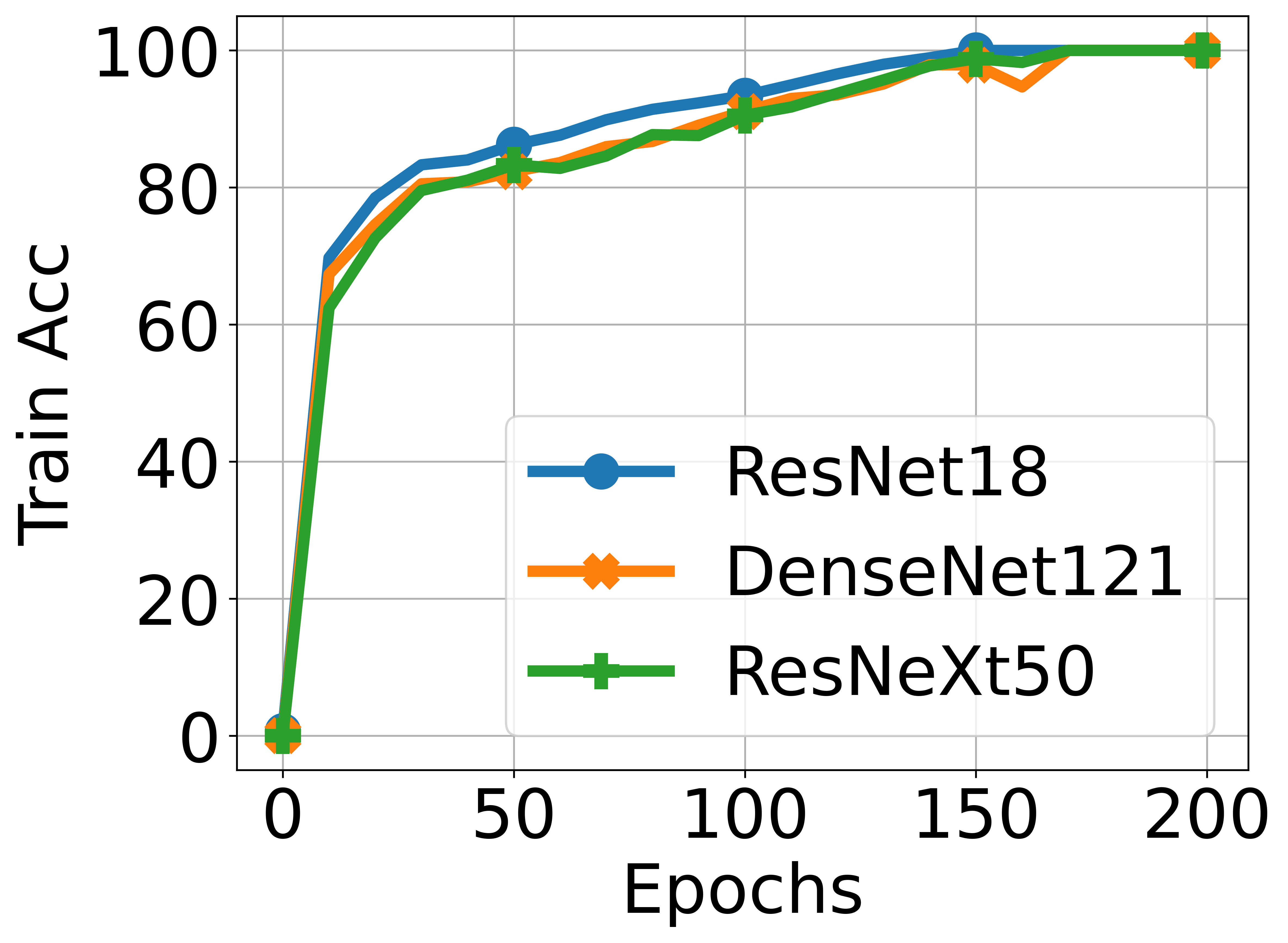}}

        \caption{\textbf{Illustration of \GNC\ and train accuracy across different network architectures on CIFAR100(top), Tiny-ImageNet(middle) and BUPT-CBFace-50(bottom) datasets.} 
        We train the networks on CIFAR100 with $d = 10 , \ K = 100$, Tiny-ImageNet with $d = 10 , \ K = 200$ and BUPT-CBFace-50 with $d = 512 , \ K = 10000$.
        }
        \label{fig:MoreNC}
\end{figure}

The results are reported in \Cref{fig:MoreNC}. 
It can be seen that in all the cases the \GNC$_1$, \GNC$_2$ and \GNC$_3$ measures converge mostly monotonically as a function of the training epochs towards the values predicted by \GNC (\ie, $0$ for \GNC$_1$ and \GNC$_3$ and the objective of Softmax Code for \GNC$_2$, see \Cref{subsec:gnc}).

\begin{figure}
\centering
\includegraphics[width=0.35\textwidth]{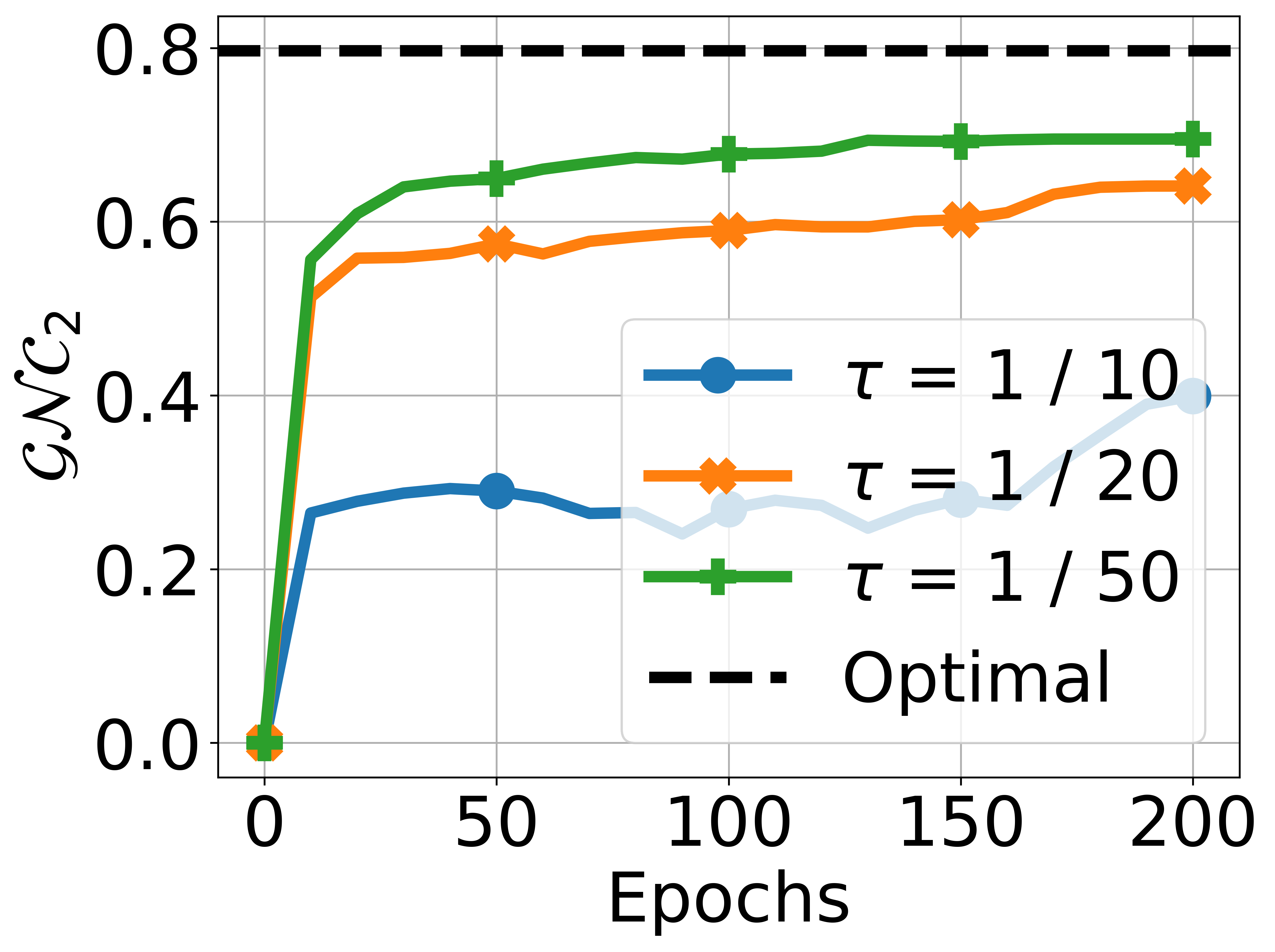}\
\caption{\textbf{Illustration of \GNC$_2$\ on BUPT-CBFace-50 dataset across varying and temperatures.} 
        We train the networks on BUPT-CBFace-50 with $d = 512 , \ K = 10,000$.
        }
    \label{fig:MoreNC_temperature}
\end{figure}

\paragraph{Effect of temperature $\tau$.} 
The results on BUPT-CBFace-50 reported in \Cref{fig:MoreNC} uses a temperature $\tau = 0.02$. To examine the effect of $\tau$, we conduct experiments with varying $\tau$ and report the \GNC$_2$ in \Cref{fig:MoreNC_temperature}. It can be seen that the \GNC$_2$ measure at convergence monotonically increases as $\tau$ decreases.

\paragraph{Implementation detail.}
We choose the temperature $\tau = 0.1$ for CIFAR100 and Tiny-ImageNet and $\tau = 0.02$ for the BUPT-CBFace-50 dataset. The CIFAR100 dataset consists of $60,000$ $32 \times 32$ color images in $100$ classes and Tiny-ImageNet contains $100,000$ $64\times 64$ color images in $200$ classes. The BUPT-CBFace-50 dataset consists of $500,000$ images in $10,000$ classes and all images are resized to the size of $50 \times 50$. To adapt to the smaller size images, we modify these architectures by changing the first convolutional layer to have a kernel size of $3$, a stride of $1$, and a padding size of $1$. For our data augmentation strategy, we employ the random crop with a size of $32$ and padding of $4$, random horizontal flip with a probability of $0.5$, and random rotation with degrees of $15$ to increase the diversity of our training data. Then we normalize the images (channel-wise) using their mean and standard deviation. For optimization, we utilized SGD with a momentum of $0.9$ and an initial learning rate of $0.1$, which decayed according to the CosineAnnealing over a span of $200$ epochs. The optimal margin (represented by the dash line) in the second from left column is obtained through numerical optimization of the Softmax Codes problem.

\begin{figure}
\centering

  \includegraphics[width=0.35\linewidth]{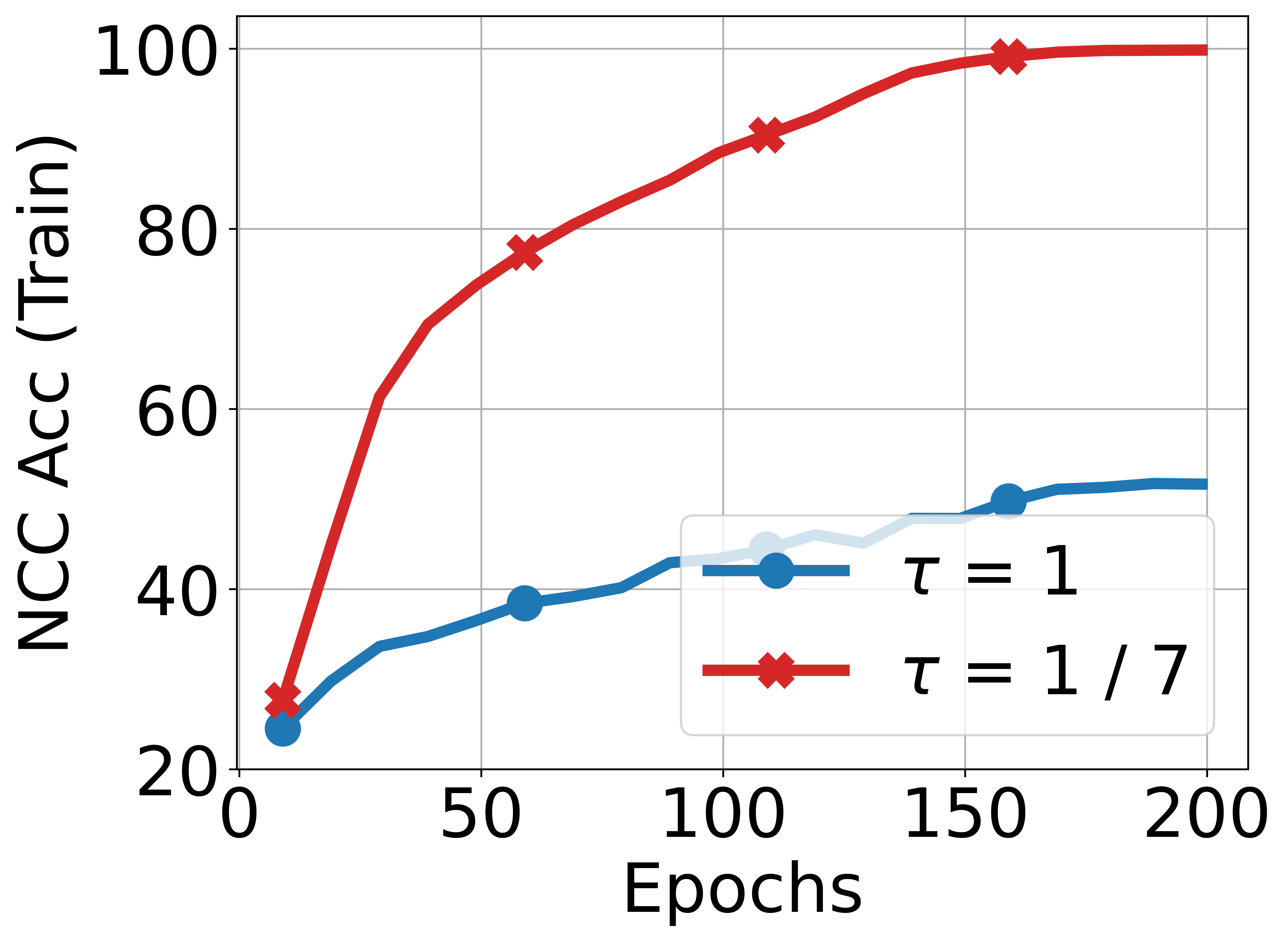}
  \captionof{figure}{\textbf{Illustration of NCC training accuracy with different temperatures.}}
  \label{fig:NCC}
\end{figure}


\subsection{The Nearest Centroid Classifier }
As the learned features exhibit within-class variability collapse and are maximally distant between classes, the classifier also converges to the nearest centroid classifier (a.k.a nearest class-center classifier (NCC), where each sample is classified with the nearest class-mean features), which is termed as \NC$_4$ in \cite{papyan2020prevalence} and exploited in \citep{galanti2022role, galanti2022generalization,rangamani2023feature} for studying \NC. To evaluate the convergence in terms of NCC accuracy, we use the same setup as in \Cref{fig:NC-dynamic}, i.e.,  train a ResNet18 network on the CIFAR100 dataset with CE loss using different temperatures. We then classify the features by the NCC. The result is presented in \Cref{fig:NCC}. We can observe that with a relatively small temperature $\tau$, the NCC accuracy converges to 100\% and hence the classifier also converges to a NCC.

\subsection{Additional results on the effect of assignment problem}
We provide additional evidence on the effect of "class assignment" problem in practical multi-class classification problems.
To illustrate this, we train a ResNet18 network with d = 2 on six
classes \{Automobile, Cat, Deer, Dog, Horse, Truck\} from CIFAR10 dataset that are selected due to their clear semantic similarity and discrepancy. Consequently, according to \Cref{thm:special-case-NC3}, there are fifteen distinct class assignments up to permutation. 

\begin{figure} [!ht]
     \centering
     \begin{subfigure}[b]{0.24\textwidth}
         \centering
         \includegraphics[width=\textwidth]{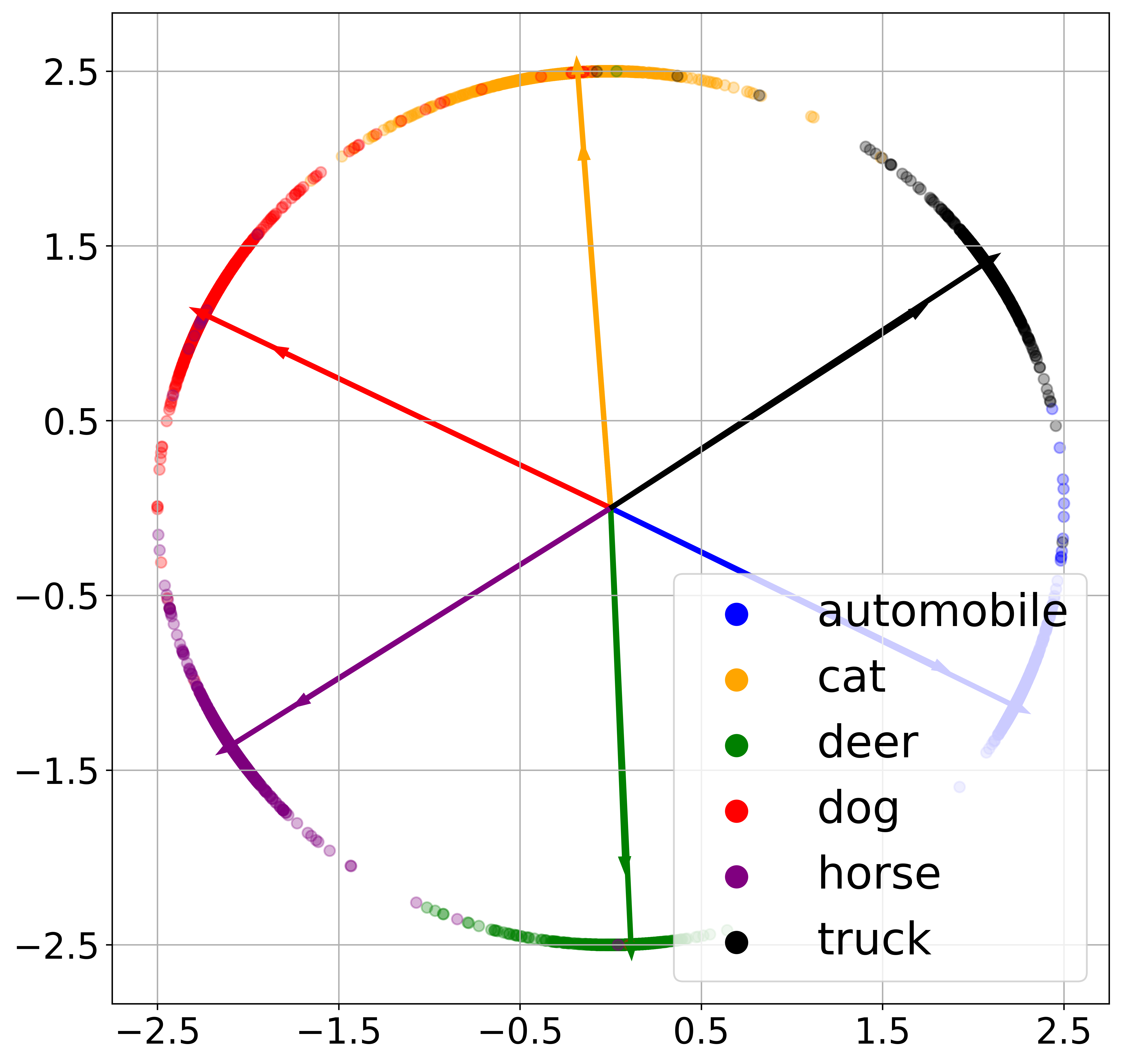}
         \caption{Trainable ($91.82\%$)}
         \label{fig:learned-assignment6c}
     \end{subfigure}
     \hfill
     \begin{subfigure}[b]{0.24\textwidth}
         \centering
         \includegraphics[width=\textwidth]{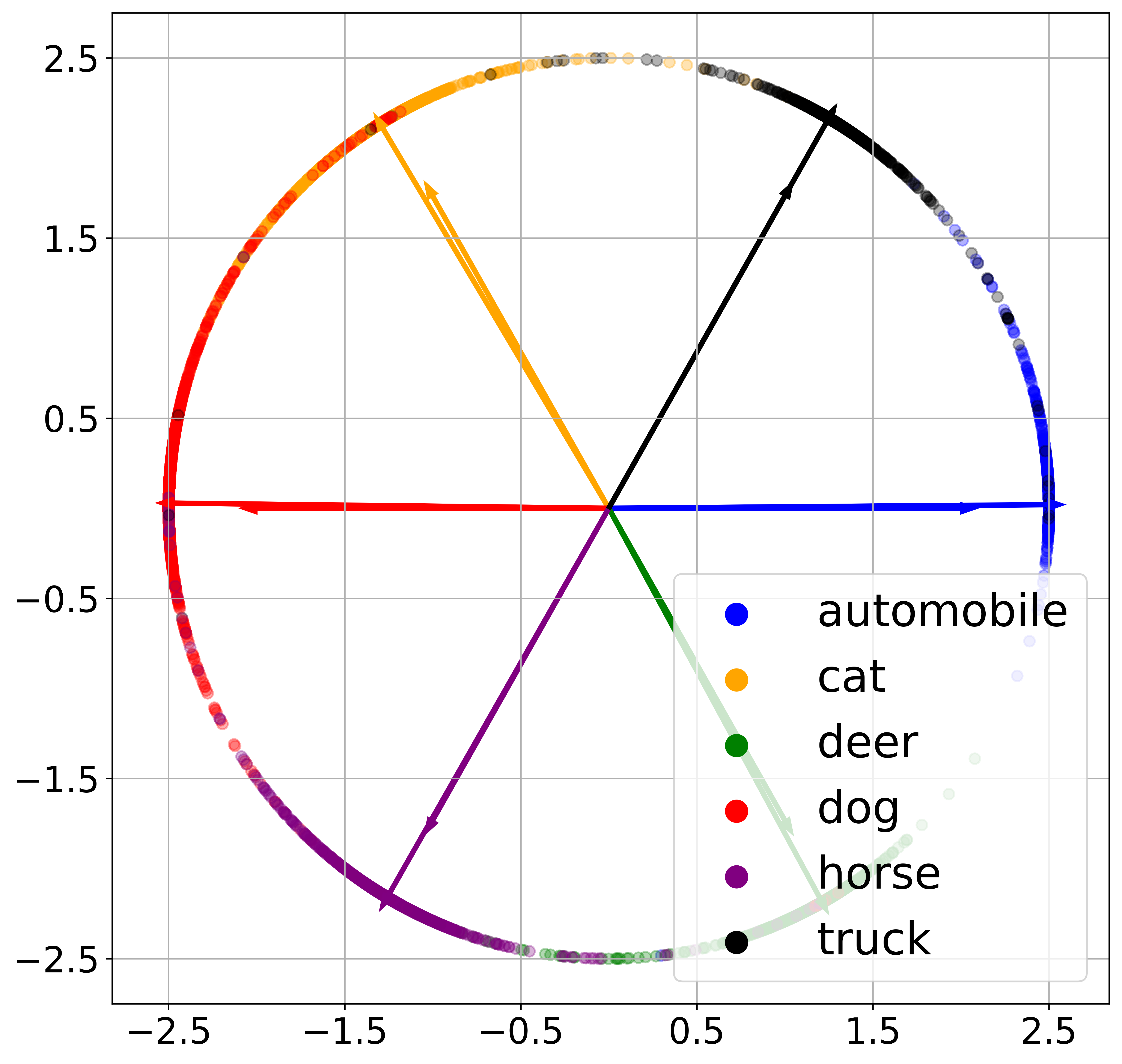}
         \caption{Fixed A1 ($91.60\%$)}
         \label{fig:assignment6c_1}
     \end{subfigure}
     \hfill
     \begin{subfigure}[b]{0.24\textwidth}
         \centering
         \includegraphics[width=\textwidth]{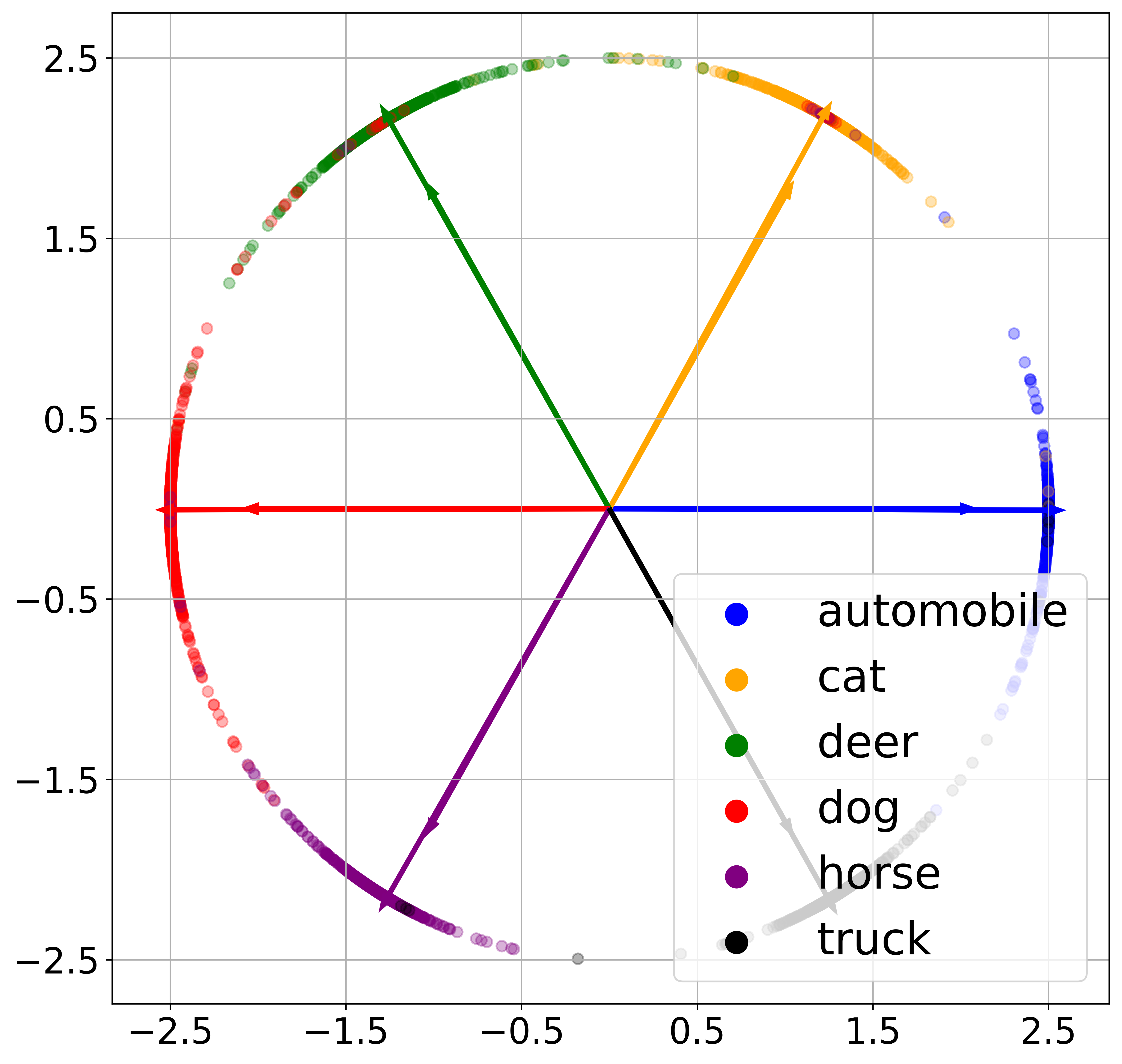}
         \caption{Fixed A2 ($91.95\%$)}
         \label{fig:assignment6c_2}
    \end{subfigure}
    \hfill
     \begin{subfigure}[b]{0.24\textwidth}
         \centering
         \includegraphics[width=\textwidth]{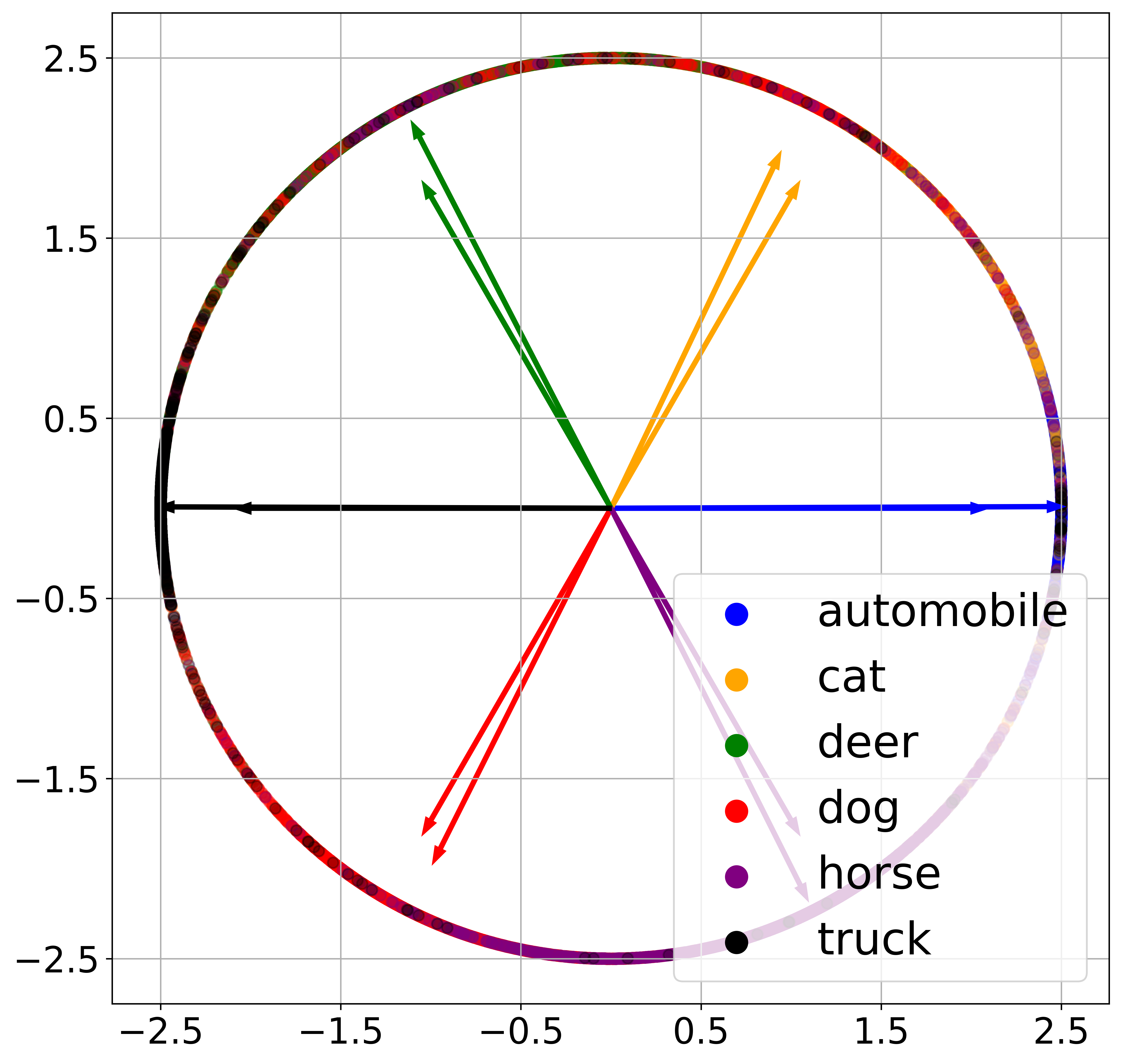}
         \caption{Fixed A3 ($86.37\%$)}
         \label{fig:assignment6c_3}
     \end{subfigure}\
        \caption{\textbf{Assignment of classes to classifier weights} for a ResNet18 with 2-dimensional feature space trained on the 6 classes \{Automobile, Cat, Deer, Dog, Horse, Truck\} from CIFAR10. \emph{(a)} Learned classifier. \emph{(b-d)} Classifiers fixed to be three different assignments. Test accuracy is reported in the bracket.} 
        \label{fig:permutation6classes}
        \vspace{-.3in}
\end{figure}

When doing standard training, the classifier consistently converges to the case where Cat-Dog, Automobile-Truck and Deer-Horse pairs are closer together across 5 different trials; \Cref{fig:learned-assignment6c} shows the learned features (dots) and classifier weights (arrows) in one of such trials.
This demonstrates the implicit algorithmic regularization in training DNNs, which naturally attracts (semantically) similar classes and separates dissimilar ones. 

We also conduct experiments with the classifier fixed to be three of the fifteen arrangements, and present the results in \Cref{fig:assignment6c_1,fig:assignment6c_3}.
Among them, we observe that the case where Cat-Dog, Automobile-Truck and Deer-Horse pairs are far apart achieves a testing accuracy of $86.37 \%$, which is lower than the other two cases with testing accuracies of $91.60 \%$ and $91.95 \%$. 
This demonstrates the important role of class assignment to the generalization of DNNs, and that the implicit bias of the learned classifier is benign, \ie, leads to a more generalizable solutions. Moreover, compared with the case of four classes in \Cref{fig:permutation}, the discrepancy of test accuracy between different assignments increases from $2.18\%$ to $5.58\%$. This suggests that as the number of classes grows, the importance of appropriate class assignments becomes increasingly paramount.

\section{Theoretical Proofs}\label{app:thm}
\subsection{Proof of \Cref{thm:converge-to-hardmax}}

We rewrite \Cref{thm:converge-to-hardmax} below for convenience.

\begin{lemma}[Convergence to the ``HardMax" problem] For any positive integers $K$ and $n$, we have
\begin{equation}
    \limsup_{\tau \to 0} \paren{ \argmin \frac{1}{nK}\sum_{k=1}^{K} \sum_{i=1}^{n} \mathcal{L}_{\text{CE}}\paren{\mW^\top\vh_{k,i} , \vy_k,\tau} }
    \subseteq \argmin \mathcal{L}_{\text{HardMax}}(\mW, \mH). 
\end{equation}
In above, all $\arg\min$ are taken over $\mW\in\OB(d,K),\mH\in\OB(d,nK)$. 
\end{lemma}

\begin{proof}
    Our proof uses the fundamental theorem of $\Gamma$-convergence \cite{braides2006handbook} (a.k.a. epi-convergence \cite{rockafellar2009variational}). 
    Denote
    \begin{equation}
        \mathcal{L}_0(\mW, \mH, \tau) \doteq \tau \cdot \log \sum_{i = 1}^{n} \sum_{k = 1}^{K} \mc{L}_{\text{CE}}(\mW^\top\vh_{k,i}, \vy_k,\tau),
    \end{equation}
    and let
    \begin{equation}\label{eq:def-wh-minus}
        \mathcal{WH}^- \doteq \{(\mW, \mH) \in \OB(d,K) \times \OB(d,K): (\vw_{k'} - \vw_k)^\top \vh_{k,i} \le 0, \forall i \in [n], k \in [K], k' \in [K] \setminus k\}.
    \end{equation}
    By \Cref{thm:uniform-convergence}, the function $\mathcal{L}_0(\mW, \mH, \tau)$ converges uniformly to $\mathcal{L}_{\text{HardMax}}(\mW, \mH)$ on $\mathcal{WH}^-$ as $\epsilon \to 0$. 
    Combining it with \cite[Proposition 7.15]{rockafellar2009variational}, we have $\mathcal{L}_0(\mW, \mH, \tau)$ $\Gamma$-converges to $\mathcal{L}_{\text{HardMax}}(\mW, \mH)$ on $\mathcal{WH}^-$ as well.
    By applying \cite[Theorem 2.10]{braides2006handbook}, we have
    \begin{equation}
    \limsup_{\tau \to 0} \argmin_{(\mW, \mH) \in \mathcal{WH}^-} \mathcal{L}_0(\mW, \mH, \tau)
    \subseteq \argmin_{(\mW, \mH) \in \mathcal{WH}^-} \mathcal{L}_{\text{HardMax}}(\mW, \mH).
    \end{equation}
    Note that in above, $\arg\min$ are taken over $\mathcal{WH}^-$ which is a strict subset of $\mW\in\OB(d,K),\mH\in\OB(d,nK)$. 
    However, by \Cref{thm:negative-correlation-is-optimal}, we know that
    \begin{equation}
        \argmin_{(\mW, \mH) \in \mathcal{WH}^-} \mathcal{L}_0(\mW, \mH, \tau) = \argmin_{(\mW, \mH) \in \OB(d,K) \times \OB(d,K)} \mathcal{L}_0(\mW, \mH, \tau),
    \end{equation}
    which holds for all $\tau$ sufficiently small. 
    Hence, we have
    \begin{equation}
    \limsup_{\tau \to 0} \argmin_{(\mW, \mH) \in \OB(d,K) \times \OB(d,K)} \mathcal{L}_0(\mW, \mH, \tau)
    \subseteq \argmin_{(\mW, \mH) \in \OB(d,K) \times \OB(d,K)} \mathcal{L}_{\text{HardMax}}(\mW, \mH),
    \end{equation}
    which concludes the proof. 
\end{proof}

\begin{lemma}\label{thm:uniform-convergence} 
$\mathcal{L}_0(\mW, \mH, \tau)$ converges uniformly to $\mathcal{L}_{\text{HardMax}}(\mW, \mH)$ in the domain $(\mW, \mH) \in \mathcal{WH}^-$ as $\tau \to 0$.
\end{lemma}

\begin{proof}
Recall from the definition of the CE loss in \eqref{def:ce} that
\begin{multline}
    \mathcal{L}_0(\mW, \mH, \tau) \doteq \tau \cdot \log \sum_{i = 1}^{n} \sum_{k = 1}^{K} \mc{L}_{\text{CE}}(\mW^\top\vh_{k,i}, \vy_k,\tau) 
    \\= \tau \log \sum_{i = 1}^{n} \sum_{k = 1}^{K}   \log\paren{1 + \sum_{k' \in [K] \setminus k} \exp\paren{(\vw_{k'} - \vw_k)^\top \vh_{k,i} /\tau} }
\end{multline}

Denote $\alpha_{i, k, k'} = (\vw_{k'} - \vw_k)^\top \vh_{k,i}, \forall i \in [n], k \in [K], k' \in [K] \setminus k$ for convenience.
Fix any $i \in [n]$ and $k \in [K]$, by the property that $\frac{x}{1+x} \le \log(1+x) \le x$ for all $x > -1$, we have
\begin{equation}
     \frac{\sum_{k' \in [K] \setminus k} \exp\paren{\frac{\alpha_{i, k, k'}}{\tau}}}{1 + \sum_{k' \in [K] \setminus k} \exp\paren{\frac{\alpha_{i, k, k'}}{\tau}}}
     \le \log\paren{1 + \sum_{k' \in [K] \setminus k} \exp\paren{\frac{\alpha_{i, k, k'}}{\tau}} } 
     \le \sum_{k' \in [K] \setminus k} \exp\paren{\frac{\alpha_{i, k, k'}}{\tau}}.
\end{equation}
Note that in the domain $(\mW, \mH) \in \mathcal{WH}^-$ we have $\alpha_{i, k, k'} \le 0$ for all $i, k, k'\ne k$. 
It follows trivially from monotonicity of exponential function that $\sum_{k' \in [K] \setminus k} \exp\paren{\frac{\alpha_{i, k, k'}}{\tau}} < K-1$ holds for all $i, k$. 
Hence, continuing from the inequality above we have
\begin{equation}
     \frac{\sum_{k' \in [K] \setminus k} \exp\paren{\frac{\alpha_{i, k, k'}}{\tau}}}{K}
     \le \log\paren{1 + \sum_{k' \in [K] \setminus k} \exp\paren{\frac{\alpha_{i, k, k'}}{\tau}} } 
     \le \sum_{k' \in [K] \setminus k} \exp\paren{\frac{\alpha_{i, k, k'}}{\tau}}.
\end{equation}
Summing over $i, k$ we obtain 
\begin{equation}
     \sum_{i, k}\frac{\sum_{k' \in [K] \setminus k} \exp\paren{\frac{\alpha_{i, k, k'}}{\tau}}}{K}
     \le \sum_{i, k}\log\paren{1 + \sum_{k' \in [K] \setminus k} \exp\paren{\frac{\alpha_{i, k, k'}}{\tau}} } 
     \le \sum_{i, k}\sum_{k' \in [K] \setminus k} \exp\paren{\frac{\alpha_{i, k, k'}}{\tau}},
\end{equation}
Using the property of max function we further obtain
\begin{multline}
     \frac{\max_{i, k, k' \in [K] \setminus k} \exp\paren{\frac{\alpha_{i, k, k'}}{\tau}}}{K}
     \le \sum_{i, k} \log\paren{1 + \sum_{k' \in [K] \setminus k} \exp\paren{\frac{\alpha_{i, k, k'}}{\tau}} } \\
     \le n\cdot K \cdot (K-1)\cdot \max_{i, k, k' \in [K] \setminus k} \exp\paren{\frac{\alpha_{i, k, k'}}{\tau}}.
\end{multline}
Taking logarithmic on both sides and multiplying all terms by $\tau$ we get
\begin{equation}
    \max_{i, k, k' \in [K] \setminus k} \alpha_{i, k, k'}- \tau \log K 
    \le \mathcal{L}_0(\mW, \mH, \tau)
    \le \tau \log(n\cdot K \cdot (K-1)) + \max_{i, k, k' \in [K] \setminus k} \alpha_{i, k, k'}.
\end{equation}
Noting that $\max_{i, k, k' \in [K] \setminus k} \alpha_{i, k, k'} = \mathcal{L}_{\text{HardMax}}(\mW, \mH)$, we have
\begin{equation}
    \mathcal{L}_{\text{HardMax}}(\mW, \mH)- \tau \log K 
    \le \mathcal{L}_0(\mW, \mH, \tau)
    \le \tau \log(n\cdot K \cdot (K-1)) + \mathcal{L}_{\text{HardMax}}(\mW, \mH).
\end{equation}
Hence, for any $\epsilon >0$, by taking $\tau_0 = \frac{\epsilon}{\max\{\log K, \log(n\cdot K \cdot (K-1))\}}$, we have that for any $(\mW, \mH) \in \mathcal{WH}^-$, we have
\begin{equation}
    |\mathcal{L}_0(\mW, \mH, \tau) -  \mathcal{L}_{\text{HardMax}}(\mW, \mH)| \le \tau \max\{\log K, \log(n\cdot K \cdot (K-1))\} < \epsilon,
\end{equation}
for any $\tau < \tau_0$.
That is, $\mathcal{L}_0(\mW, \mH, \tau)$ converges uniformly to $\mathcal{L}_{\text{HardMax}}(\mW, \mH)$.

\end{proof}

\begin{lemma}\label{thm:negative-correlation-is-optimal} 
Given any $n, K$ there exists a constant $\tau_0$ such that for any $\tau < \tau_0$ we have
\begin{equation}
    \argmin_{(\mW, \mH) \in \OB(d,K) \times \OB(d,K)}\mathcal{L}_0(\mW, \mH, \tau) \in \mathcal{WH}^-.
\end{equation}
\end{lemma}
\begin{proof}
    Note that for any $(\mW, \mH) \notin \OB(d,K) \times \OB(d,K)$, there exists $\bar{i} \in [n], \bar{k} \in [K]$, and $\bar{k}' \in [K] \setminus \bar{k}$ such that $(\vw_{\bar{k}'} - \vw_{\bar{k}})^\top \vh_{\bar{k},\bar{i}} \ge 0$. Hence, we have 
    \begin{multline}
        \mathcal{L}_0(\mW, \mH, \tau) =  \tau \log \sum_{i = 1}^{n} \sum_{k = 1}^{K}   \log\paren{1 + \sum_{k' \in [K] \setminus k} \exp\paren{(\vw_{k'} - \vw_k)^\top \vh_{k,i} /\tau} }  \\
        \ge \tau \log \log(1 + \exp(\vw_{\bar{k}'} - \vw_{\bar{k}})^\top \vh_{\bar{k},\bar{i}} / \tau) 
        \ge \tau \log \log(2), \quad \forall \tau > 0.
    \end{multline}

    On the other hand, we show that one may construct a $(\mW^*, \mH^*) \in \mathcal{WH}^-$ such that $\mathcal{L}_0(\mW^*, \mH^*, \tau) < \tau \log \log(2)$ for any small enough $\tau$.
    Towards that, we take $\mW^*$ to be any matrix in $\OB(d,K)$ with distinct columns. Denote $M = \max_{k \ne k'} \langle \vw_k^*, \vw_{k'}^* \rangle$ the inner product of the closest pair of columns from $\mW^*$, which by construction has value $M < 1$.
    Take $\vh_{k, i} = \vw_k$ for all $k \in [K], i \in [n]$. 
    We have $(\vw_{k'}^* - \vw_{k}^*)^\top \vh_{k, i}^* \le - (1-M)$, for all $i\in [n], k \in [K]$, and $k' \in [K]\setminus k$. 
    Plugging this into the definition of $\mathcal{L}_0(\mW, \mH, \tau)$ we have
    \begin{multline}
        \mathcal{L}_0(\mW^*, \mH^*, \tau) =  \tau \log \sum_{i = 1}^{n} \sum_{k = 1}^{K}   \log\paren{1 + \sum_{k' \in [K] \setminus k} \exp\paren{(\vw_{k'}^* - \vw_k^*)^\top \vh_{k,i}^* /\tau} }\\
        \le \tau \log \paren{ n K \log (1 + (K-1) \exp{(-\frac{1-M}{\tau})}) } < \tau \log \log 2, \quad \forall \tau < \tau_0
    \end{multline}
    for some $\tau_0>0$ that depends only on $M$. 
    In above, the last inequality holds because one can always find a $\tau_0 > 0$ such that $n K \log (1 + (K-1) \exp{(-\frac{1-M}{\tau})}) <  \log 2$ for $\tau < \tau_0$. 
    This implies that any $(\mW, \mH) \notin \OB(d,K) \times \OB(d,K)$ is not a minimizer of $\mathcal{L}_0(\mW, \mH, \tau)$ for a sufficiently small $\tau$, which finishes the proof.
\end{proof}

\begin{figure}[!h]
\includegraphics[width=0.5\textwidth]{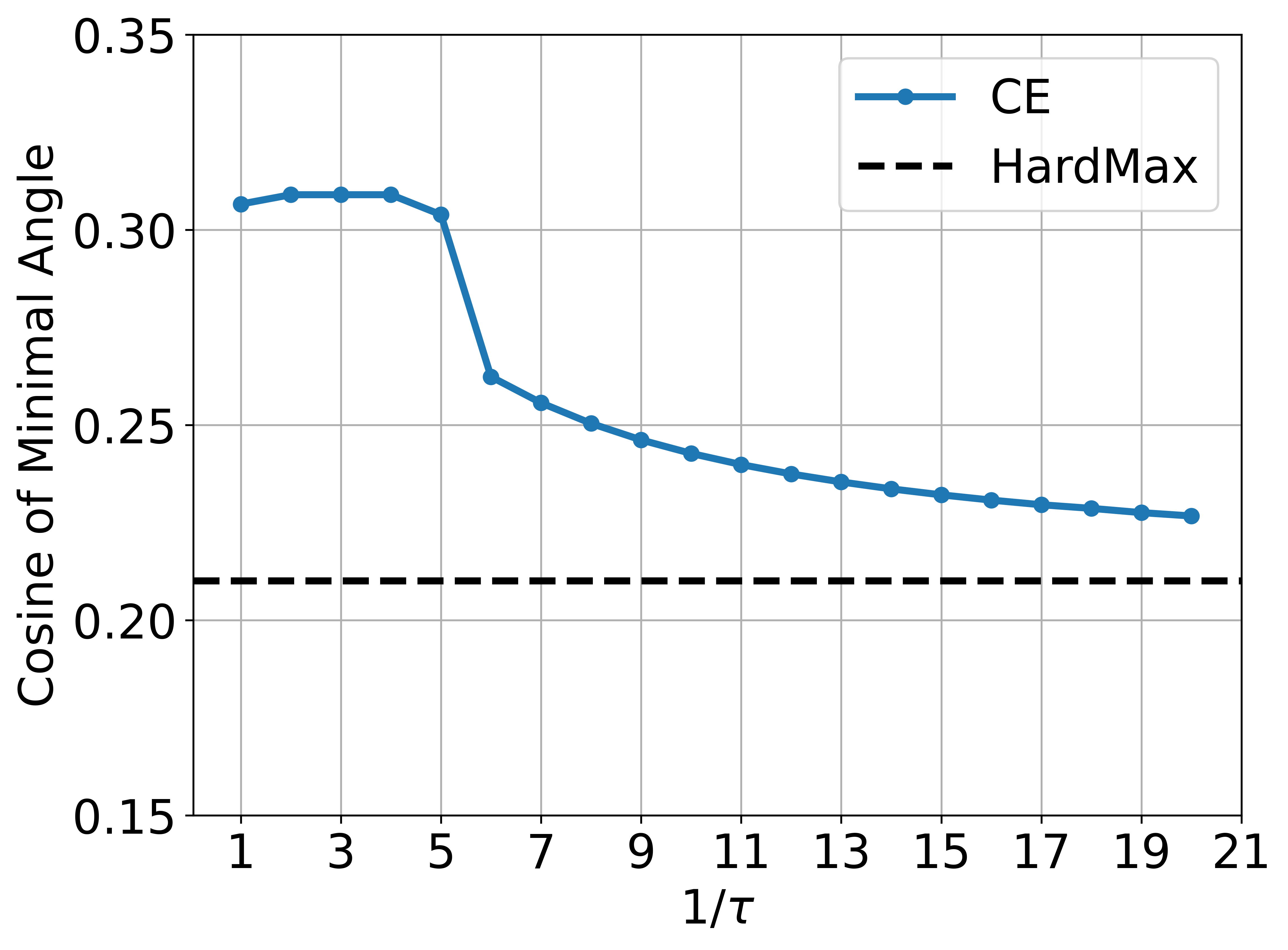}
\centering
\caption{Verifying \Cref{thm:converge-to-hardmax} under $d = 3$ and $K = 7$.}
\label{fig:hardmax-temp}
\end{figure}

As depicted in Figure \ref{fig:hardmax-temp}, we plot the cosine value of the minimal angle obtained from optimizing the CE loss (blue line) and the "HardMax" approach (black line) for different temperature parameters. The figure demonstrates that as the temperature parameter $\tau \rightarrow 0$, the blue line converges to the black line, thus validating our proof.

\subsection{An important lemma}


We present an important lemma which will be used to prove many of the subsequent results.

\begin{lemma}[Optimal Features for Fixed Classifier]\label{thm:max_min_separation_margin}
For any $k \in [K]$, suppose $\vw_k \notin \conv(\{\vw_j\}_{j\in[K]\setminus k})$, then 
\begin{equation}
     \min_{\vh\in\setS^{d-1}}  \max_{k' \neq k}\left\langle \vw_{k'} - \vw_k, \vh \right\rangle = - \dist(\vw_k,\{\vw_j\}_{j\in[K]\setminus k}).
\end{equation}
In addition, the optimal $\vh$ is given by 
\begin{equation}
    \vh = \mathcal{P}_{\setS^{d-1}} \left(\vw_k - \mathcal{P}_{ \{\vw_j\}_{j\in[K]\setminus k}}(\vw_k) \right)
\end{equation}
where $\mathcal{P}_\mathcal{W}(\vv) \doteq \argmin_{\vw \in \conv(\mathcal{W})}\{\|\vv - \vw\|_2 \}$ denotes the projection of $\vv$ on $\conv(\mathcal{W})$.
\end{lemma}

\begin{proof}
The proof follows from combining \Cref{thm:optimize-h-equivalent-form} and \Cref{thm:optimize-h-duality}. 
\end{proof}

\begin{lemma}\label{thm:optimize-h-equivalent-form}
Suppose $\vw_k \notin \conv(\{\vw_j\}_{j\in[K]\setminus k})$. Then
\begin{equation}\label{eq:optimize-h-equivalent-form}
     \min_{\vh\in\setS^{d-1}}  \max_{k' \neq k}\left\langle \vw_{k'} - \vw_k, \vh \right\rangle \equiv \min_{\|\vh\|_2 \le 1}  \max_{k' \neq k}\left\langle \vw_{k'} - \vw_k, \vh \right\rangle
\end{equation}
where $\equiv$ means the two problems are equivalent, i.e., have the same optimal solutions.
\end{lemma}
\begin{proof}[Proof of \Cref{thm:optimize-h-equivalent-form}]
    By the separating hyperplane theorem (e.g. \cite[Example 2.20]{boyd2004convex}), there exist a nonzero vector $\bar{\vh}$ and $b \in \RR$ such that $\max_{k' \ne k} \langle \vw_{k'}, \bar{\vh} \rangle < b$ and $\langle \vw_k, \bar{\vh} \rangle > b$, i.e., $\max_{k' \neq k}\left\langle \vw_{k'} - \vw_k, \bar{\vh} \right\rangle < 0$.
    Let $\vh^*$ be any optimal solution to the RHS of \eqref{eq:optimize-h-equivalent-form}. 
    Then, it holds that
    $$\max_{k' \neq k}\left\langle \vw_{k'} - \vw_k, \vh^* \right\rangle \le \max_{k' \neq k}\left\langle \vw_{k'} - \vw_k, \frac{\bar{\vh}}{\|\bar{\vh}\|_2} \right\rangle < 0.$$
    Hence, it must be the case that $\|\vh^*\|_2 = 1$; otherwise, taking $\vh = \vh^* / \|\vh^*\|_2$ gives lower objective for the RHS of \eqref{eq:optimize-h-equivalent-form}, contradicting the optimality of $\vh^*$.
\end{proof}

\begin{lemma}\label{thm:optimize-h-duality}
Suppose $\vw_k \notin \conv(\{\vw_j\}_{j\in[K]\setminus k})$. Consider the (primal) problem
\begin{equation}
    \min_{\|\vh\|_2 \le 1}  \max_{k' \neq k}\left\langle \vw_{k'} - \vw_k, \vh \right\rangle.
\end{equation}
Its dual problem is given by
\begin{equation}
    \max_{\vv \in \RR^d} -\|\w_k - \sum_{k' \ne k}v_{k'}\w_{k'}\|_2 ~~\st \sum_{k' \ne k} v_{k'} = 1, ~\text{and}~v_{k'}\ge 0, \forall k' \ne k.
\end{equation}
with zero duality gap. Moreover, for any primal optimal solution $\vh^*$ there is a dual optimal solution $\vv^*$ and they satisfy
\begin{equation}
    \vh^* = \frac{\vw_k - \sum_{k'}v_{k'}^*\vw_{k'}}{\|\vw_k - \sum_{k'}v_{k'}^*\vw_{k'}\|_2}.
\end{equation}
\end{lemma}
\begin{proof}[Proof of \Cref{thm:optimize-h-duality}]
    We rewrite the primal problem as
    \begin{equation}
        \min_{\|\vh\|_2 \le 1, \vp \in \RR^d}  \max_{k' \neq k} p_{k'} ~~\st~ p_{k'} = \left\langle \vw_{k'} - \vw_k, \vh \right\rangle.
    \end{equation}
    Introducing the dual variable $\vv \in \RR^d$, the Lagragian function is
    \begin{equation}
        \mathcal{L}(\h, \vp, \vv) = \max_{k' \neq k} p_{k'} -  \sum_{k' \ne k} v_{k'}(p_{k'} - \left\langle \vw_{k'} - \vw_k, \vh \right\rangle), ~~\|\vh\|_2 \le 1.
    \end{equation}
    We now derive the dual problem, defined as
    \begin{equation}\label{eq:prf-optimize-h-dual-derivation}
    \begin{split}
        \max_{\vv} \min_{\|\vh\|_2 \le 1, \vp \in \RR^d}  \max_{k' \neq k} \mathcal{L}(\h, \vp, \vv) 
        &= \max_{\vv} \left( \min_{\vp} \left(\max_{k' \neq k} p_{k'} - \sum_{k' \ne k} v_{k'}p_{k'}\right)
        + \min_{\|\h\|_2 \le 1} \left\langle \sum_{k' \ne k} v_{k'}\vw_{k'} - \vw_k, \vh \right\rangle \right)\\
        &= \max_{\vv} \min_{\|\h\|_2 \le 1} \left\langle \sum_{k' \ne k} v_{k'}\vw_{k'} - \vw_k, \vh \right\rangle ~~\st \sum_{k' \ne k} v_{k'} = 1, ~v_{k'}\ge 0 ~\forall k' \ne k\\
        &= \max_{\vv} -\|\w_k - \sum_{k' \ne k}v_{k'}\w_{k'}\|_2 ~~\st \sum_{k' \ne k} v_{k'} = 1, ~v_{k'}\ge 0 ~\forall k' \ne k.
    \end{split}
    \end{equation}
    In above, the second equality follows from the fact that the conjugate (see e.g. \cite{boyd2004convex}) of the max function is the indicator function of the probability simplex.
    The third equality uses the assumption that $\vw_k \notin \conv(\{\vw_j\}_{j\in[K]\setminus k})$, which implies that $\sum_{k' \ne k} v_{k'}\vw_{k'} - \vw_k \ne 0$ under the simplex constraint of $\vv$, hence the optimal $\vh$ to the optimization in the second line can be easily obtained as $\vh = \frac{\vw_k - \sum_{k'}v_{k'}\vw_{k'}}{\|\vw_k - \sum_{k'}v_{k'}\vw_{k'}\|_2}$.

    Finally, the rest of the claims hold as the primal problem is convex with the Slater's condition satisfied.
\end{proof}

\subsection{Proof of \Cref{thm:NC2}}

Here we prove the following result which is a stronger version of \Cref{thm:NC2}.

\begin{theorem}\label{thm:NC2-stronger-version}
Let $(\mW^\star,\mH^\star)$ be an optimal solution to  \eqref{def:hardmax}. Then, it holds that $\mW^\star$ is a Softmax Code, \ie,
\begin{equation}
    \mW^\star \in \argmax\limits_{\mW\in\OB(d,K)} \rho_\onevsrest(\mW).
\end{equation}
Conversely, let $\mW^\text{SC}$ be any Softmax Code. Then, there exists a $\mH^\text{SC}$ such that $(\mW^\text{SC},\mH^\text{SC})$ is an optimal solution to \eqref{def:hardmax}.
\end{theorem}

\begin{proof}
The proof is divided into two parts.

\flushleft\textbf{Any optimal solution to \eqref{def:hardmax} is a Softmax Code.}
Our proof is based on providing a lower bound on the objective $\mathcal{L}_{\text{HardMax}}(\mW, \mH)$. 
We distinguish two cases in deriving the lower bound.
\begin{itemize}
    \item $\mW$ has distinct columns. In this case we use the following bound:
    \begin{multline}\label{eq:prf-hardmax-lower-bound}
        \mathcal{L}_{\text{HardMax}}(\mW, \mH) 
        = \max_{k \in [K]} \max_{i \in [n]} \max_{k' \ne k} \langle \w_{k'} - \w_{k} , \h_{k,i} \rangle \\
        \ge \max_{k \in [K]} \max_{i \in [n]} \min_{\bar{\vh}_{k, i} \in \setS^{d-1}} \max_{k' \ne k} \langle \w_{k'} - \w_{k} , \bar{\vh}_{k,i} \rangle
        = -\min_{k \in [K]} \dist(\vw_k,\{\vw_j\}_{j\in[K]\setminus k}).
    \end{multline}
    In above, the first equality follows directly from definition of the HardMax function. 
    The inequality follows trivially from the property of the $\min$ operator. 
    The last equality follows from \Cref{thm:max_min_separation_margin}, which requires that $\mW$ has distinct columns.
    Continuing on the rightmost term in \eqref{eq:prf-hardmax-lower-bound}, we have
    \begin{equation}\label{eq:prf-hardmax-lower-bound-continued}
        -\min_{k \in [K]} \dist(\vw_k,\{\vw_j\}_{j\in[K]\setminus k}) 
        = -\rho_\onevsrest(\mW) 
        \ge -\rho_\onevsrest(\mW^\text{SC}),
    \end{equation}
    where $\mW^\text{SC}$ is any Softmax Code. In above, the equality follows from the definition of the operator $\rho_\onevsrest()$, and the inequality follows from the definition of the Softmax Code. 
    In particular, by defining $\widehat{\mW} = \mW^\text{SC}$ and $\widehat{\mH}$ as
    \begin{equation}
        \widehat{\vh}_{k, i} = 
        \frac{\widehat{\vw}_k - \proj(\widehat{\vw}_k,\{\widehat{\vw}_j\}_{j\in[K]\setminus k})}{\|\widehat{\vw}_k - \proj(\widehat{\vw}_k,\{\widehat{\vw}_j\}_{j\in[K]\setminus k})\|_2},
    \end{equation}
    all inequalities in \eqref{eq:prf-hardmax-lower-bound} and \eqref{eq:prf-hardmax-lower-bound-continued} holds with equality by taking $\mW = \widehat{\mW}$ and $\mH = \widehat{\mH} $, at which we have that $\mathcal{L}_{\text{HardMax}}(\widehat{\mW} , \widehat{\mH} ) = -\rho_\onevsrest(\mW^\text{SC})$.
    
    \item $\mW$ does not have distinct columns. Hence, there exists $k_1$, $k_2$ such that $k_1 \ne k_2$ but $\vw_{k_1} = \vw_{k_2}$. We have
    \begin{equation}
        \mathcal{L}_{\text{HardMax}}(\mW, \mH) 
        = \max_{k \in [K]} \max_{i \in [n]} \max_{k' \ne k} \langle \w_{k'} - \w_{k} , \h_{k,i} \rangle 
        \ge \max_{i \in [n]} \langle \vw_{k_2} - \vw_{k_1}, \vh_{k_1, i} \rangle = 0.
    \end{equation}
\end{itemize}

Combining the above two cases, and by noting that $ -\rho_\onevsrest(\mW^\text{SC}) < 0$, we have that $(\widehat{\mW} , \widehat{\mH} )$ is an optimal solution to the HardMax problem in \eqref{def:hardmax}. 
Moreover, since $(\mW^*, \mH^*)$ is an optimal solution to the HardMax problem, it must attain the lower bound, \ie, 
\begin{equation}\label{eq:prf-hardmax-optimal-objective}
    \mathcal{L}_{\text{HardMax}}(\mW^* , \mH^* ) = -\rho_\onevsrest(\mW^\text{SC}).
\end{equation} 
Hence, $\mW^*$ has to attain the equality in \eqref{eq:prf-hardmax-lower-bound-continued}. 
By definition, this means that $\mW^*$ is a Softmax Code, which concludes the proof of this part.

\flushleft\textbf{From any Softmax Code we can construct an optimal solution to \eqref{def:hardmax}.}
Let
\begin{equation}
    \mW^\text{SC} \in \argmax\limits_{\mW\in\OB(d,K)} \rho_\onevsrest(\mW)
\end{equation}
be any Softmax Code. 
Moreover, define $\mH^\text{SC}$ to be such that 
\begin{equation}\label{eq:prf-nc2-constructed-h}
\vh_{k, i}^\text{SC} = \argmin_{\vh_k \in \setS^{d-1}} \max_{k' \ne k} \langle \vw_{k'}^\text{SC} - \vw_{k}^\text{SC}, \vh_{k} \rangle, \forall k \in [K], \forall i \in [n].
\end{equation}
Note that the following result holds which will be used in the subsequent proof:
\begin{equation}\label{eq:prf-nc2-softmaxcode-property}
\begin{aligned}
    \mW^\text{SC} 
    &\in \argmax_{\mW\in\OB(d,K)} \min_{k} \dist\!\Big(\vw_k, \{\vw_j\}_{j\in[K]\setminus k}\Big) && \text{(Definition of Softmax Code)} \\
    &\in \argmin_{\mW\in\OB(d,K)} \max_k \min_{\vh_k \in\setS^{d-1}} \max_{k' \ne k} \langle \vw_{k'} - \vw_k, \vh_k \rangle. && \text{(\Cref{thm:max_min_separation_margin})} \\
\end{aligned}
\end{equation}
For any $(\widehat{\mW}, \widehat{\mH})$, we have
\begin{equation}
\begin{aligned}
    \mathcal{L}_{\text{HardMax}}(\widehat{\mW}, \widehat{\mH}) 
    &= \max_{k \in [K]} \max_{i \in [n]} \max_{k' \ne k} \langle \widehat{\vw}_{k'} - \widehat{\vw}_k, \widehat{\vh}_{k, i} \rangle && \text{(Definition of $\mathcal{L}_\text{HardMax}$)} \\
    &\ge \max_{k \in [K]} \min_{\vh_k \in\setS^{d-1}} \max_{k' \ne k} \langle \widehat{\vw}_{k'} - \widehat{\vw}_k, \vh_k \rangle \\
    &\ge \max_{k \in [K]} \min_{\vh_k \in\setS^{d-1}} \max_{k' \ne k} \langle {\vw}^\text{SC}_{k'} - {\vw}^\text{SC}_k, \vh_k \rangle && \text{(Eq. \eqref{eq:prf-nc2-softmaxcode-property})} \\
    &= \max_{k \in [K]} \max_{i \in [n]} \max_{k' \ne k} \langle {\vw}^\text{SC}_{k'} - {\vw}^\text{SC}_k, {\vh}_{k, i}^\text{SC} \rangle && \text{(Eq. \eqref{eq:prf-nc2-constructed-h})} \\
    &= \mathcal{L}_{\text{HardMax}}(\mW ^\text{SC}, \mH^\text{SC}). && \text{(Definition of $\mathcal{L}_\text{HardMax}$)} 
\end{aligned}
\end{equation}
This implies that $(\mW ^\text{SC}, \mH^\text{SC})$ is an optimal solution to the HardMax problem in \eqref{def:hardmax}, which concludes the proof of this part.

\end{proof}

\subsection{Proof of \Cref{thm:special-case-NC3}}



\begin{theorem}
For any positive integers $K$ and $d$, let $\mW^\star \in \OB(d,K)$ be a Softmax Code. 
Then, 
\begin{itemize}[leftmargin=0.22in,topsep=0.2em,itemsep=0.11em]
    \item $d=2$: $\{\vw_k^\star\}$ is uniformly distributed on the unit circle, i.e., $\{\vw_k^\star\} = \{\big(\cos(\frac{2\pi k}{K} + \alpha), \sin(\frac{2\pi k}{K} + \alpha)\big)\}$ for some $\alpha$;
    \item $K\le d+1$: $\{\vw_k^\star\}$ forms a simplex ETF, i.e., $\W^\star = \sqrt{\frac{K}{K-1}}\P(\I_K - \frac{1}{K}\1_K \1_K^\top)$ for some orthonomal $\P \in \RR^{d\times K}$;
    \item $d+1< K\le 2d $: $\min_{k} \dist(\vw_k^\star,\{\vw_j^\star\}_{j\in[K]\setminus k}) =1$ which can be achieved when $\{\vw_k^\star\}$ are a subset of vertices of a cross-polytope; 
    \item $K\rightarrow\infty$: $\{\vw_k^\star\}$ are uniformly distributed on the unite sphere  $\mathbb{S}^{d-1}$;
\end{itemize}
\end{theorem}

\begin{proof} We prove the results case by case as follows.

\begin{itemize}[leftmargin=0.22in,topsep=0.2em,itemsep=0.11em]
    \item \textbf{$\boldsymbol{d=2}$: $\{\vw_k^\star\}$ are uniformly distributed on the unite sphere of $\mathbb{S}^1$.}

    Denote $\vw_k^\star = [\cos \alpha_k^\star, \sin \alpha_k^\star], \forall k \in [K]$. Without loss of generality we may assume that $0 < \alpha_1^\star < \alpha_2^\star < \ldots < \alpha_K^\star \le 2\pi$. Define 
    \begin{equation}\label{eq:prf-d=2-def-theta}
        \theta_k^\star = 
        \begin{cases}
            \alpha_{k+1}^\star - \alpha_k^\star, & ~~\forall k \in [K-1]\\
            \alpha_1^\star - \alpha_K^\star + 2\pi, & ~~k = K.
        \end{cases}
    \end{equation}
    For convenience, we also define $\alpha^\star_{K+1} \doteq \alpha^\star_1$ and $\theta^\star_{K+1} \doteq \theta^\star_1$. 

    Geometrically, $\theta_k^\star$ is the angular distance between $\alpha_k^\star$ and $\alpha_{k+1}^*$. 
    Moreover, by summing up all terms in \eqref{eq:prf-d=2-def-theta} over $k \in [K]$ we have
    \begin{equation}\label{eq:prf-d=2-sum-theta}
        \sum_{k \in [K]} \theta_k^\star = 2\pi.
    \end{equation}
    To prove the theorem we only need to show that $\theta_k^\star = \frac{2\pi}{K}$ for all $k \in [K]$, which implies that $\{\vw_k^\star\}$ are uniformly distributed on the unit circle.

    We start by noting that the following result holds:
    \begin{equation}\label{eq:prf-d=2-theta_k+theta_k+1}
        \theta_k^\star + \theta_{k+1}^\star \ge 2 \times \frac{2\pi}{K}, ~~\forall k \in [K].
    \end{equation}
    To see why, let $\{\widehat{\vw}_k = [\cos \widehat{\alpha}_k, \sin \widehat{\alpha}_k]\}_{k \in [K]}$ with $\widehat{\alpha}_k = \frac{k \times 2\pi}{K}, \forall k \in [K]$ be a collection of points on the unit circle that is distributed uniformly. 
    Moreover, define $\{\widehat{\theta}_k\}_{k \in [K]}$ as
    \begin{equation}
        \widehat{\theta}_k = 
        \begin{cases}
            \widehat{\alpha}_{k+1} - \widehat{\alpha}_k, & ~~\forall k \in [K-1]\\
            \widehat{\alpha}_1 - \widehat{\alpha}_K + 2\pi, & ~~k = K.
        \end{cases}
    \end{equation}
    Since $\mW^\star$ is a Softmax Code, we have
    \begin{equation}
        \rho_\onevsrest(\mW^\star) \ge \rho_\onevsrest(\widehat{\mW}),
    \end{equation}
    which implies, using the definition of $\rho_\onevsrest()$,
    \begin{equation}\label{eq:prf-d=2-distance-relation}
        \min_{k \in [K]} \dist(\vw_k^\star, \{\vw_j^\star\}_{j \in [K] \setminus k}) \ge \min_{k \in [K]} \dist(\widehat{\vw}_k, \{\widehat{\vw}_j\}_{j \in [K] \setminus k}).
    \end{equation}
    If there exists a $\bar{k} \in [K]$ such that $\theta_{\bar{k}}^\star + \theta_{\bar{k}+1}^\star < 2 \times \frac{2\pi}{K}$, by noting that $\widehat{\theta}_{\bar{k}} + \widehat{\theta}_{\bar{k}+1} = 2 \times \frac{2\pi}{K}$, it is easy to see geometrically that 
    $$\dist(\vw_{\bar{k}}^\star, \{\vw_j^\star\}_{j \in [K] \setminus {\bar{k}}}) < \dist(\widehat{\vw}_{\bar{k}}, \{\widehat{\vw}_j\}_{j \in [K] \setminus \bar{k}}) = \min_{k \in [K]} \dist(\widehat{\vw}_k, \{\widehat{\vw}_j\}_{j \in [K] \setminus k}),$$
    where the last equality follows from the fact that $\{\widehat{\vw}_k\}_{k \in [K]}$ are uniformly distributed. 
    This result contradicts \eqref{eq:prf-d=2-distance-relation}, which implies that \eqref{eq:prf-d=2-theta_k+theta_k+1} holds.

    Taking the summation on both sizes of \eqref{eq:prf-d=2-theta_k+theta_k+1} over all $k \in [K]$ and divide both sides by $2$, we obtain
    \begin{equation}
        \sum_{k \in [K]} \theta^*_k \ge 2\pi.
    \end{equation}
    Comparing this with \eqref{eq:prf-d=2-sum-theta}, we obtain that the inequality in \eqref{eq:prf-d=2-theta_k+theta_k+1} holds with equality for all $k \in [K]$, that is, 
    \begin{equation}\label{eq:prf-d=2-theta_k+theta_k+1-equal}
        \theta_k^\star + \theta_{k+1}^\star = 2 \times \frac{2\pi}{K}, ~~\forall k \in [K].
    \end{equation}
    If there exists a $\bar{k} \in [K]$ such that $\theta_{\bar{k}}^\star \ne \theta_{\bar{k}+1}^\star$, then it is easy to see geometrically that 
    $$\dist(\vw_{\bar{k}}^\star, \{\vw_j^\star\}_{j \in [K] \setminus {\bar{k}}}) < \dist(\widehat{\vw}_{\bar{k}}, \{\widehat{\vw}_j\}_{j \in [K] \setminus \bar{k}}) = \min_{k \in [K]} \dist(\widehat{\vw}_k, \{\widehat{\vw}_j\}_{j \in [K] \setminus k}),$$
    which contradicts \eqref{eq:prf-d=2-distance-relation}. 
    Hence, it follows that $\theta_k^\star = \frac{2\pi}{K}$ for all $k \in [K]$.

    \item \textbf{$\boldsymbol{K\le d+1}$: $\{\vw_k^\star\}$ forms a simplex ETF.}

    We first consider optimal configuration of $K$ unit-length vectors $\u_1,\ldots,\u_K$. Note that
\begin{align*}
& 0 \le \norm{\sum_{k=1}^K \u_k}{2}^2 = \sum_k \sum_{k'} \langle \u_k, \u_{k'} \rangle \le K + K(K-1) \max_{k\neq k'} \langle \u_k, \u_{k'} \rangle,
\end{align*}
where the first inequality  achieves equality only when $\sum_{k=1}^K \u_k = 0$ and the second inequality becomes equality only when $\langle \u_k, \u_{k'} \rangle = -\frac{1}{K-1}$ for any $k\neq k'$. These two conditions mean that $\u_1,\ldots,\u_K$ form a simplex ETF. The above equation further impleis that 
\begin{align}
\max_{k\neq k'} \langle \u_k, \u_{k'} \rangle \ge -\frac{1}{K-1}, \ \forall \u_1,\ldots,\u_K \in \mathbb{S}^{d-1},
\label{eq:ETF-requirement}\end{align}
and the equality holds only when $\u_1,\ldots,\u_K$ form a simplex ETF.

We will also need the following result:
\begin{align}
\frac{1}{2}\sum_{k\neq k'}\|\vw_k - \vw_{k'}\|^2 = K^2  - \big\|\sum_{k} \u_k\big\|^2 \le K^2,
\label{eq:proof-ETF-sum}\end{align}
where the last inequality becomes equality when $\sum_{k} \u_k = 0$. 

We now prove the form of the optimal Softmax Code for $K\le d+1$. Noting the equivalence between Softmax Code and the HardMax problem as proved in \Cref{thm:NC2}, we will analyze the HardMax problem for this case. Specifically, note that
\begin{align*}
& K(K-1) \max_{k \neq k'} \langle \w_{k'} - \w_{k} , \h_k \rangle \ge \sum_{k\neq k'} \langle \w_{k'} - \w_{k} , \h_k \rangle = \frac{1}{2} \sum_{k\neq k'} \langle \w_{k'} - \w_{k} , \h_k - \h_{k'} \rangle \\ 
& \ge  -\frac{1}{4} \sum_{k\neq k'}\|\vw_k - \vw_{k'}\|^2 - \frac{1}{4} \sum_{k\neq k'}\|\vh_k - \vh_{k'}\|^2 \ge - K^2,
\end{align*}
where the first inequality achieves equality only when $\langle \w_{k'} - \w_{k} , \h_k \rangle = \langle \w_{j'} - \w_{j} , \h_j\rangle$ for any $k'\neq k, j'\neq j$, the second inequality follows from the Cauchy–Schwarz inequality and acheives inequality only when $\vw_{k'} - \vw_k = \vh_{k'} - \vh_k$ for any $k'\neq k$, and the third inequality follows from \eqref{eq:proof-ETF-sum} and achieves equality only when $\sum_k \vw_k = \sum_k \vh_k = 0$. Assuming all these conditions hold, then  $
 \langle \w_{k'} - \w_{k} , \h_k \rangle = - \frac{K}{K-1}$, which together with the requirement $\vw_{k'} - \vw_k = \vh_{k'} - \vh_k$ implies that
\[
\langle \h_{k'} - \h_{k} , \h_k \rangle = - \frac{K}{K-1}, \quad  \Rightarrow \quad \langle \h_{k'} , \h_k \rangle = - \frac{1}{K-1}, \forall k\neq k',
\]
which holds only when $\mH$ forms a simplex ETF according to the derivation for \eqref{eq:ETF-requirement}. Using the condition  $\vw_{k'} - \vw_k = \vh_{k'} - \vh_k$ which indicates $\langle \w_{k'} , \w_k \rangle = \langle \h_{k'} , \h_k \rangle$, we can obtain that $\mW$ is also a simplex ETF, which completes the proof. 

\item \textbf{$\boldsymbol{d+1< K\le 2d }$: $\rho_\onevsrest(\mW^\star) =1$ which can be achieved when $\{\vw_k^\star\}$ are some vertices of a cross-polytope.}

We first present and prove the following result that establishes an upper bound for $\rho_{\onevsrest}(\mW)$ when $K\ge d+2$.
\begin{lemma}
Suppose $K \ge d+2$, then for any $\mW \in \OB(d,K)$, it holds that $\rho_{\onevsrest}(\mW) \le 1$, with equality only if $\ \0 \in \conv(\mW)$,  where $\conv(\mW)$ is the convex hull of $ \{\vw_j\}_{j\in[K]}$.
\label{lemma:K>d+2}
\end{lemma}

\begin{proof}[Proof of  Lemma \ref{lemma:K>d+2}]
Let $\v$ denote the (unique) point in $\conv(\mW)$ of minimum $l_2$ norm. By Carathéodory's theorem, $\v$ resides in the convex hull of $d+1$ of the points in $\mW$. Since $K \ge d+2$ by assumption, there exists $k \in [K]$ such that $\v \in  \conv(\{\vw_j\}_{j\in[K] \setminus k})$ where $\v$ is the projection of $\w_k$ to $\conv(\{\vw_j\}_{j\in[K] \setminus k})$. The result follows by noting the following two cases.
\begin{itemize}
[leftmargin=0.22in,topsep=0.2em,itemsep=0.11em]
    \item \textbf{Case I:} $\v = \0.$ Then $\rho(\mW) \le \dist(\vw_{k}, \conv(\{\vw_j\}_{j\in[K] \setminus k})) \le \|\vw_{k} - \v\|_2 = \|\vw_{k} \| = 1.$
    \item \textbf{Case II:} $\v \neq \0.$ Since $\v$ is the projection of $\0$ onto $\conv(\mW)$, the supporting hyperplane at $\v$ gives $\langle \x, \v \rangle \ge \|\v\|_{2}^{2}$ for every $\x \in \conv(\mW)$. In particular, taking $\x = \vw_k$ implies
    \begin{align*}
    \|\vw_k - \v\|^2 = \|\vw_k\|^2-2\langle \vw_k,\v\rangle + \|\v\|^2  \le 1 - \|\v\|^2 < 1,
    \end{align*}
    and so
    \begin{align*}
    \rho_{\onevsrest}(\mW) \le \dist(\vw_k, \conv(\{\vw_j\}_{j\in[K] \setminus k})) \le \|\vw_k - \v\| < 1.
    \end{align*}
\end{itemize}
\end{proof}

According to Lemma \ref{lemma:K>d+2}, when $K \ge d+2$, for any $\mW \in \OB(d,K)$, it holds that $\rho_{\onevsrest}(\mW) \le 1$. Moreover, when $d+2\le K \le 2d $, we can verify that $\mW^\star = \{\vw_k^\star\}$ as some vertices of a cross-polytope satisfies $\rho_{\onevsrest}(\mW^\star) = 1$. Hence, such a $\mW^\star$ is a Softmax Code. 

In addition, since $\mW^\star = \{\vw_k^\star\}$ as some vertices of a cross-polytope and $K\ge d+2$, we have $0\in \conv(\{\vw_j^\star\}_{j\in[K]\k})$ for any $k$, and hence it also holds that $\dist(\vw_k^\star,\{\vw_j^\star\}_{j\in[K]\k}) = 1$. Thus, there is no rattler for this case.

\end{itemize}
\end{proof}

\subsection{Proof of \Cref{thm:NC1}}

\begin{theorem}[\GNC1]
Let $(\mW^\star,\mH^\star)$ be an optimal solution to  \eqref{def:hardmax}. For all $k$ that is not a rattler of $\mW^\star$, it holds that
\begin{equation}
    \overline{\vh}_{k}^\star \doteq \vh_{k,1}^\star=\cdots=\vh_{k,n}^\star = \mathcal{P}_{\setS^{d-1}} \left(\vw_k^\star - \mathcal{P}_{ \{\vw_j^\star\}_{j\in[K]\setminus k}}(\vw_k^\star) \right).
\end{equation}
\end{theorem}

\begin{proof}
Let $\bar{k}$ be any non-rattler of $\mW^*$. We have
\begin{equation}
\begin{aligned}
    -\rho_\onevsrest(\mW^*) 
    &= -\min_{k \in [K]} \dist(\vw_k^*,\{\vw_j^*\}_{j\in[K]\setminus k}) && \text{(Definition of $\rho_\onevsrest())$}\\
    &= -\dist(\vw_{\bar{k}}^*,\{\vw_j^*\}_{j\in[K]\setminus {\bar{k}}})  && \text{(Definition of rattler)}\\
    &= \min_{\vh \in \setS^{d-1}} \max_{k' \ne \bar{k}} \langle \vw_{k'}^* - \vw_{\bar{k}}^*, \vh \rangle        && \text{(\Cref{thm:max_min_separation_margin})}\\
    &\le \max_{i \in [n]} \max_{k' \ne \bar{k}} \langle \vw_{k'}^* - \vw_{\bar{k}}^*, \vh^*_{\bar{k}, i} \rangle && \text{(Property of $\max$)}\\
    &\le \max_{i \in [n]} \max_{k \in [K]} \max_{k' \ne k} \langle \vw_{k'}^* - \vw_{k}^*, \vh^*_{k, i} \rangle  && \text{(Property of $\max$)}\\
    &=\mathcal{L}_{\text{HardMax}}(\mW^* , \mH^* ) && \text{(Definition of $\mathcal{L}_\text{HardMax}$)} \\
    &= -\rho_\onevsrest(\mW^\text{SC})             && \text{(Eq. \eqref{eq:prf-hardmax-optimal-objective})} \\
    &= -\rho_\onevsrest(\mW^*)             && \text{(\Cref{thm:NC2})}
\end{aligned}
\end{equation}
Since the first and last expressions are identical, all inequalities holds with equality. Hence
\begin{equation}
    \min_{\vh \in \setS^{d-1}} \max_{k' \ne \bar{k}} \langle \vw_{k'}^* - \vw_{\bar{k}}^*, \vh \rangle   
    = \max_{i \in [n]} \max_{k' \ne \bar{k}} \langle \vw_{k'}^* - \vw_{\bar{k}}^*, \vh^*_{\bar{k}, i} \rangle.
\end{equation}
By \Cref{thm:max_min_separation_margin}, the optimal $\vh$ to the optimization problem on the left is unique. 
Hence, all $\vh^*_{\bar{k}, i}, i \in [n]$ must be equal. 
This concludes the proof.

\end{proof}

\subsection{Proof of \Cref{thm:NC3}}

\begin{proof}
$(\implies)$ Assume that any $(\mW^\star, \mH^\star) \in \argmin_{\mW\in\OB(d,K),\mH\in\OB(d,nK)}  \mathcal{L}_{\text{HardMax}}(\mW, \mH)$ satisfies $\vh_{k, i}^\star = \vw_k^*, \forall i \in [n], \forall k \in [K]$. We show that the Tammes problem and Softmax code are equivalent.
This can be established trivially from the following two claims, namely, 
\begin{equation}\label{eq:prf-nc3-equiv-hardmax-onevsrest}
    \argmin_{\mW\in\OB(d,K)} \min_{\mH\in\OB(d,nK)}  \mathcal{L}_{\text{HardMax}}(\mW, \mH) = \argmax_{\mW\in\OB(d,K)} \rho_\onevsrest(\mW),
\end{equation}
and 
\begin{equation}\label{eq:prf-nc3-equiv-hardmax-onevsone} 
    \argmin_{\mW\in\OB(d,K)} \min_{\mH\in\OB(d,nK)}  \mathcal{L}_{\text{HardMax}}(\mW, \mH) = \argmax_{\mW\in\OB(d,K)} \rho_\onevsone(\mW).
\end{equation}
In the rest of the proof we show that the claims \eqref{eq:prf-nc3-equiv-hardmax-onevsrest} and\eqref{eq:prf-nc3-equiv-hardmax-onevsone} hold true.  

We first establish \eqref{eq:prf-nc3-equiv-hardmax-onevsrest}. From \Cref{thm:NC2}, any solutions to the HardMax problem is a Softmax Code, \ie,
\begin{equation} 
    \argmin_{\mW\in\OB(d,K)} \min_{\mH\in\OB(d,nK)}  \mathcal{L}_{\text{HardMax}}(\mW, \mH) \subseteq \argmax_{\mW\in\OB(d,K)} \rho_\onevsrest(\mW).
\end{equation}
Conversely, from \Cref{thm:NC2-stronger-version}, any Softmax Code must also be a solution to the HardMax problem, \ie, 
\begin{equation} 
    \argmax_{\mW\in\OB(d,K)} \rho_\onevsrest(\mW) \subseteq \argmin_{\mW\in\OB(d,K)} \min_{\mH\in\OB(d,nK)}  \mathcal{L}_{\text{HardMax}}(\mW, \mH) 
\end{equation}
Combining the above two relations we readily obtain \eqref{eq:prf-nc3-equiv-hardmax-onevsrest}. 

We now establish \eqref{eq:prf-nc3-equiv-hardmax-onevsone}.
Let $(\mW^\star, \mH^\star) \in \argmin_{\mW\in\OB(d,K),\mH\in\OB(d,nK)}  \mathcal{L}_{\text{HardMax}}(\mW, \mH)$ be any solution to the HardMax problem, which by our assumption satisfies $\vh_{k, i}^\star = \vw_k^*, \forall i \in [n], \forall k \in [K]$. 
Using this condition and plugging in the definition of the HardMax function we obtain
\begin{equation}\label{eq:prf-gnc3-implies-hardmax-tammes}
\begin{split}
    \mW^* \in \argmin_{\mW\in\OB(d,K)} \max_{k \in [K]}  \max_{k' \ne k} \langle \w_{k'} - \w_{k} , \w_{k} \rangle = \argmin_{\mW\in\OB(d,K)} \max_{k \in [K]}\max_{k' \ne k} \langle \w_{k'}, \w_{k} \rangle = \argmax_{\mW\in\OB(d,K)} \rho_\onevsone(\mW),
\end{split}
\end{equation}
where the first equality uses the fact that $\w_k$ has unit $\ell_2$ norm for all $k \in [K]$, and the second equality follows trivially from the definition of $\rho_\onevsone()$. 
This implies
\begin{equation}\label{eq:prf-nc3-hardmax-is-subset-of-tammes}
    \argmin_{\mW\in\OB(d,K)} \min_{\mH\in\OB(d,nK)}  \mathcal{L}_{\text{HardMax}}(\mW, \mH) \subseteq \argmax_{\mW\in\OB(d,K)} \rho_\onevsone(\mW).
\end{equation}
We argue that the converse of the set inclusion above also holds. To see that, let
$$\mW^\text{Tammes} \in  \argmax_{\mW\in\OB(d,K)} \rho_\onevsone(\mW)$$ 
be any solution to the Tammes problem, and let $\mH ^\text{Tammes} \in\OB(d,nK)$ be such that $\vh_{k, i}^\text{Tammes} = \vw_k^\text{Tammes}, \forall i \in [n], \forall k \in [K]$. We have
\begin{equation}
\begin{aligned}
    \mathcal{L}_{\text{HardMax}}(\mW^\text{Tammes}, \mH^\text{Tammes})
    &= \max_{k \in [K]} \max_{i \in [n]} \max_{k' \ne k} \langle \vw_{k'}^\text{Tammes} - \vw_{k}^\text{Tammes} , \vh_{k,i}^\text{Tammes} \rangle  && \text{(Eq. \eqref{def:hardmax})} \\
    &= \max_{k \in [K]} \max_{k' \ne k} \langle \vw_{k'}^\text{Tammes} - \w_{k}^\text{Tammes} , \vw_k^\text{Tammes}\rangle                         && \text{(Using $\vh_{k, i}^\text{Tammes} = \vw_k^\text{Tammes}$)}\\
    &= \max_{k \in [K]} \max_{k' \ne k} \langle \vw_{k'}^\text{Tammes}, \vw_k^\text{Tammes}\rangle - 1                                 && \text{(Using $\|\vw_k^\text{Tammes}\|_2 = 1$)}\\
    &= \min_{\mW\in\OB(d,K)} \max_{k \in [K]} \max_{k' \ne k}  \langle \vw_{k'}, \vw_k\rangle - 1              && \text{(Definition of $\mW^\text{Tammes}$)}\\
    &= \max_{k \in [K]} \max_{k' \ne k} \langle \vw_{k'}^\star, \vw_k^\star \rangle - 1                        && \text{(Eq. \eqref{eq:prf-gnc3-implies-hardmax-tammes})}\\
    &= \max_{k \in [K]} \max_{k' \ne k} \langle \vw_{k'}^\star - \w_{k}^\star , \vw_k^\star\rangle             && \text{(Using $\|\vw_k^\star\|_2 = 1$)}\\
    &= \max_{k \in [K]} \max_{i \in [n]} \max_{k' \ne k} \langle \vw_{k'}^\star - \vw_{k}^\star , \vh_{k,i}^\star \rangle  && \text{(Using $\vh_{k, i}^\star = \vw_k^\star$)} \\
    &= \mathcal{L}_{\text{HardMax}}(\mW^\star, \mH^\star).\\
\end{aligned}
\end{equation}
This implies that $(\mW^\text{Tammes}, \mH^\text{Tammes})$ is also a solution to the HardMax problem. 
Hence,
\begin{equation}\label{eq:prf-nc3-tammes-is-subset-of-hardmax}
    \argmax_{\mW\in\OB(d,K)} \rho_\onevsone(\mW) \subseteq \argmin_{\mW\in\OB(d,K)} \min_{\mH\in\OB(d,nK)}  \mathcal{L}_{\text{HardMax}}(\mW, \mH).
\end{equation}
Combining \eqref{eq:prf-nc3-hardmax-is-subset-of-tammes} and \eqref{eq:prf-nc3-tammes-is-subset-of-hardmax} we obtain \eqref{eq:prf-nc3-equiv-hardmax-onevsone}.

$(\impliedby)$ 
Assume that the Tammes problem and the Softmax Codes are equivalent. Take any optimal solution $(\mW^\star,\mH^\star)$ to  \eqref{def:hardmax}, i.e., 
\begin{multline}\label{eq:prf-nc3-def-wh-star}
    (\mW^\star,\mH^\star) 
    = \argmax_{\mW\in\OB(d,K), \mH\in\OB(d,nK)}\mathcal{L}_{\text{HardMax}}(\mW, \mH) \\
    = \argmax_{\mW\in\OB(d,K), \mH\in\OB(d,nK)} \max_{k \in [K]} \max_{i \in [n]} \max_{k' \ne k} \langle \w_{k'} - \w_{k} , \h_{k,i} \rangle.
\end{multline}
We show that $\vh_{k, i}^\star = \vw_k^*, \forall i \in [n], \forall k \in [K]$.

From \Cref{thm:NC2}, $\mW^\star$ is a Softmax Code, \ie,
\begin{equation}
    \mW^* = \argmax_{\mW\in\OB(d,K)} \min_{k \in [K]} \dist\!\Big(\vw_k^\star, \{\vw_j^\star\}_{j\in[K]\setminus k}\Big) = 
    \argmin_{\mW\in\OB(d,K)} \max_{k \in [K]} \min_{\vh_k \in \setS^{d-1}} \max_{k' \ne k} \langle \vw_{k'} - \vw_k, \vh_k \rangle,
\end{equation}
where the second equality follows from \Cref{thm:max_min_separation_margin}.
By the assumption that Softmax Code has no rattler, we know that
\begin{equation}\label{eq:prf-nc3-sc-no-rattler-implication}
    \min_{\vh_k \in \setS^{d-1}} \max_{k' \ne k} \langle \vw_{k'}^\star - \vw_k ^\star, \vh_k \rangle
\end{equation}
is independent of $k \in [K]$. 
We now use this result to show that, for any $\bar{k} \in [K]$ and $\bar{i} \in [n]$ the following result holds:
\begin{equation}\label{eq:prf-nc3-sc-optimal-h}
    \vh_{\bar{k}, \bar{i}}^* \in \argmin_{\vh_{\bar{k}} \in \setS^{d-1}} \max_{k' \ne \bar{k}} \langle \vw_{k'}^\star - \vw_{\bar{k}}^\star, \vh_{\bar{k}} \rangle.
\end{equation}
To prove \eqref{eq:prf-nc3-sc-optimal-h} by constructing a contradition, we assume that \eqref{eq:prf-nc3-sc-optimal-h} does not hold, \ie,
\begin{equation}\label{eq:prf-nc3-sc-optimal-h-opposite}
    \vh_{\bar{k}, \bar{i}}^* \notin \argmin_{\vh_{\bar{k}} \in \setS^{d-1}} \max_{k' \ne \bar{k}} \langle \vw_{k'}^\star - \vw_{\bar{k}}^\star, \vh_{\bar{k}} \rangle.
\end{equation}
Using \eqref{eq:prf-nc3-sc-optimal-h-opposite} we have
\begin{equation}
\begin{aligned}
    \mathcal{L}_{\text{HardMax}}(\mW^\star, \mH^\star) 
    &=  \max_{k \in [K]} \max_{i \in [n]} \max_{k' \ne k} \langle \vw_{k'}^\star - \vw_{k}^\star , \vh_{k,i}^\star \rangle && \text{(Definition of $\mathcal{L}_{\text{HardMax}}$)}\\
    &\ge  \max_{k' \ne \bar{k}} \langle \vw_{k'}^\star - \vw_{\bar{k}}^\star , \vh_{\bar{k}, \bar{i}}^\star \rangle  && \text{(Property of $\max$)}\\
    &> \min_{\vh_{\bar{k}} \in \setS^{d-1}}\max_{k' \ne \bar{k}} \langle \vw_{k'}^\star - \vw_{\bar{k}}^\star , \vh_{\bar{k}} \rangle && \text{(Eq. \eqref{eq:prf-nc3-sc-optimal-h-opposite})}\\
    &= \min_{\vh_{k} \in \setS^{d-1}}\max_{k' \ne k} \langle \vw_{k'}^\star - \vw_{k}^\star , \vh_{k} \rangle, \forall k \in [K] && \text{(Eq. \eqref{eq:prf-nc3-sc-no-rattler-implication})}\\
    &= \max_{k \in [K]} \max_{i \in [n]}\min_{\vh_{k, i} \in \setS^{d-1}}\max_{k' \ne k} \langle \vw_{k'}^\star - \vw_{k}^\star , \vh_{k, i} \rangle\\
    &= \min_{\mH \in \setS^{d-1}} \max_{k \in [K]} \max_{i \in [n]}\max_{k' \ne k} \langle \vw_{k'}^\star - \vw_{k}^\star , \vh_{k, i} \rangle\\
    &\ge \min_{\mW\in\OB(d,K)} \min_{\mH \in \setS^{d-1}} \max_{k \in [K]} \max_{i \in [n]}\max_{k' \ne k} \langle \vw_{k'} - \vw_{k} , \vh_{k, i} \rangle && \text{(Property of $\max$)}\\
    &= \min_{\mW\in\OB(d,K)} \min_{\mH \in \setS^{d-1}} \mathcal{L}_{\text{HardMax}}(\mW, \mH) && \text{(Definition of $\mathcal{L}_{\text{HardMax}}$)}\\
    &= \mathcal{L}_{\text{HardMax}}(\mW^\star, \mH^\star) && \text{(Eq. \eqref{eq:prf-nc3-def-wh-star})}
\end{aligned}
\end{equation}
which is a contradiction. 
Hence, we have proved that \eqref{eq:prf-nc3-sc-optimal-h} holds true.
Now, using \eqref{eq:prf-nc3-sc-optimal-h} we have that for any $\bar{k} \in [K]$ and $\bar{i} \in [n]$, it holds that
\begin{equation}\label{eq:prf-nc3-sc-result}
\begin{aligned}
    &\max_{k' \ne \bar{k}} \langle \vw_{k'}^\star - \vw_{\bar{k}}^\star, \vh_{\bar{k}, \bar{i}}^\star \rangle 
    \le \max_{k' \ne \bar{k}} \langle \vw_{k'}^\star - \vw_{\bar{k}}^\star, \vw_{\bar{k}}^\star \rangle \\
    \implies &\left(\max_{k' \ne \bar{k}} \langle \vw_{k'}^\star, \vh_{\bar{k}, \bar{i}}^\star \rangle\right) - \langle \vw_{\bar{k}}^\star, \vh_{\bar{k}, \bar{i}}^\star  \rangle
    \le \left( \max_{k' \ne \bar{k}} \langle \vw_{k'}^\star, \vw_{\bar{k}}^\star \rangle \right) - \langle \vw_{\bar{k}}^\star, \vw_{\bar{k}}^\star \rangle \\
    \implies &\left(\max_{k' \ne \bar{k}} \langle \vw_{k'}^\star, \vh_{\bar{k}, \bar{i}}^\star \rangle\right)
    \le \left( \max_{k' \ne \bar{k}} \langle \vw_{k'}^\star, \vw_{\bar{k}}^\star \rangle \right) + \langle \vw_{\bar{k}}^\star, \vh_{\bar{k}, \bar{i}}^\star  \rangle - 1.
\end{aligned}
\end{equation}

On the other hand, using the result that Tammes problem and Softmax Code are equivalent, and the fact that $\mW ^\star$ is a Softmax Code, we know that $\mW ^\star$ is also a solution to the Tammes problem, \ie,
\begin{equation}\label{eq:prf-nc3-tammes-def-w}
    \mW^\star = \argmin_{\mW\in\OB(d,K)} \max_{k \in [K]} \max_{k' \ne k} \langle \vw_{k'}, \vw_k \rangle.
\end{equation}
Hence, it must hold that for any $\bar{k} \in [K]$ and $\bar{i} \in [n]$,
\begin{equation}\label{eq:prf-nc3-tammes-optimal-h}
    \max_{k' \ne \bar{k}} \langle \vw_{k'}^\star, \vw_{\bar{k}}^\star \rangle \le \max_{k' \ne \bar{k}} \langle \vw_{k'}^\star, \vh_{\bar{k}, \bar{i}}^\star \rangle.
\end{equation}
To see why \eqref{eq:prf-nc3-tammes-optimal-h} holds, assume for the purpose of arriving at a contradiction that it does not hold, \ie, 
\begin{equation}\label{eq:prf-nc3-tammes-optimal-h-opposite}
    \max_{k' \ne \bar{k}} \langle \vw_{k'}^\star, \vw_{\bar{k}}^\star \rangle > \max_{k' \ne \bar{k}} \langle \vw_{k'}^\star, \vh_{\bar{k}, \bar{i}}^\star \rangle.
\end{equation}
Then, consider $\mW^0 \in\OB(d,K)$ with $\vw^0_{k} := \vh_{\bar{k}, \bar{i}}^\star$ for $k = \bar{k}$, and $\vw^0_{k} := \vw^\star_{k}$ otherwise. 
It can be verified that
\begin{equation}
\begin{aligned}
    \max_{k \in [K]} \max_{k' \ne k} \langle \vw_{k'}^0, \vw_k^0 \rangle 
    &= \max\left(  \max_{k' \ne \bar{k}} \langle \vw_{k'}^0, \vw_{\bar{k}}^0 \rangle, \max_{\{k, k'\} \subseteq [K] \setminus \{\bar{k}\}, k' \ne k} \langle \vw_{k'}^0, \vw_k^0 \rangle  \right)\\
    &= \max\left(  \max_{k' \ne \bar{k}} \langle \vw_{k'}^\star, \vh_{\bar{k}, \bar{i}}^\star \rangle, \max_{\{k, k'\} \subseteq [K] \setminus \{\bar{k}\}, k' \ne k} \langle \vw_{k'}^\star, \vw_k^\star \rangle  \right)\\
    &\le \max\left(  \max_{k' \ne \bar{k}} \langle \vw_{k'}^\star, \vh_{\bar{k}, \bar{i}}^\star \rangle, \max_{\{k, k'\} \subseteq [K], k' \ne k} \langle \vw_{k'}^\star, \vw_k^\star \rangle  \right)\\
    &\le \max_{\{k, k'\} \subseteq [K], k' \ne k} \langle \vw_{k'}^\star, \vw_k^\star \rangle 
    = \max_{k \in [K]} \max_{k' \ne k} \langle \vw_{k'}^\star, \vw_k^\star \rangle,
\end{aligned}
\end{equation}
where the last inequality is obtained from using \eqref{eq:prf-nc3-tammes-optimal-h-opposite}. 
This implies, using the definition of $\mW^\star$ in \eqref{eq:prf-nc3-tammes-def-w}, that $\mW^0$ is also a solution to the Tammes problem. 
Moreover, because of \eqref{eq:prf-nc3-tammes-optimal-h-opposite} it can be seen that $\vw^0_{\bar{k}}$ is a rattler, contradicting with the assumption that the Tammes problem has no rattlers. 
Hence, we have proved that \eqref{eq:prf-nc3-tammes-optimal-h} holds true.
Combining \eqref{eq:prf-nc3-tammes-optimal-h} with \eqref{eq:prf-nc3-sc-result}, we have
\begin{equation}
    \max_{k' \ne \bar{k}} \langle \vw_{k'}^\star, \vw_{\bar{k}}^\star \rangle 
    \le \max_{k' \ne \bar{k}} \langle \vw_{k'}^\star, \vh_{\bar{k}, \bar{i}}^\star \rangle 
    \le \left( \max_{k' \ne \bar{k}} \langle \vw_{k'}^\star, \vw_{\bar{k}}^\star \rangle \right) + \langle \vw_{\bar{k}}^\star, \vh_{\bar{k}, \bar{i}}^\star  \rangle - 1, 
\end{equation}
hence $\langle \vw_{\bar{k}}^\star, \vh_{\bar{k}, \bar{i}}^\star  \rangle \ge 1$.
Since $\vw_{\bar{k}}^\star$ and $ \vh_{\bar{k}, \bar{i}}^\star$ are of unit $\ell_2$ norm, we have $\vh_{\bar{k}, \bar{i}}^\star = \vw_{\bar{k}}^\star$, which concludes the proof.
\end{proof}

\subsection{Proof of \Cref{thm:Tammes-equivalent-Softmax}}
The equivalence can be established by noting that the optimal solutions for the Tammes problem for the case $d = 2$ \citep{cohn2022small} and $K \le d+1$ \citep{fickus2017brief} are the same as those for the Softmax codes in \Cref{thm:special-case-NC3}.

\subsection{Proof of \Cref{thm:softmax-code-objective-bound}}

We first show that one-vs-rest and one-vs-one distances are mutually bounded by each other.

\begin{theorem} For any classifier weights $\mW$ with normalization $\norm{\vw_k}{2}=1,\forall~k\in[K]$, the one-vs-rest distance and one-vs-one distance obey the following relation:
\begin{align}\label{eq:margin-equivalency}
    \frac{\rho^2_\onevsone(\mW) }{2} \le \rho_\onevsrest(\mW)  \le \rho_\onevsone(\mW) .
\end{align}
\end{theorem}

\begin{proof} On one hand, according to the definitions of the two margins, it is clear that $\rho_\onevsrest(\mW) \le \rho_\onevsone(\mW)$. On the other hand, according to \Cref{thm:max_min_separation_margin}, we have
\begin{align*}
\rho_\onevsrest(\mW) &=\min_{k} \dist(\vw_k,\{\vw_j\}_{j\in[K]\setminus k}) =  \min_{k} \max_{\mH\in\OB(d,K)} \ \paren{  - \max_{j \neq k}\langle \vw_{j} - \vw_k, \vh_k \rangle }\\
& \ge \min_{k} \paren{1 - \max_{j\neq k}\langle \vw_j, \vw_k \rangle  } = \min_{k}  \min_{j\neq k} \frac{\|\vw_j - \vw_k\|^2}{2}
\end{align*}
where the first inequality follows by setting $\vh_k = \vw_k$. The proof is completed by noting that $\rho^2_\onevsone(\mW) = \min_{k}  \min_{j\neq k} \|\vw_j - \vw_k\|^2$.
\end{proof}

Combining this theorem with the existing upper bound (see~\cite{MOORE1974213}) and a lower bound (see Lemma~\ref{thm:onevsone-lowerbound}) on the one-vs-one distance, we obtain the following result.

\begin{theorem} Assuming $K \geq \sqrt{2 \pi \sqrt{e} d}$ and letting $\Gamma(\cdot)$ denote the Gamma function, we have
\begin{equation}
    \frac{1}{2} \biggr[\frac{\sqrt{\pi}}{K}\frac{\Gamma\paren{\frac{d+1}{2}}}{\Gamma\paren{\frac{d}{2} +1}} \biggr]^{\frac{2}{d-1}} \le \max_{\mW \in \OB(d,K)}\rho_\onevsrest(\mW) \le 2 \biggr[ \frac{2\sqrt{\pi}}{K} \frac{\Gamma \paren{\frac{d+1}{2}}}{\Gamma\paren{\frac{d}{2}}} \biggr]^{\frac{1}{d-1}}.
\end{equation}\label{appeq:margin-bound}
\end{theorem}

Using the property $a^{1-s}<\Gamma(a+1)/\Gamma(a+s)< (a+1)^{1-s}$ for any $a>0,s\in(0,1)$, 
we can simplify the bounds in \eqref{eq:margin-bound} to
\begin{align*}
   \frac{1}{2} \left[\frac{\sqrt{\pi}}{K \sqrt{\frac{d}{2} +1}}\right]^{\frac{2}{d-1}} \le \max_{\mW \in \OB(d,K)}\rho_\onevsrest(\mW) \le 2 \left[ \frac{2\sqrt{\pi (\frac{d+1}{2})}}{K}  \right]^{\frac{1}{d-1}},
\end{align*}

\begin{lemma}\label{thm:onevsone-lowerbound}
For any $K$ and $d$, we have
\begin{equation}\label{eq:onevsone-lowerbound}
    \biggr[\frac{\sqrt{\pi}}{K}\frac{\Gamma\paren{\frac{d+1}{2}}}{\Gamma\paren{\frac{d}{2} +1}} \biggr]^{\frac{1}{d-1}} \le \max_{\mW \in \OB(d,K)}\rho_\onevsone(\mW).
\end{equation}
\end{lemma}

\begin{proof}
    We prove the theorem by constructing a $\widehat{\mW} \in \OB(d,K)$ and deriving a lower bound on $\rho_\onevsone(\widehat{\mW})$. 
    Then, this lower bound is a lower bound for $\max_{\mW \in \OB(d,K)}\rho_\onevsone(\mW)$ as well owning to the relation 
    \begin{equation}\label{eq:prf-rho-onevsone-bound-by-construction}
        \rho_\onevsone(\widehat{\mW}) \le \max_{\mW \in \OB(d,K)}\rho_\onevsone(\mW).
    \end{equation} 
    The tightness of this lower bound naturally depends on the choice of $\widehat{\mW}$, which we explain below.

    Instead of constructing $\widehat{\mW}$ directly \emph{for a given $K$}, we first consider the construction of a $\mW^0(\rho_0)$ \emph{for any given $\rho_0 > 0$}. 
    Later, we will specify a $\rho_0$ and construct a $\widehat{\mW}$ from $\mW^0(\rho_0)$. 
    To simplify the notation, we write $\mW^0(\rho_0)$ as $\mW^0$.
    
    We construct $\mW^0$ using the following procedure from the proof of \cite[Lemma 12]{you2018sparse}:
    \begin{itemize}[leftmargin=0.12in,topsep=0.01em,itemsep=0.01em]
    \item Set $\vw_1^0$ to be an arbitrary vector in $\mathbb{S}^{d-1}$.
    \item For any $k > 1$, take $\vw_k^0$ to be any vector in $\mathbb{S}^{d-1}$ that satisfies $\|\vw_k^0 - \vw_{k'}^0\|_2 > \rho_0$, for all $k' < k$.
    \item Terminate the process when no such point exists. 
    \end{itemize}
    It is easy to see that this procedure terminates in finite number of steps; see \cite[Lemma 12]{you2018sparse} for a rigorous argument. 
    Assume that the total number of generated vectors is $K_0$. Collect these vectors into the columns of a matrix $\mW^0$.  
    By construction, $\mW^0$ has the following property, which will be used later:
    \begin{equation}\label{eq:prf-rho-w0-bound}
        \rho_\onevsone(\mW^0) > \rho_0.
    \end{equation} 

    We now derive a lower bound for $K_0$. 
    First, note that $\mW^0$ provides a \emph{$\rho_0$-covering} of the unit sphere $\mathbb{S}^{d-1}$. 
    That is, for any $\vv \in \mathbb{S}^{d-1}$, there must exist a $k \in \{1, \ldots, K_0\}$ such that $\|\vv - \vw_k^0\|_2 \le \rho_0$. 
    Otherwise, there would exist a $\vw \in \mathbb{S}^{d-1}$ that satisfies $\|\vw - \vw_{k}^0\|_2 > \rho_0$ for all $k \le K_0$, contradicting the termination condition in the construction of $\mW^0$.
    
    Geometrically, a {$\rho_0$-covering} is a set of points such that, the union of Euclidean balls of radius $\rho_0$ centered at those points cover the entire unit sphere. 
    Leveraging this interpretation, we provide a geometric method for bounding the number of points in a {$\rho_0$-covering} from below. 
    Concretely, given any $\vw \in \mathbb{S}^{d-1}$, we denote $\mathbb{S}^{d-1}_{\rho_0}(\vw) = \{ \vv \in \mathbb{S}^{d-1}, \|\vv - \vw\|_2 \le \rho_0\}$, which is the spherical cap centered at $\vw$ with radius $\rho_0$. 
    Since $\mW^0$ provides a $\rho_0$-covering, we have
    \begin{equation}
        \bigcup_{k=1}^{K_0}\mathbb{S}^{d-1}_{\rho_0}(\vw_k^0) \subseteq \mathbb{S}^{d-1}
        \implies
        \sum_{k=1}^{K_0} \sigma_{d-1}(\mathbb{S}^{d-1}_{\rho_0}(\vw_k^0)) \ge \sigma_{d-1}(\mathbb{S}^{d-1}),
    \end{equation}
    where $\sigma_{d-1}$ denotes the uniform area measure on $\mathbb{S}^{d-1}$.
    By noting that $\sigma_{d-1}(\mathbb{S}^{d-1}_{\rho_0}(\vw_k^0))$ is independent of $k$, we obtain
    \begin{equation}\label{eq:prf-k0-spherical-cap}
        K_0 \cdot \sigma_{d-1}(\mathbb{S}^{d-1}_{\rho_0}(\vw)) \ge \sigma_{d-1}(\mathbb{S}^{d-1}),
    \end{equation}
    for an arbitrary choice of $\vw \in \mathbb{S}^{d-1}$. 
    In above, note that the quantity $\sigma_{d-1}(\mathbb{S}^{d-1}_{\rho_0}(\vw)) / \sigma_{d-1}(\mathbb{S}^{d-1})$ is the proportion of the area of $\mathbb{S}^{d-1}$ that lies in the spherical cap $\mathbb{S}^{d-1}_{\rho_0}(\vw)$. 
    By a geometric argument, \cite[Lemma 9]{you2018sparse} provides an upper bound for it which we rewrite here: 
    \begin{equation}\label{eq:prf-spherical-cap-bound}
        \frac{\sigma_{d-1}(\mathbb{S}^{d-1}_{\rho_0}(\vw))}{\sigma_{d-1}(\mathbb{S}^{d-1})} \le \frac{v_{d-1}}{v_d} \left( \rho_0 \sqrt{1 - \frac{\rho_0^2}{4}} \right)^{d-1}.
    \end{equation}
    In above, $v_{d} \doteq \frac{\pi^{\frac{d}{2}}}{\Gamma(\frac{d}{2} + 1)}$ is the volumn of the unit Euclidean ball in $\mathbb{R}^d$. 
    Combining \eqref{eq:prf-spherical-cap-bound} with \eqref{eq:prf-k0-spherical-cap}, we obtain a lower bound on $K_0$ as 
    \begin{equation}\label{eq:prf-k0-bound}
        K_0 
        \ge \frac{\sigma_{d-1}(\mathbb{S}^{d-1})}{\sigma_{d-1}(\mathbb{S}^{d-1}_{\rho_0}(\vw))} 
        \ge \frac{v_d}{v_{d-1}} \frac{1}{\left( \rho_0 \sqrt{1 - \frac{\rho_0^2}{4}} \right)^{d-1}}
        \ge \frac{v_d}{v_{d-1}} \frac{1}{\rho_0^{d-1}}.
    \end{equation}
    
    We are now ready to construct a $\widehat{\mW} \in \OB(d,K)$. 
    Take 
    \begin{equation}\label{eq:prf-define-rho0}
        \rho_0 = \left(\frac{v_d}{v_{d-1} \cdot K}\right)^\frac{1}{d-1}.
    \end{equation}  
    By using \eqref{eq:prf-k0-bound} we have $K_0 \ge K$.
    Hence, we may construct $\widehat{\mW}$ as the set of any $K$ distinct columns of $\mW^0$.
    In particular, from \eqref{eq:prf-rho-onevsone-bound-by-construction}, \eqref{eq:prf-rho-w0-bound} and \eqref{eq:prf-define-rho0}, we obtain
    \begin{multline}
        \max_{\mW \in \OB(d,K)}\rho_\onevsone(\mW) \ge \rho_\onevsone(\widehat{\mW}) \ge \rho_\onevsone(\mW^0) \\
        > \rho_0 = \left(\frac{v_d}{v_{d-1} \cdot K}\right)^\frac{1}{d-1} = \biggr[\frac{\sqrt{\pi}}{K}\frac{\Gamma\paren{\frac{d+1}{2}}}{\Gamma\paren{\frac{d}{2} +1}} \biggr]^{\frac{1}{d-1}} ,
    \end{multline}
    where the second inequality follows trivially from the definition of $\rho_\onevsone(\cdot)$.
\end{proof}

\end{document}